\documentclass[prodmode,acmtecs]{acmsmall} 
\usepackage[ruled]{algorithm2e}

\SetAlFnt{\small}
\SetAlCapFnt{\small}
\SetAlCapNameFnt{\small}
\SetAlCapHSkip{0pt}
\IncMargin{-\parindent}

\usepackage{times}
\usepackage{url}
\usepackage{latexsym}
\usepackage{helvet}
\usepackage{courier}
\usepackage{graphicx}
\usepackage{tabularx}
\usepackage{verbatim}
\usepackage{subfig}
\usepackage{amsmath}
\usepackage{stfloats}
\usepackage{amsfonts}
\usepackage{lipsum}
\usepackage{color}

\usepackage{epstopdf}
\usepackage{flushend}

\usepackage{labelfig}
\usepackage{algorithmic}
\newcolumntype{C}[1]{>{\centering\let\newline\\\arraybackslash\hspace{0pt}}m{#1}}
\usepackage{array}
\usepackage{ragged2e}
\newcolumntype{P}[1]{>{\RaggedRight\hspace{0pt}}p{#1}}
\usepackage{pbox}

\usepackage[para,online,flushleft,normal]{threeparttable}

\newcommand{\nop}[1]{}

\newcommand{\eg}{{e.g.}}
\newcommand{\ie}{{i.e.}}
\newcommand{\etc}{{etc.}}

\newcommand{\comments}[1]{}

\begin{document}
\markboth{C. Wang et al.}{World Knowledge as Indirect Supervision for Document Clustering}

\title{World Knowledge as Indirect Supervision for Document Clustering}
\author{CHENGUANG WANG
\affil{Peking University}
YANGQIU SONG
\affil{Hong Kong University of Science and Technology}
DAN ROTH
\affil{University of Illinois at Urbana-Champaign}
MING ZHANG\footnote{Corresponding Author: MING ZHANG}
\affil{Peking University}
JIAWEI HAN
\affil{University of Illinois at Urbana-Champaign}
}

\begin{abstract}
One of the key obstacles in making learning protocols realistic in applications is the need to supervise them, a costly process that often requires hiring domain experts.
We consider the framework to use the world knowledge as indirect supervision.
World knowledge is general-purpose knowledge, which is not designed for any specific domain.
Then the key challenges are how to adapt the world knowledge to domains and how to represent it for learning.
In this paper, we provide an example of using world knowledge for domain dependent document clustering.
We provide three ways to specify the world knowledge to domains by resolving the ambiguity of the entities and their types, and represent the data with world knowledge as a heterogeneous information network.
Then we propose a clustering algorithm that can cluster multiple types and incorporate the sub-type information as constraints.
In the experiments, we use two existing knowledge bases as our sources of world knowledge.
One is Freebase, which is collaboratively collected knowledge about entities and their organizations.
The other is YAGO2, a knowledge base automatically extracted from Wikipedia and maps knowledge to the linguistic knowledge base, WordNet.
Experimental results on two text benchmark datasets (20newsgroups and RCV1) show that incorporating world knowledge as indirect supervision can significantly outperform the state-of-the-art clustering algorithms as well as clustering algorithms enhanced with world knowledge features.

A preliminary version of this work appeared in
the proceedings of KDD 2015~\cite{wang2015knowledge}. This journal version has made several
major improvements. First, we have proposed a new and general learning framework for machine learning with world knowledge as indirect supervision, where document clustering is a special case in the original paper.
Second, in order to make our unsupervised semantic parsing method more understandable, we add several real cases from the original sentences to the resulting logic forms with all the necessary information.
Third, we add details of the three semantic filtering methods and conduct deep analysis of the three semantic filters, by using case studies to show why the conceptualization based semantic filter can produce more accurate indirect supervision.
Finally, in addition to the experiment on 20 newsgroup data and Freebase, we have extended the experiments on clustering results by using all the combinations of text (20 newsgroup, MCAT, CCAT, ECAT) and world knowledge sources (Freebase, YAGO2).

\end{abstract}

\terms{Data Mining, Algorithms, Performance}


\begin{bottomstuff}
Author's addresses: C. Wang {and} M. Zhang, School of EECS, Peking University; email: \{wangchenguang, mzhang\_cs\}@pku.edu.cn;
Y. Song, Department of Computer Science and Engineering, Hong Kong University of Science and Technology; email: yqsong@cse.ust.hk; D. Roth {and} J. Han, Department of Computer Science, University of Illinois at Urbana-Champaign; email: \{danr, hanj\}@illinois.edu.
\end{bottomstuff}

\maketitle

\section{Introduction}
Machine learning algorithms have become pervasive in multiple domains, impacting a wide variety of applications.
Nonetheless, a key obstacle in making learning protocols realistic in applications is the need to supervise them, a costly process that often requires hiring domain experts.
In the past decades, machine learning community has elaborated to reduce the labeling work done by human for supervised machine learning algorithms or to improve unsupervised learning with only minimum supervision.
For example, semi-supervised learning~\cite{ChaSchZie06} is proposed to use only partially labeled data and a lot of unlabeled data to perform learning with the hope that it can perform as good as fully supervised learning.
Transfer learning~\cite{Pan2010} uses the labeled data from other relevant domains to help the learning task in the target domain.
However, there are still many cases that neither semi-supervised learning nor transfer learning can help.
For example, in the era of big data, we can have a lot textual information from different Web sites, \eg, blogs, forums, mailing lists.
It is impossible to ask human to annotate all the required tasks.
It is also difficult to find relevant labeled domains.
Recognizing that some domains can be very specific and really need the domain experts to perform annotation, \eg, the medical domain publication classification.
Therefore, we should consider a more general approach to further reducing the labeling cost for learning tasks in diverse domains.

Fortunately, with the proliferation of general-purpose knowledge bases (or knowledge graphs), e.g., Cyc project~\cite{researchCyc}, Wikipedia, Freebase~\cite{freebase}, KnowItAll~\cite{knowitall}, TextRunner~\cite{BankoCSBE07}, ReVerb~\cite{FaderSE11}, Ollie~\cite{MausamSSBE12}, WikiTaxonomy~\cite{PonzettoS07}, Probase~\cite{wu2011taxonomy}, DBpedia~\cite{auer2007dbpedia}, YAGO~\cite{suchanek2007yago}, NELL~\cite{DBLP:conf/aaai/MitchellCHTBCMG15} and Knowledge Vault~\cite{dong2014knowledge}, we have an abundance of available world knowledge.
We call these knowledge bases world knowledge~\cite{gabrilovich2005feature}, because they are universal knowledge that are either collaboratively annotated by human labelers or automatically extracted from big data.
In general, world knowledge can be common sense knowledge, common knowledge, or domain dependent knowledge.
Common sense knowledge is the facts that an ordinary person is expected to know, but we seldom need to learn them on purpose.
For example, we all know that a dog is an animal.
Common knowledge is widely accessible knowledge. For example, the population of Illinois is around 12.8M. An ordinary person may not know the exact number of population, but we would find the answer easily from numerous sources.
Domain knowledge can be very specific, such as the complete set of species of animals, or the taxonomy classification of trees.

When world knowledge is annotated or extracted, it is not collected for any specific domain.
However, because we believe the facts in world knowledge bases are very useful and of high quality,
we propose using them as supervision for many machine learning problems.
People have found it useful to use world knowledge as distant supervision for entity and relation extraction and embedding~\cite{mintz2009distant,wang2014knowledge,xu2014rc}.
This is a direct use of the facts in world knowledge bases, where the entities in the knowledge bases are matched in the context regardless the ambiguity.
A more interesting question is {\it can we use the world knowledge to indirectly ``supervise'' more machine learning algorithms or applications?}
For example, if we can use world knowledge as indirect supervision, then we can extend the knowledge about entities and relations to more generic text analytics problems, \eg, categorization and information retrieval.

\begin{figure*}[t]
\centering
\includegraphics[width=1.0\textwidth]{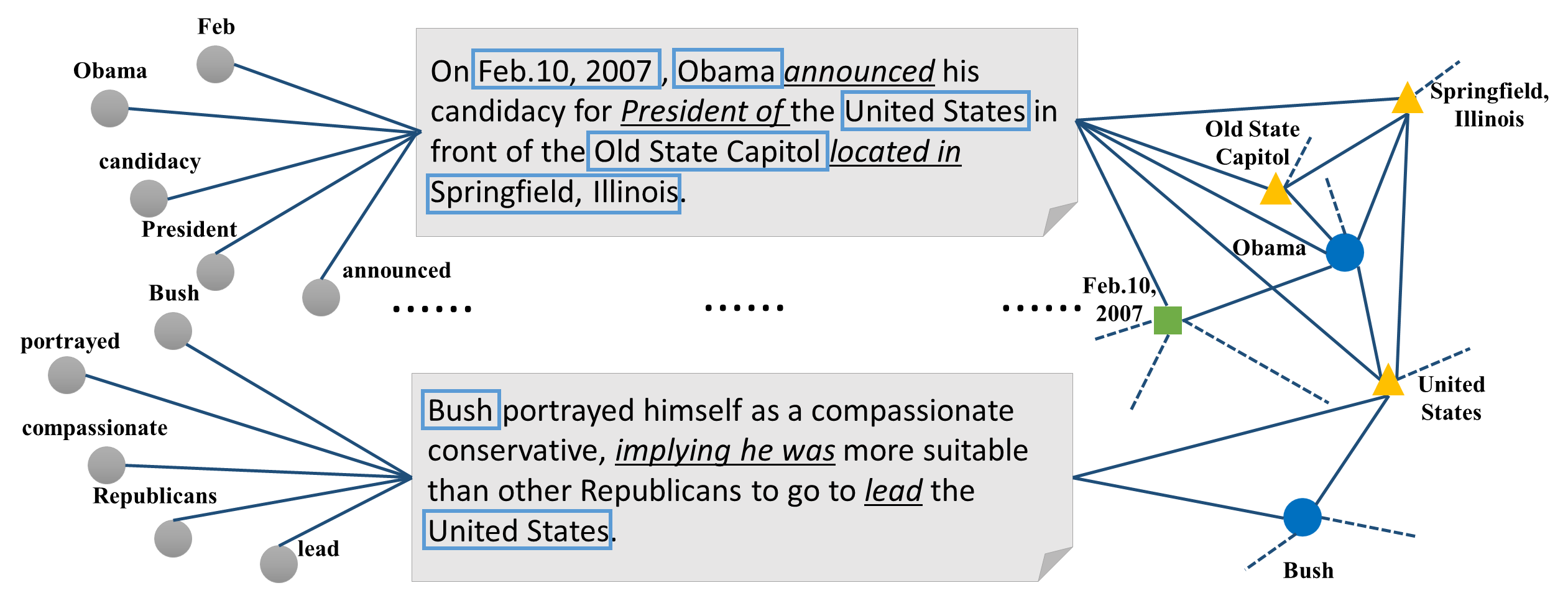}
\caption{Heterogeneous information network example. The network $\mathcal G$ contains five entity types: document, word, date, person and location, which are represented with gray rectangle, gray round, green square, blue round, and yellow triangle, respectively.}
\label{fig:hinex}
\end{figure*}

Thus, we consider a general machine learning framework that can incorporate world knowledge into machine learning algorithms.
In general, there are three challenges of incorporating world knowledge into machine learning algorithms: (1) {\it domain specification}, (2) {\it knowledge representation}, and (3) {\it propagation of indirect supervision}.

First, as mentioned, world knowledge is not designed for any specific domain.
For example, when we want to cluster the documents about entertainment or sports, then the world knowledge about names of celebrities and athletes may help while the terms used in science and technology may not be very useful.
Thus, a key issue is how we should adapt world knowledge to the domain specific tasks.
Second, when we have the world knowledge, we should consider how we can represent it for the domain dependent tasks.
An intuitive way is to use knowledge bases to generate features for machine learning algorithms~\cite{gabrilovich2005feature,SongLLW15}.
We call these features ``flat'' because they do not consider the link information in the knowledge bases.
However, most knowledge bases use a linked network to organize the knowledge.
For example, a {\it CEO} is connected to an {\it IT company}, and the {\it IT company} is a {\it company}.
Thus, the structure of the knowledge also provides rich information about the connections of entities and relations.
Therefore, we should also carefully consider the best way to represent the world knowledge for machine learning algorithms.
Third, given the world knowledge about entities and their relations, as well as the types of entities and relations, we should consider an effective algorithm that can propagate the knowledge about entity and relation categories to the categories of data that contain the entities and relations.
This is a non-trivial task because we should consider both the data representation and the structural representation of the world knowledge.

In this paper, we illustrate the framework of machine learning with world knowledge using a document clustering problem.
We select two knowledge bases, \ie, Freebase, YAGO2, as the sources of world knowledge.
Freebase~\cite{freebase} is a collaboratively collected knowledge base about entities and their organizations.
YAGO2~\cite{suchanek2007yago} is a knowledge base automatically extracted from Wikipedia and maps the knowledge to the linguistic knowledge base, WordNet~\cite{wordnet}.
To adapt the world knowledge to domain specific tasks, we first use semantic parsing to ground any text to the knowledge bases~\cite{berant2013semantic}.
We then apply entity frequency, document frequency, and conceptualization~\cite{song2015short} based semantic filters to resolve the ambiguity problem when adapting world knowledge to the domain tasks.
After that, we have the documents as well as the extracted entities and their relations.
Since the knowledge bases provide the entity types, the resulting data naturally form a heterogeneous information network (HIN)~\cite{Han2010MKD}. We show an example of such HIN in Figure~\ref{fig:hinex}. The specified world knowledge, such as named entities (``Bush'', ``Obama'') and their types ({\it Person}), as well as the documents and the words form the HIN.
We then formulate the document clustering problem as an HIN partitioning problem, and provide a new algorithm to better perform clustering by incorporating the rich structural information as constraints in the HIN.
For example, the HIN builds a link (a must-link constraint) between ``Obama'' of sub-type {\it Politician} in one document and ``Bush'' of sub-type {\it Politician} in another document. Such link and type information could be very useful if the target clustering domain is ``Politics.''
Then we use the sub-type information as supervision information for the entities, and propagate the information to the documents through the network. Therefore we call the sub-type information as indirect supervision of documents.

The main contributions of this work are highlighted as follows:
\begin{itemize}
\item We propose a new learning framework of machine learning with world knowledge as indirect supervision. We give the general data mining and machine learning framework, and use a specific problem of document clustering to illustrate the process.
\item We propose to use semantic parsing and semantic filtering to specify world knowledge to the domain dependent documents, and develop a new constrained HIN clustering algorithm to make better use of the structural information from the world knowledge for document clustering task.
\item We conduct experiments on two benchmark datasets (20newsgroups and RCV1) to evaluate the clustering algorithm using HIN, compared with the state-of-the-art document clustering algorithms and clustering with ``flat'' world knowledge features. We show that our approach can be $13.3\%$ better than the semi-supervised clustering algorithm incorporating 250K constraints which are generated by ground-truth labels.
\end{itemize}

This paper is an extension of our previous work~\cite{wang2015knowledge}. We make the detailed algorithms clearer, and illustrate the effectiveness and efficiency of the algorithms with more extensive experiments.

The remainder of the paper is organized as follows.
Section~\ref{sec:mlframework} introduces the general learning framework of machine learning with world knowledge.
Section~\ref{sec:framework} presents our world knowledge specification approach.
The representation of world knowledge is introduced in Section~\ref{sec:representation}.
The model for document clustering with world knowledge is shown in Section~\ref{sec:clustering}.
Experiments and results are discussed in Section~\ref{sec:exp}.
Section~\ref{sec:related} discusses the related work, and we conclude this study in Section~\ref{sec:con}.

\section{{Machine Learning} with World Knowledge Framework}
\label{sec:mlframework}
\begin{figure*}[t]
\centering
\includegraphics[width=0.7\textwidth]{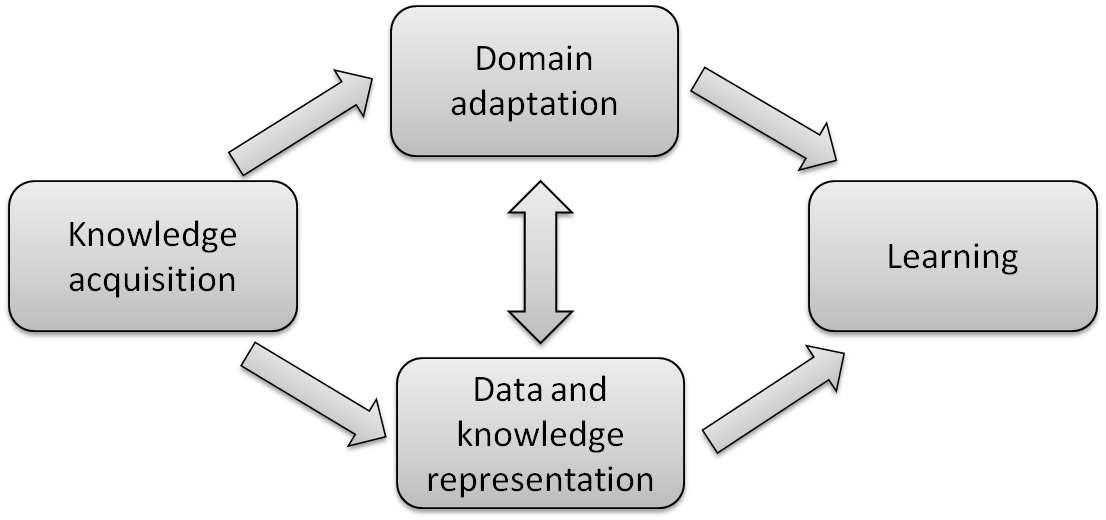}
\caption{Major steps of incorporating world knowledge into machine learning algorithms.}
\label{fig:steps}
\end{figure*}

In this section, we discuss the general framework on how we enable world knowledge to indirectly ``supervise'' machines, and give an overview on how to conduct document clustering in this framework.
{
In general, performing machine learning with world knowledge, we should follow four steps.
\begin{enumerate}
  \item Knowledge acquisition. World knowledge acquisition is a challenging problem. There exist some world knowledge bases. They are either collaboratively constructed by humans (such as Cyc project~\cite{researchCyc}, Wikipedia, Freebase~\cite{freebase}) or automatically extracted from big data (such as KnowItAll~\cite{knowitall}, TextRunner~\cite{BankoCSBE07},  ReVerb~\cite{FaderSE11}, Ollie~\cite{MausamSSBE12}, WikiTaxonomy~\cite{PonzettoS07}, Probase~\cite{wu2011taxonomy}, DBpedia~\cite{auer2007dbpedia}, YAGO~\cite{suchanek2007yago}, NELL~\cite{DBLP:conf/aaai/MitchellCHTBCMG15} and Knowledge Vault~\cite{dong2014knowledge}). Since we assume the world knowledge is given, we skip this step in this study. We select two knowledge bases in this paper, which are Freebase and YAGO2. Different knowledge bases have different characteristics. For example, Freebase is collaboratively collected. It focuses on named entities and their organizations. YAGO2 is automatically extracted from Wikipedia and mapped to WordNet. It will be interesting to compare the effects of using different world knowledge bases.
  \item Data adaptation. Given the world knowledge, it is not necessary to use the whole knowledge base to perform inference, since not all the world knowledge is related to the specific domains. Therefore, we should consider specifying the world knowledge to the domain dependent data, and adapting the world knowledge to better characterize the specific domains. Moreover, since the knowledge can be ambiguous without context, we should consider using domain dependent data to find the best knowledge to use. For example, when a text mentions ``apple,'' it can refer to a company or a fruit. In the knowledge base, we have both. Therefore, we should choose the right one to use. {Notice that the general data adaptation process contains the disambiguation phase. For example, in Section~\ref{subsec:sf}, we describe the way to data adaptation for documents including the entity disambiguation phrase~\cite{li2013mining} (semantic filtering procedure). The traditional entity disambiguation problem is focusing on leveraging the purely context information, such as the co-occurrence of words/phrases appearing in certain window of the entity to disambiguate the entity, while we are considering to use more information from the world knowledge base, such as the types of the near entities and relations to disambiguate the entities. Besides, we further explore the semantic context of the entity by considering the relations between entities in the world knowledge bases. Thus, disambiguation is more on entity side in the original raw documents, while data adaptation is more general.} 
  \item Data and knowledge representation. Traditionally, machine learning algorithms use feature vectors to represent data. Some interesting algorithms can represent data as trees or graphs, and compute kernel based on trees and graphs for machine learning~\cite{Collins2002,Vishwanathan2010}. Given the specified knowledge we have as well as the domain dependent data, we should use a better representation which considers the structure information of the linked knowledge rather than just considering the knowledge as flat features. Therefore, we propose to use a typed graph, which is called HIN to represent the data.
  \item Learning. After we have the representation, we can design a learning algorithm for domain dependent task. The learning algorithm is dependent to the problem as well as the data and knowledge representation. We will show how to handle the HIN for our clustering problem. Particularly, by representing the world knowledge as HIN, we have type information for the named entities we detected from the documents. Moreover, the world knowledge also provides the sub-type information. The coarse-grained type information is denser than the fine-grained sub-type information. For example, we can have much more entities annotated as {\it Person} than {\it Politician}. Thus, we use the coarse-grained type information to construct the HIN, and use the fine-grained sub-type information as further supervision for the entities, which is then used as indirect supervision for the documents.
\end{enumerate}
The illustration of these four steps are shown in Figure~\ref{fig:steps}, where steps two and three are sometimes dependent.
For example, we can do domain adaptation and knowledge representation jointly.}
The above four steps are general, which means they may apply to many applications.
In the following sections, we demonstrate how to select the right knowledge to use and to represent this knowledge for the task of document clustering.
After that, we will introduce the learning algorithm to perform better document clustering given the representation.

\begin{figure*}[t]
\centering
\includegraphics[width=0.7\textwidth]{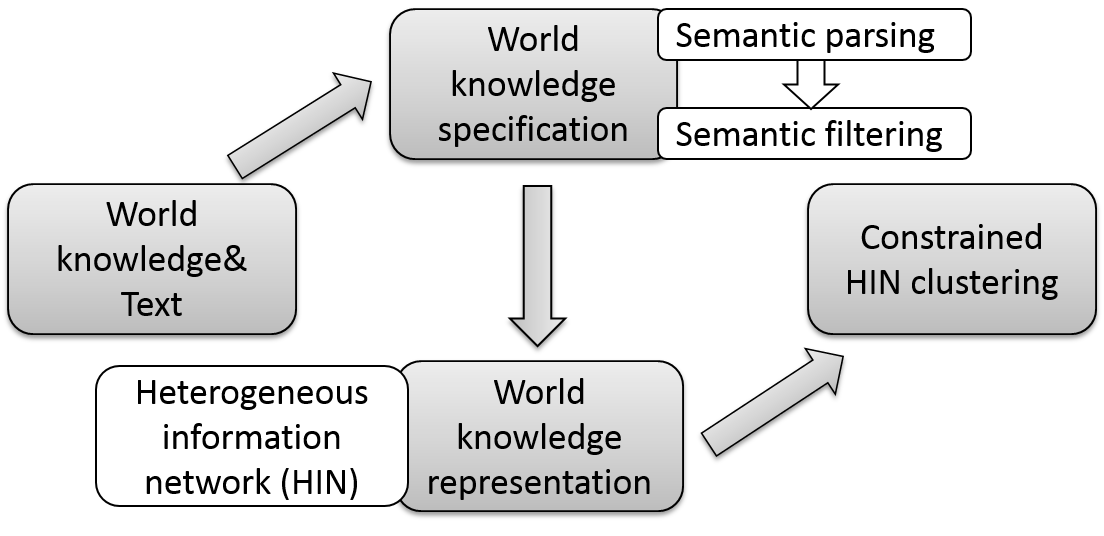}
\caption{{The overview of adapting machine learning with world knowledge framework to document clustering.}}
\label{fig:ov}
\end{figure*}

{In Figure~\ref{fig:ov}, we show the general overview of the framework illustrating the major procedures of
how to adapt the framework to the document clustering task. Notice that each module of the general
learning framework shown in Figure~\ref{fig:steps} is directly specified for the particular document clustering
with world knowledge approach (Figure~\ref{fig:ov}), e.g., data adaptation is specified as world knowledge specification.
Generally, we first assume the knowledge acquisition is done, i.e., the world knowledge bases (e.g.,
Freebase) are given. Second, robust unsupervised semantic parsing (Section~\ref{subsec:sp}) and three alternative semantic filters (Section~\ref{subsec:sf}) (e.g., conceptualization based semantic filtering) are introduced
for data adaptation. We third generate the new representation of the documents by modelling the unstructured texts in the heterogeneous information network (HIN) with entities including documents
themselves, words in the document set and relevant named entities and relations with proper types from the
knowledge bases. Finally, for the learning phase in the framework, based on the new HIN representation of the documents, we propose the constrained HIN clustering model to take the type and link
information obtained from the knowledge bases into consideration via constraints and probabilistic
distributions. In simpler terms, the first three steps are performed to construct the document-based
HIN. Then new learning algorithms (e.g., HIN based clustering algorithms) are proposed to handle
the document clustering in the new network representation and make use of the relevant knowledge specified from the world knowledge, to improve the performance of the traditional clustering algorithms.}

\section{World Knowledge Specification}
\label{sec:framework}
In this section, we propose a world knowledge specification approach to generate specified world knowledge given a set of domain dependent documents.
We first use semantic parsing to ground any text to the knowledge base, then provide three semantic filtering approaches to avoid ambiguity of the extracted information.

\subsection{Semantic Parsing}
\label{subsec:sp}
Semantic parsing is the task of mapping a piece of natural language text to a formal meaning representation~\cite{Mooney07}.
This can support question answering by querying a knowledge base~\cite{KwiatkowskiZGS11}.
Most previous semantic parsing algorithms or tools developed are for small scale problems but with complicated logical forms.
More recently, large scale semantic parsing grounding to world knowledge bases has been investigated, e.g., using Freebase~\cite{Krishnamurthy2012,CaiY13,KwiatkowskiCAZ13,berant2013semantic,BerantL14,YaoD14,ReddyLS14} or ReVerb~\cite{FaderZE13}.
These methods use semantic parsing to answer questions with the world knowledge bases.
Intuitively, they need to identify the candidate answers in the parsing results so that they can be ranked to answer the questions.
Therefore, they either need the question-answer pairs as supervision, or need a large amount of resources as well as the questions as distant/weak supervision~\cite{ReddyLS14}.
Similar to them, we are also working {with} very large scale world knowledge bases, but unlike them, we do not match question and answers.
We have already got all the entities in the document that can be matched to the world knowledge bases.
Our task is then to ground the text to the knowledge base entities and their relationships in the prescribed logical form.
Because we do not have and do not necessarily have the question-answer pairs, our problem is a fully unsupervised problem.
Then the remaining problems are two-folds.
First, we need to identify the relations between entities to map them to the knowledge base relations.
Second, we need to resolve the ambiguity of the entities and relations.

We first introduce the problem formulation and then introduce how we perform unsupervised semantic parsing.
Let $\mathcal{E}$ be a set of entities and $\mathcal{R}$ be a set of relations in the knowledge base.
Then the knowledge base $\mathcal K$ consists of triplets in the form of $(e_1, r, e_2)$, where $e_1, e_2 \in \mathcal E$ and $r \in \mathcal R$.
We follow~\cite{berant2013semantic} to use a simple version of Lambda Dependency-Based Compositional Semantics ($\lambda$-DCS)~\cite{Liang13} as the logic language to query the knowledge base.
We use $\lambda$-DCS because that it can generate logic forms simpler than lambda calculus forms. $\lambda$-DCS can reduce the number of variables by making existential quantification implicit. The logical form in simple $\lambda$-DCS is either in the form of unary (a subset of $\mathcal E$) or binary (a subset of $\mathcal E \times \mathcal E$).
We briefly introduce the definition of basic $\lambda$-DCS logical forms $x$ and the corresponding denotations $x_{\mathcal K}$ as below:
(1) Unary base: an entity $e \in \mathcal E$ is a unary logic form (\eg, {\it Obama}) with $e_{\mathcal K} = \{e\}$; (2) Binary base: a relation $r \in \mathcal R$ is a binary logic form (\eg, {\it PresidentofCountry}) with $r_{\mathcal K} = \{(e_1,e_2):(e_1,r,e_2)\in \mathcal K\}$; (3) Join: $b.u$ is a unary logic form, denoting a join and projection, where $b$ is a binary and $e$ is a unary. ${b.u}_{\mathcal K} = \{e_1 \in \mathcal E : \exists e_2.(e_1,e_2) \in b_{\mathcal K} \wedge e_2\in u_{\mathcal K}\}$ (\eg, {\it PresidentofCountry.Obama}); (4) Intersection: $u_1 \sqcap u_2$ ($u_1$ and $u_2$ are both unaries) denotes set intersection: ${u_1 \sqcap u_2}_{\mathcal K} = {u_1}_{\mathcal K} \cap {u_2}_{\mathcal K}$ (\eg, {\it Location.Olympics $\sqcap$ PresidentofCountry.Obama}).
\nop{For example, ``Gone with the Wind is written by Margaret Mitchell'' can be represented as {\it {Book.Written$\wedge$Work.People.Person}} in $\lambda$-DCS. {\color{red} this is not in the correct form! we need a better example here.}}

\begin{figure*}[t]
\centering
\includegraphics[width=0.7\textwidth]{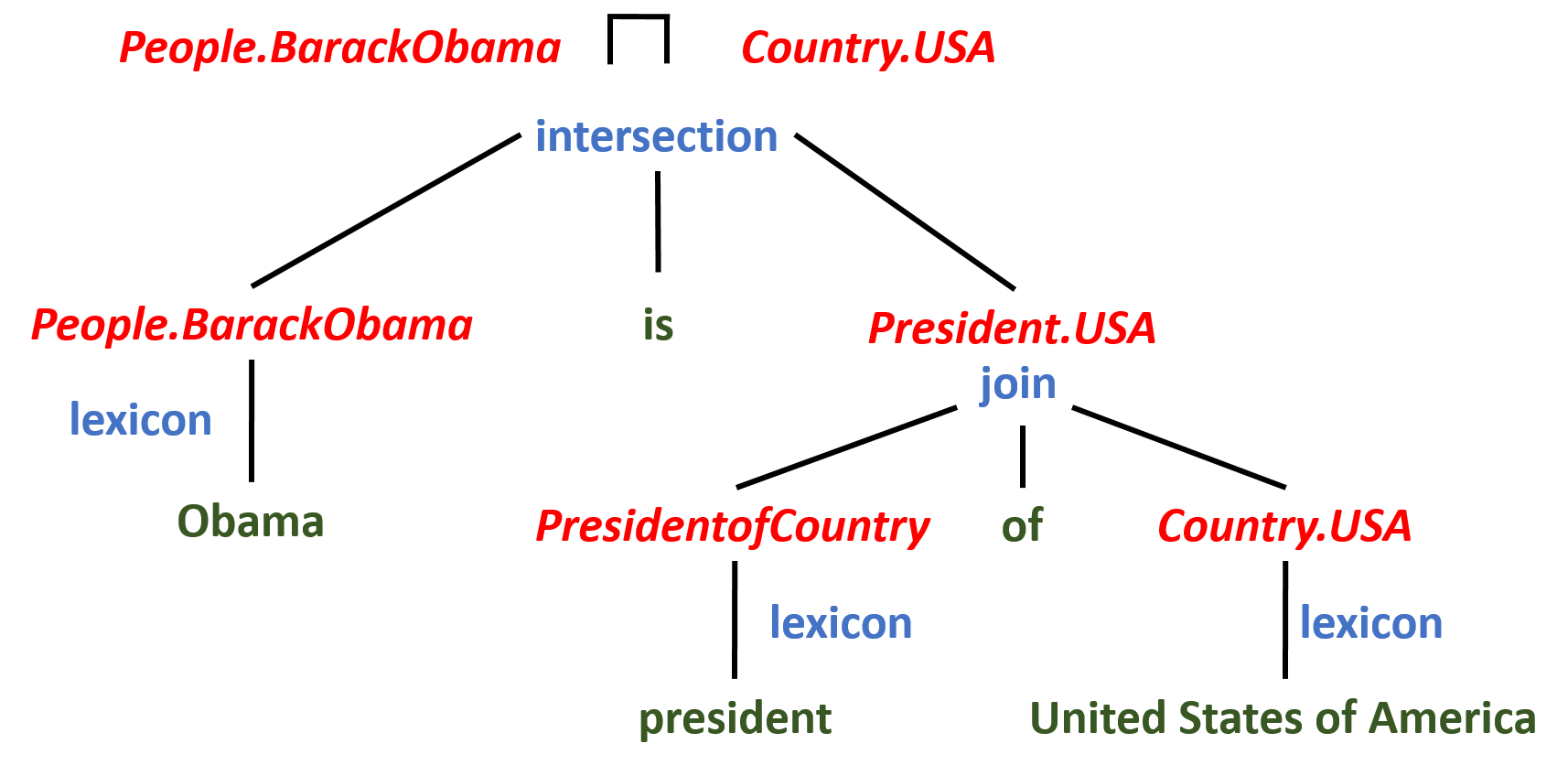}
\caption{Semantic parsing example. The figure shows a derivation $d$ of the input text ``Obama is president of United States.'' and the sub-derivations. Each is labeled with composition rule (in blue) and logical form (in red). The derivation $d$ ignores words ``is'' and ``of.''}
\label{fig:spex}
\end{figure*}

\begin{figure*}[t]
\centering
\includegraphics[width=1.0\textwidth]{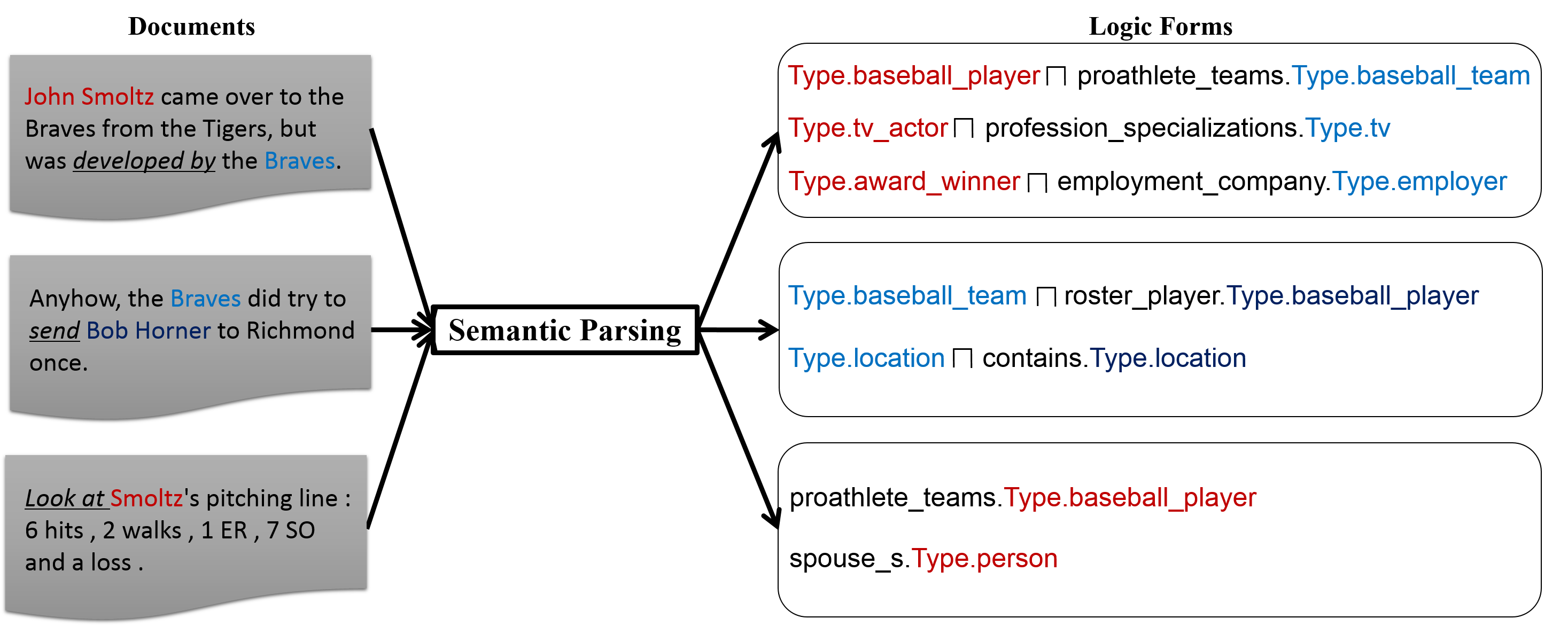}
\caption{A real example of the input documents and output logic forms for 20 newsgroups dataset. Left: a set of given documents; middle: semantic parser; right: resulting candidate logic forms for each document. we only show some examples of the parsed logic forms according to entities ``John Smoltz,'' ``Braves,'' and ``Bob Horner'' in the given documents for simplicity.}
\label{fig:spexp}
\end{figure*}

We generally adopt the semantic parsing framework proposed in~\cite{berant2013semantic}. Given a piece of text $s$, the semantic parser produces a distribution over possible derivations $D(s)$. Each derivation $d\in D(s)$ is a tree that indicates applying a certain set of composition rules that ends in the root of the tree, \ie, the logical form $d.z$.
We illustrate the semantic parsing process with an example as shown in Figure~\ref{fig:spex}.
In simpler terms, the semantic parsing process can be understood as the following.
First, given a piece of text ``Obama is president of United States of America,'' it maps the entities as well as the relation phrases in the text to knowledge base.
So ``Obama'' and ``United States of America'' are mapped to knowledge base, resulting in two unary logic forms {\it People.BarackObama} and {\it Country.USA}, where {\it People} and {\it Country} are the type information in Freebase.
The relation phrase ``president'' is mapped to a binary logic form {\it PresidentofCountry.} Notice that, the mapping process skips the words ``is'' and ``of.''
The mapping dictionary is constructed by aligning a large text corpus to the knowledge base. A phrase and a knowledge base entity or relation can be aligned if they co-occur with many of the same entities.
We select two knowledge bases, i.e., Freebase and YAGO2.
For Freebase, we just use the mapping already existing in the released tool shown in~\cite{berant2013semantic}.
For YAGO2, we follow~\cite{berant2013semantic} and download a subset of ClueWeb09\footnote{\url{http://www.lemurproject.org/clueweb09.php/}} to find the new mapping for YAGO2 entities and relations.
Second, it uses some rules (\ie, grammar) to combine the basic logic forms to generate the derivations, and rank the results (\ie, derivations).
In detail, the semantic parsing framework constructs a set of derivations for each span of the input text. First, for each span of the input text, it generates the single-predicate derivations based on the lexicon mapping from text to knowledge base predicates (\eg, ``president'' maps to {\it PresidentofCountry}). According to the set of composition rules, given any logical form $x_1$ and $x_2$ that are constructed over span $[i_1, j_1]$ and $[i_2, j_2]$, we then generate the logic forms based on the span $[min(i_1, i_2), max(j_1,j_2)]$ as the following: $x_1 \sqcap x_2$ (intersection), $x_1.x_2$ (join), or $x_1 \sqcap r.x_2$ (briging) for any relation $r \in \mathcal R$ (bridging operation is defined to generate additional predicates based on neighboring predicates).
For the example shown in this paragraph, {\it People.BarackObama $\sqcap$ President.USA} is generated to represent its semantic meaning. Notice that, {\it President.USA} is generated by joining the unary {\it Country.USA} with the binary {\it PresidentofCountry.} Figure~\ref{fig:spexp} shows a real example in 20 newsgroups dataset. Given the documents on the left side, the semantic parsing model is performed to generate the logic form candidates on the right side. Notice that, we only show some examples of the parsed logic forms according to entities ``John Smoltz,'' ``Braves,'' and ``Bob Horner'' in the given documents for simplicity.

\begin{figure*}[t]
\centering
\includegraphics[width=1.0\textwidth]{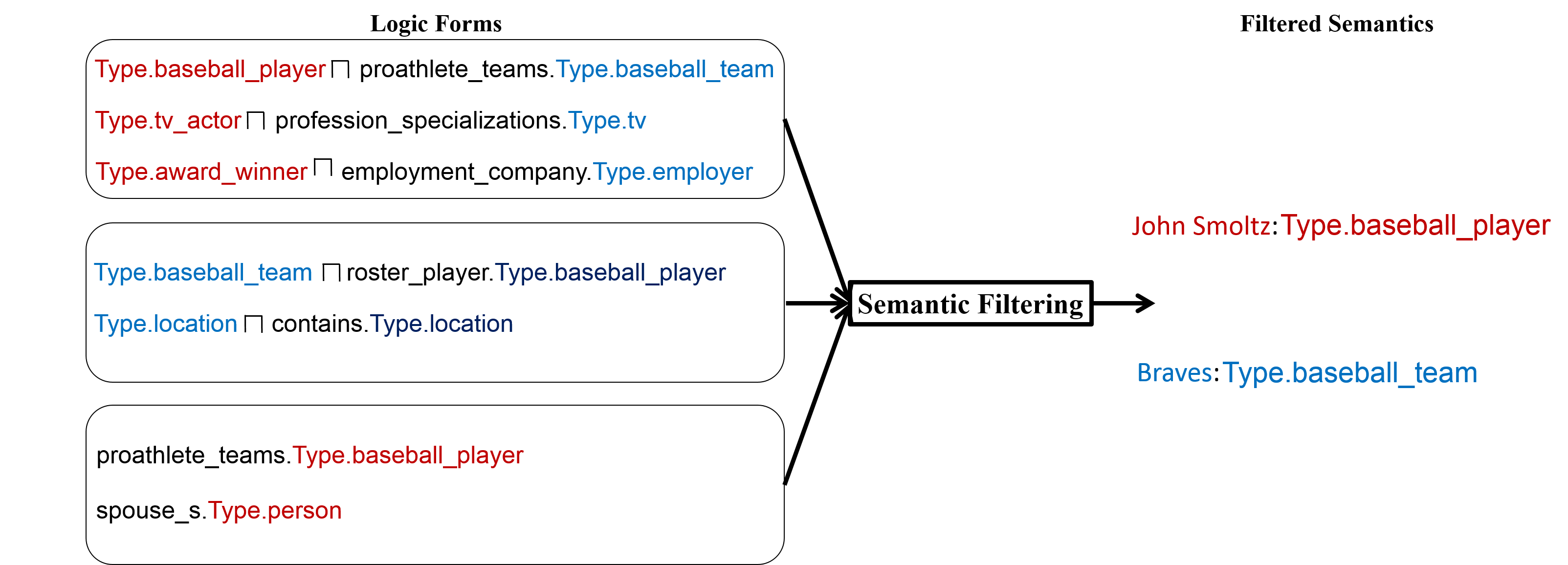}
\caption{Illustration of semantic filtering results based on the example shown in Figure~\ref{fig:spexp}. Left: candidate logic forms according to each document; middle: semantic filter; right: filtered semantics. ``John Smoltz'' is actually a baseball player, and ``Braves'' means the Atlanta Braves, which is a baseball team playing in national league. By using CBSF, the entity {ambiguation} problem is resolved to some extent. For example, ``Bob Horner'' could help ``John Smoltz'' disambiguate the correct entity type to be {\it baseball\_player} from {\it tv\_actor}, by using the context information provided in the first two documents. Similarly, the context information in the first and the third documents could help to disambiguate the entity type of ``Braves'' to be {\it baseball\_team}.}
\label{fig:sfexp}
\end{figure*}

When there are more than one candidate semantic meanings (\ie, derivations) for a sentence, in~\cite{berant2013semantic}, they learn the ranks based on the annotated question-answer pairs.
For our task, this annotation is not available.
Therefore, instead of ranking or enumerating all the possible logic forms (which is found to be not feasible in limited time), we constrain the entities to be the maximum length spanning phrases recognized by a state-of-the-art named entity recognition tool~\cite{RatinovRo09}. We then perform the two steps introduced above by using the maximum length spanning noun phrase as entities, and use the phrase between them in the text as relation phrase. We propose to use the following three semantic filtering methods to resolve the {ambiguation} problem.


\subsection{Semantic Filtering}
\label{subsec:sf}
For each sentence in the given document, the output of semantic parsing is a set of derivations that represent the semantic meaning. However, the extracted entities (\ie, unaries in the resulting derivations) can be ambiguous. For example, ``apple'' may be associated with type {\it Company} or {\it Fruit}.
Therefore, we should filter out the noisy entities and their types to ensure that the knowledge we have is good enough as indirect supervision for document clustering.
We assume that in the domain specific tasks, given the context, the entities seldom have multiple mutually exclusive meanings.
Given that we have the domain dependent corpus containing the documents to be clustered, we are given a lot of evidence to disambiguate the entities.
We propose the following three approaches to select the best knowledge to use for further learning process.

{\it Frequency based semantic filter (FBSF).} For each entity $e_{i}$ in document $d_j$, which can have multiple types choosing from all the types $t_1,\ldots,t_{\bar{T}}$ where we assume there are in total $\bar{T}$ types. Then we can use the frequency $n_{d_j}(e_i, t_k)$ of a type $t_k$ for an entity $e_i$ appearing in a document $d_j$ as the criterion to decide whether the entity should be extracted for the domain specific task in a sentence.
Then we use a threshold to cut the entity types that appear less than the frequency. Here we assume that the most frequent type(s) of an entity appearing in the document are the correct semantic meaning(s) in the context.

{\it Document frequency based semantic filter (DFBSF).} Similar to the frequency based method, we use the document frequency $\sum_j I_{d_j}(e_i, t_k)$ of a type $t_k$ of an entity $e_{i}$ as the criterion to find the most likely semantic meaning, where $I_{d_j}(e_i, t_k)=1$ if $e_{i}$ in $d_j$ is with type $t_k$, otherwise $I_{d_j}(e_i, t_k)=0$. Here we assume that if an entity appears in multiple documents with the same type, then the type should be the correct semantic meaning in whole document collection.

{\it Conceptualization based semantic filter (CBSF)}. Motivated by the approaches of conceptualization~\cite{song2011short,song2015short} and entity disambiguation~\cite{li2013mining}, we represent each entity with a feature vector ${\bf t}_i = (t_1, \ldots, t_{\bar{T}})^T$ of entity types, and use standard Kmeans to cluster the entities in a document. Suppose in one cluster we have a set of entities $\mathcal E = \{e_1, \cdots, e_{N_{\mathcal E}}\}$. We then use the probabilistic conceptualization proposed in~\cite{song2011short} to find the most likely entity types for the entities in the cluster. We make the naive Bayes assumption and use
\begin{equation}
P(t_k|\mathcal E)  \propto P(t_k)\prod_{i=1}^{N_{\mathcal E}}P(e_i|t_k)
\end{equation}
as the score of entity type $t_k$. Here, $P(e_i|t_k) = \frac{n(e_i,t_k)}{n(t_k)}$ where $n(e_i,t_k)$ is the co-occurrence count of entity type $t_k$ and entity $e_i$ in the knowledge base, and $n(t_k)$ is the overall number of entities with type $t_k$ in the knowledge base. Besides, $P(t_k) = \frac{n(t_k)}{\sum_k^{\bar{T}} n(t_k)}$.
Note that, in this formulation, we have $n(e_i,t_k)\equiv1$ for Freebase and YAGO, since the evidence in Freebase and YAGO is deterministic.
We also want to leverage the information provided by the document or corpus. Therefore, we can also replace $n(e_i,t_k)$ with the frequency $n_{d_j}(e_i, t_k)$ or document frequency $\sum_j I_{d_j}(e_i, t_k)$ used in the previous two methods.
In this case, we mean we construct a sub-knowledge base with the edges being weighted by the evidence shown in the document or the corpus.
The probability $P(t_k|\mathcal E)$ is used to rank the entity types and the largest ones are selected.
In this case, different entities in a document can be used to disambiguate each other. For each cluster, only the common types are retained, and concepts with conflicts are filtered out. Here we also assume that the type that can best fit the context is the correct semantic meaning. Different from the FBSF method, we also consider the entity cluster information. Therefore, CBSF will use more accurate context information about the entity types. Figure~\ref{fig:sfexp} shows the semantic filtering results based on the example shown in Figure~\ref{fig:spexp}. ``John Smoltz'' is actually a baseball player, and ``Braves'' means the Atlanta Braves, which is a baseball team playing in national league. Note that, based on CBSF, the entity {disambiguation} problem is resolved to some extent. For example, ``Bob Horner'' could help ``John Smoltz'' disambiguate the correct entity type to be {\it baseball\_player} from {\it tv\_actor}, by using the context information provided in the first two documents. Similarly, the context information in the first and the third documents could help to disambiguate the entity type of ``Braves'' to be {\it baseball\_team}.


\section{World Knowledge Representation}
\label{sec:representation}

The output of semantic parsing and semantic filtering is then the document associated with the entities, which are further associated with the types (or concepts, categories, the names can be different for different knowledge bases and relations).
For example, in Freebase, we select the top level named entity {categories} (\ie, domains) as the types, \eg, {\it Person}, {\it Location}, and {\it Organization}.
In addition to the named entities, we also regard the document and word as two types.
Then we use an HIN to represent the data we get after semantic parsing and semantic filtering.

\begin{definition}
\label{def:hin}
A \textbf{heterogeneous information network} (HIN) is a graph ${\mathcal G} = ({\mathcal V}, {\mathcal E})$ with an entity type mapping $\phi$: ${\mathcal V} \to \mathcal A$ and a relation type mapping $\psi$: ${\mathcal E} \to \mathcal R$, where ${\mathcal V}$ denotes the entity set and ${\mathcal E}$ denotes the link set, $\mathcal A$ denotes the entity type set and $\mathcal R$ denotes the relation type set, and the number of entity types $|\mathcal A|>1$ or the number of relation types $|\mathcal R|>1$.
\end{definition}

The network schema provides a high-level description of a given heterogeneous information network.

\begin{definition}
Given an HIN ${\mathcal G} = ({\mathcal V}, {\mathcal E})$ with the entity type mapping $\phi$: ${\mathcal V} \to \mathcal A$ and the relation type mapping $\psi$: $\mathcal E \to \mathcal R$, the \textbf{network schema} for network $G$, denoted as $\mathcal T_{\mathcal G} = (\mathcal A, \mathcal R)$, is a graph with nodes as entity types from $\mathcal A$ and edges as relation types from $\mathcal R$.
\end{definition}

\begin{figure}[t]
\centering
\includegraphics[width=0.5\textwidth]{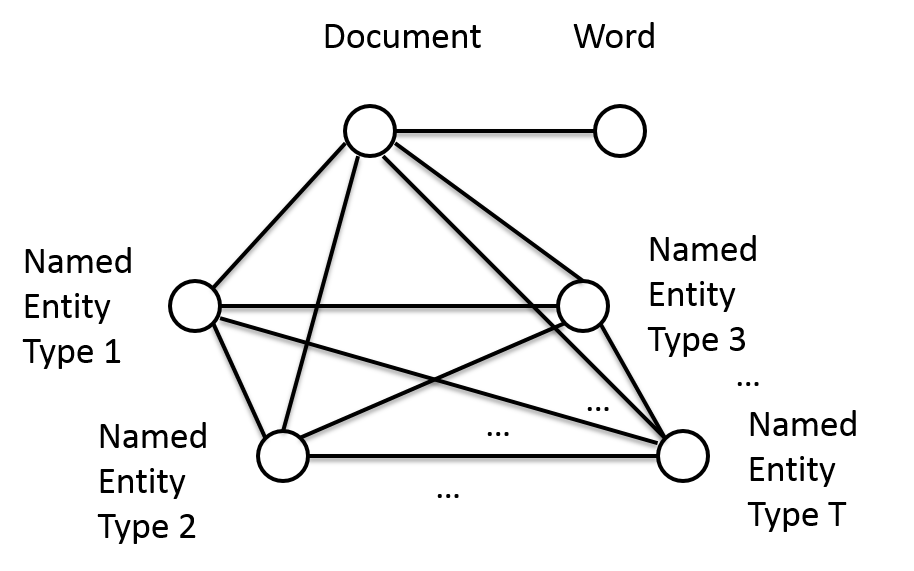}
\vspace{-0.05in}
\caption{Heterogeneous information network schema. The specified knowledge is represented in the form of heterogeneous information network. The schema contains multiple entity types: document $\mathcal D$, word $\mathcal W$, named entities $\{{\rm {\mathcal E^I}}\}_{I=1}^T$, and the relation types connecting the entity types.}
\label{fig:schema}
\end{figure}

Then for our world knowledge dependent network, we use the network schema shown in Figure~\ref{fig:schema} to represent the data.
The network contains multiple entity types: {\bf document} $\mathcal D$, {\bf word} $\mathcal W$, {\bf named entities} $\{{\rm {\mathcal E^I}}\}_{I=1}^T$, and a few relation types connecting the entity types.
Notice that, we use ``entity type'' to represent the node type in HIN, as Definition~\ref{def:hin} showed. We use ``named entity type'' to represent the type of the name mentioned in text (widely used in NLP community), \eg, person, location, and organization names. The entities in HIN do not have to be named entities, \eg, the categories of animals or diseases.
We denote the document set as ${\mathcal D} =
\{d_1, d_2, \ldots, d_{M}\}$, where $M$ is the size of ${\mathcal D}$, the word set as ${\mathcal{W}} =
\{w_1,w_2,\ldots,w_{N}\}$, where $N$ is the size of ${\mathcal{W}}$,
and the entity set as ${\mathcal E^t} = \{e^t_1, e^t_2, \ldots, e^t_{V_t}\}$, where $V_t$ is the size of ${\mathcal E^t}$.
We have $t=1, ..., T$ where $T$ is the total number of named entity types we find in the knowledge base.
Note that if there are no named entities, then the network reduces to a bipartite graph containing only documents and words.

\begin{figure*}[t]
\centering
\includegraphics[width=1.0\textwidth]{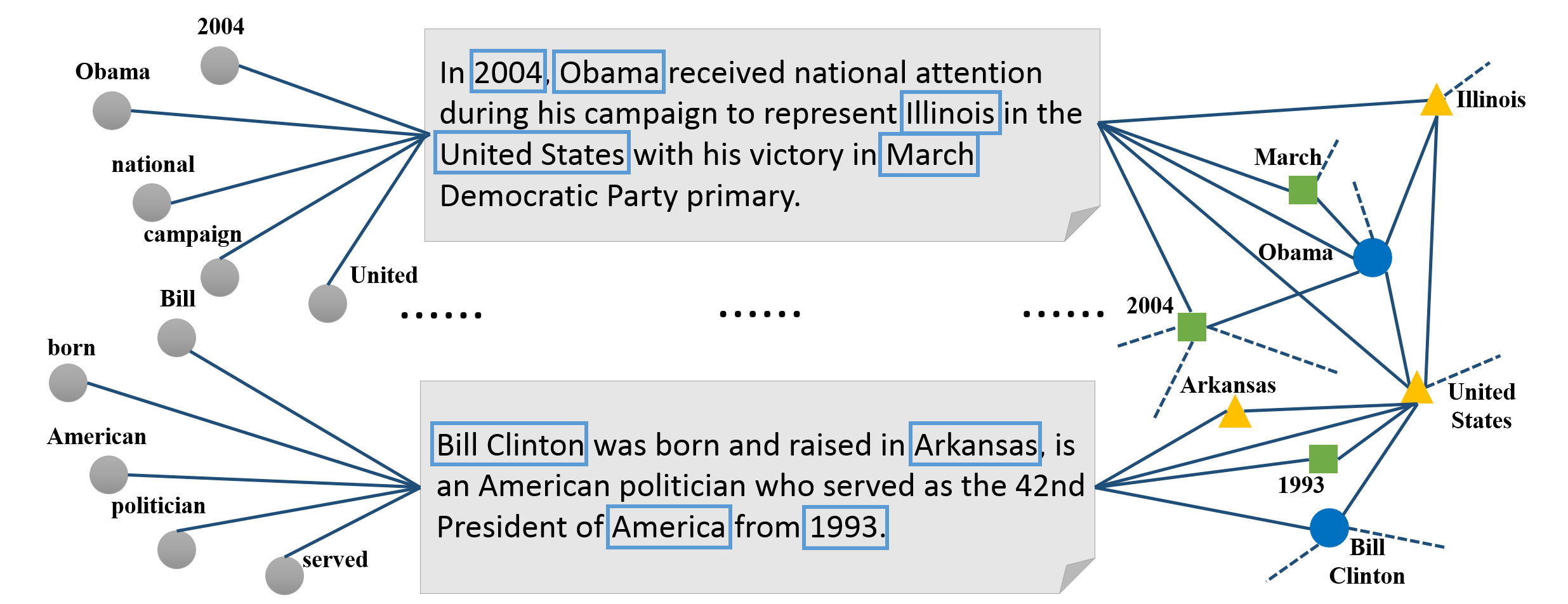}
\caption{Document based heterogeneous information network example. The network $\mathcal G$ contains five entity types: document, word, date, person and location, which are represented with gray rectangle, gray round, green square, blue round, and yellow triangle, respectively.}
\label{fig:dochinex}
\end{figure*}

{In Figure~\ref{fig:dochinex}, we show a real example of the world knowledge dependent network. From the figure, we can see that two documents represented as gray rectangles are modelled in the HIN. Besides the two documents, we have words represented as gray rounds. The link between a word and the corresponding document indicates that the document contains the word. We also have named entities that associated with certain types (\eg, date, person, location), relevant to the documents specified from the knowledge base. The link between a document and a named entity means that the document contains the named entity. The link between two named entities represents the relation between the named entities in the knowledge base generated by world knowledge specification. In summary, after performing world knowledge specification for documents, world knowledge representation aims to construct a document based HIN that contains the link and type information that are useful to understand the documents, thus lead to better text mining performance.}

\section{Document Clustering with World Knowledge}
\label{sec:clustering}
In this subsection, we present our clustering algorithm using HIN, constructed from domain dependent documents and the world knowledge.
Given the HIN, it is natural to perform HIN partitioning to obtain the document clusters.
In addition to the HIN itself, let us revisit the structural information in a typical world knowledge base, e.g., Freebase.
In the world knowledge base, the named entities are often organized in a hierarchy of categories.
Although there are additional category information for each entity, we only use the top level named entity types as the entity types in HIN.
For example, ``Barack Obama'' is a person, where person is the top level category.
In addition, he is the president of the ``United States,'' a politician, a celebrity, \etc.
Another example is that ``Google'' is a software company, plus it has a CEO.
This shows that the entities can have some attributes.
We choose to use top level entity types for the HIN schema since then we will have a relatively dense graph for each pairwise nodes in the network schema.
The fine-grained named entity sub-types or the attributes are also very useful to identify the topics or the clusters of the documents.
Therefore, in this section, we introduce how we incorporate the fine-grained level of named entity types as constraints in the HIN clustering algorithm.

\subsection{Constrained Clustering Modeling}
To formulate the clustering algorithm for the domain dependent documents, we denote latent label sets of the documents as ${\mathcal L}_d = \{l_{d_1},l_{d_2},\ldots,l_{d_{M}}\}$.
We also denote ${\mathcal L}_w = \{l_{w_1},l_{w_2},\ldots,l_{w_{N}}\}$ for words, and ${\mathcal L}_{e^t} = \{l_{e^t_1},l_{e^t_2},\ldots,l_{e^t_{V_t}}\}$ for the $t^{th}$ named entities set.
In general, we follow the framework of information-theoretic co-clustering (ITCC)~\cite{dhillon2003information} and constrained ITCC~\cite{song2013constrained,wang2015constrained} to formulate our approach.
Instead of only performing on the bipartite graph, we need to handle multi-type relational data, as well as more complicated constraints.

The original ITCC uses a variational function to approximate the joint probability of documents and words, which is:
\begin{equation}
q(d_m, w_i)
= p(\hat{d}_{k_d}, \hat{w}_{k_w}) p(d_m|\hat{d}_{k_d}) p(w_i|\hat{w}_{k_w}),
\end{equation}
where $\hat{d}_{k_d}$ and
$\hat{w}_{k_w}$ are cluster indicators to formulate the conditional probability, and $k_d$ and $k_w$ are the corresponding cluster
indices. $q(d_m, w_i)$ is used to approximate $p(d_m, w_i)$ by minimizing the
Kullback-Leibler (KL) divergence:
\begin{equation}\label{Eq_objective_fuction_factor}
\begin{array}{ll}
& D_{KL}( p({\mathcal D},{\mathcal{W}}) ||  q({\mathcal D},{\mathcal{W}}) )\\
= & D_{KL}( p({\mathcal D},{\mathcal{W}}, \hat{\mathcal D}, \hat{\mathcal{W}}) ||  q({\mathcal
D},{\mathcal{W}}, \hat{\mathcal D}, \hat{\mathcal{W}}) )\\
= & \sum_{{k_d}}^{K_d} \sum_{d_m:l_{d_m} = k_d} p(d_m)
D_{KL}(p({\mathcal W} | d_m) || p({\mathcal W} | \hat{d}_{k_d})) \\
= & \sum_{{k_w}}^{K_w} \sum_{w_i:l_{w_i} = k_w} p(w_i)
D_{KL}(p({\mathcal D} | w_i) || p({\mathcal D} | \hat{w}_{k_w})),
\end{array}
\end{equation}
where $\hat{\mathcal D}$ and $\hat{\mathcal{W}}$ are the cluster sets,
$p({\mathcal W} | \hat{d}_{k_d}) $ denotes a multinomial
distribution based on the probabilities
$$p({\mathcal W} | \hat{d}_{k_d}) = (p(w_1|\hat{d}_{k_d}), \ldots, p(w_{N}|\hat{d}_{k_d}))^T,$$
$$p(w_i | \hat{d}_{k_d}) = p(w_i | \hat{w}_{k_w})p(\hat{w}_{k_w} | \hat{d}_{k_d}) \; {\rm and},$$
$$ p(w_i | \hat{w}_{k_w}) = p(w_i) / p( l_{w_i} = \hat{w}_{k_w}).$$
Symmetrically, we have
$$p({\mathcal D} | \hat{w}_{k_w}) = (p(d_1|\hat{w}_{k_w}), \ldots, p(d_{M}|\hat{w}_{k_w}))^T,$$
$$p(d_i | \hat{w}_{k_w}) = p(d_i | \hat{d}_{k_d})p(\hat{d}_{k_d} | \hat{w}_{k_w})\; {\rm and},$$
$$  p(d_i | \hat{d}_{k_d}) = p(d_i) / p( l_{d_i} = \hat{d}_{k_d}).$$
Moreover, $p(\hat{w}_{k_w} | \hat{d}_{k_d})$ and $p(\hat{d}_{k_d} | \hat{w}_{k_w})$ are computed based on the joint probability $q(\hat{d}_{k_d}, \hat{w}_{k_w}) = \sum_{l_{d_m} = k_d}\sum_{l_{w_i} = k_w} p(d_m, w_i)$.

Motivated by ITCC, according to the network schema shown in Figure~\ref{fig:schema}, our problem of HIN clustering is formulated as
\begin{equation}\label{eq_costorg}
\begin{array}{ll}
{\mathcal J}_{{\rm HINC}} &= D_{KL}( p({\mathcal D},{\mathcal{W}) ||  q({\mathcal D},{\mathcal{W}}}) )  \\
& + \sum_{t=1}^T D_{KL}( p({\mathcal D},{\mathcal{E}^t}) ||  q({\mathcal D},{\mathcal{E}^t}) ) \\
& + \sum_{t=1}^T\sum_{s=1}^T D_{KL}( p({\mathcal{E}^t},{\mathcal{E}^s}) ||  q({\mathcal{E}^t},{\mathcal{E}^s}) ),
\end{array}
\end{equation}
{where the equation aims to leverage the link and type information (documents, words, types of named entities) in the HIN shown in Figure~\ref{fig:schema} to estimate the cluster label of each document. The first part of the problem, $D_{KL}( p({\mathcal D},{\mathcal{W}) ||  q({\mathcal D},{\mathcal{W}}}) )$, according to Eq.~\ref{Eq_objective_fuction_factor}, is exactly the original ITCC on the document-word bipartite graph (the top portion of the Figure~\ref{fig:schema} with two types of entities, document and word). It aims to use a variational function $q({\mathcal D},{\mathcal{W}})$ to approximate the joint probability of documents and words $p({\mathcal D},{\mathcal{W}})$ measured by KL divergence, to minimize the mutual information loss due to the mapping to generate the cluster indices of the documents and words as shown in~\cite{dhillon2003information}'s LEMMA 2.1. Similarly, the second part $\sum_{t=1}^T D_{KL}( p({\mathcal D},{\mathcal{E}^t}) ||  q({\mathcal D},{\mathcal{E}^t}) )$ means to use the variational function $q({\mathcal D},{\mathcal{E}^t})$ to approximate the joint probability of documents and named entities belong to every type $t$. This equals to perform ITCC on every document-named entity of type $t$ bipartite graph (the middle portion of the Figure~\ref{fig:schema}). The third part is also defined similarly, which defines the variational function $q({\mathcal{E}^t},{\mathcal{E}^s})$ to approximate the joint probability of named entities of every type $t$ and named entities belong to every type $s$. This equals to perform ITCC on every named entity of type $t$-named entity of type $s$ bipartite graph (the bottom portion of the Figure~\ref{fig:schema}). All the probabilities of the second and third parts can be similarly defined as the first part, document-word bipartite graph. We omit the detailed definitions for brevity.}
A summary of the notations is shown in Table~\ref{tab:notation}.

\begin{table}[htbp]
\centering
\caption{Notations for clustering algorithm. The indicators are used for the probability representation, while the indices are used as ids for the clusters.}
\begin{tabular}{|c|c|c|c|}
\hline
 Meaning & Document & Word & Named Entity \\ \hline
 Cluster Index & ${k_d}$ &  ${k_w}$ & $k_{e^t}$ \\ \hline
 Cluster Indicator & $\hat{d}_{k_d}$ &  $\hat{w}_{k_w}$ & $\hat{e^t}_{k_{e^t}}$ \\ \hline
 Data Indicator  & $d_m$ & $w_i$ & $e_i^t$ \\ \hline
 Data Indicator Set & ${\mathcal D}$ & ${\mathcal W}$ & ${\mathcal E}^t$ \\ \hline
 Label & $l_{d_m}$ & $l_{w_i}$ & $l_{e^t_i}$ \\ \hline
 Label Indicator Set & ${\mathcal L}_d$ & ${\mathcal L}_w$ & ${\mathcal L}_{e^t}$ \\ \hline
\end{tabular}
\label{tab:notation}
\end{table}

To incorporate the side information of the fine-grained named entity sub-types or the attributes as indirect supervision for document clustering,
we define the constraints for the named entities we find after semantic parsing.
We take the $t^{th}$ entity label set ${\mathcal E}^t$ as an example, and use must-links and cannot-links as the constraints.
We denote the must-link set associated with $e^t_i$ as ${\mathcal M}_{e^t_i}$, and the cannot-link set as ${\mathcal C}_{e^t_i}$.
{The way how we build must-links and cannot-links is described in the experiment (Section~\ref{subsec:const}).}
For must-links, the cost function is defined as
\begin{equation}
\begin{array}{ll}\label{Eq_must_engergy}
&V_{\mathcal M}(e^t_{i_1}, e^t_{i_2}\in{\mathcal M}_{e^t_{i_1}}) \\  = &
w_{\mathcal M} D_{KL}(p({{\mathcal D}} | e^t_{i_1}) || p({{\mathcal D}} | e^t_{i_2})) \cdot {\mathcal I}_{l_{e^t_{i_1}} \neq l_{e^t_{i_2}}},
\end{array}
\end{equation}
where $w_{\mathcal M}$ is the weight for must-links, and $p({{\mathcal D}} | e^t_{i_1})$ denotes a multinomial
distribution based on the probabilities $(p(d_1| e^t_{i_1}), \ldots,
p(d_M | e^t_{i_1}))^T$, and $\mathcal I_{true} = 1$, $\mathcal I_{false}  = 0$.
The above must-link cost function means that if the label of $e^t_{i_1}$ is not equal to the label of $e^t_{i_2}$,
then we should take into account the cost function of how dissimilar the two entities $e^t_{i_1}$ and $e^t_{i_2}$ are.
The dissimilarity is computed based on the probability of document ${\mathcal D}$ given the entities $e^t_{i_1}$ and $e^t_{i_2}$ as Eq.~(\ref{Eq_must_engergy}).
The more dissimilar the two entities are, the larger cost is imposed. {Please refer to the experimental section (Section~\ref{subsec:clus}) for details about the weight setting ($w_{\mathcal M}$) for the must-links.}

For cannot-links, the cost function is defined as
\begin{equation} \label{Eq_cannot_engergy}
\begin{array}{ll}
&V_{\mathcal C}(e^t_{i_1}, e^t_{i_2}\in{\mathcal C}_{e^t_{i_1}}) \\  =
& w_{\mathcal C} (D^t_{max} - D_{KL}(p({{\mathcal D}} | e^t_{i_1}) || p({{\mathcal D}} | e^t_{i_2})) ) \cdot {\mathcal I}_{l_{e^t_{i_1}} \neq l_{e^t_{i_2}}},
\end{array}
\end{equation}
where $w_{\mathcal C}$ is the weight for cannot-links, and $D^t_{max}$ is the maximum value for all
the  $D_{KL}(p({{\mathcal D}} | e^t_{i_1}) || p({{\mathcal D}} | e^t_{i_2}))$.
The cannot-link cost function means that if the label of $e^t_{i_1}$ is equal to the label of $e^t_{i_2}$,
then we should take into account the cost function of how similar they are.
{Also, please refer to the experimental section (Section~\ref{subsec:clus}) for details about how we set $w_{\mathcal C}$ for the cannot-links.}

Integrating the constraints for ${\mathcal L}_{e^1},\ldots,{\mathcal L}_{e^T}$ to Eq.~(\ref{eq_costorg}), the objective function of constrained HIN clustering is:
\begin{equation}\label{Eq_objective_fuction}
\begin{array}{ll}
{\mathcal J}_{{\rm CHINC}} &= D_{KL}( p({\mathcal D},{\mathcal{W}) ||  q({\mathcal D},{\mathcal{W}}}) )  \\
& + \sum_{t=1}^T D_{KL}( p({\mathcal D},{\mathcal{E}^t}) ||  q({\mathcal D},{\mathcal{E}^t}) ) \\
& + \sum_{t=1}^T\sum_{s=1}^T D_{KL}( p({\mathcal{E}^t},{\mathcal{E}^s}) ||  q({\mathcal{E}^t},{\mathcal{E}^s}) ) \\
& + \sum_{t=1}^T  \sum_{e^t_{i_1}=1}^{V_t} \sum_{e^t_{i_2}\in{\mathcal M}_{e^t_{i_1}}}
V_{\mathcal M}(e^t_{i_1}, e^t_{i_2}\in{\mathcal M}_{e^t_{i_1}}) \\
& + \sum_{t=1}^T  \sum_{e^t_{i_1}=1}^{V_t} \sum_{e^t_{i_2}\in{\mathcal C}_{e^t_{i_1}}}
V_{\mathcal C}(e^t_{i_1}, e^t_{i_2}\in{\mathcal C}_{e^t_{i_1}}).
\end{array}
\end{equation}
From this objective function we can see that, the must-links and cannot-links are imposed to the entities that the semantic parsing detects.
Since the task is document clustering, the sub-types of entities serve as indirect supervision because they cannot directly affect the cluster labels of the documents. However, the constraints can affect the labels of entities, and then the labels of entities can be transferred to the document side to affect the labels of documents.

\begin{algorithm}
[tb]
\caption{Alternating Optimization for CHINC.} \label{Alg_triiHMRFITCC}
{
\begin{algorithmic}
\STATE {\bf Input:} HIN defined on
documents ${\mathcal D}$, words ${\mathcal W}$, and entities \\ ${\mathcal E^t}$, $t=1,...,T$;
Set maxIter and  max$\delta$.
\WHILE {iter $<$  maxIter and $\delta > $ max$\delta$}
\STATE {{\bf ${\mathcal D}$ Label Update}: minimize Eq.~(\ref{Eq_document_optimization}) w.r.t. ${\mathcal L}_d$.}
\STATE {{\bf ${\mathcal D}$ Model Update}: update $q(d_m, w_i)$ and $q(d_m, e^t_i)$.}
\vspace{0.05in}
\FOR {$t=1,...,T$}
\STATE {{\bf ${\mathcal E^t}$ Label Update}: minimize Eq.~(\ref{Eq_entity_optimization}) w.r.t. ${\mathcal L}_{e^t}$.}
\STATE {{\bf ${\mathcal E^t}$ Model Update}: update $q(d_m, e^t_i)$ and $q(e^s_j, e^t_i)$.}
\ENDFOR
\vspace{0.05in}
\STATE {{\bf ${\mathcal D}$ Label Update}: minimize Eq.~(\ref{Eq_document_optimization}) w.r.t. ${\mathcal L}_d$.}
\STATE {{\bf ${\mathcal D}$ Model Update}: update $q(d_m, w_i)$ and $q(d_m, e^t_i)$.}
\vspace{0.05in}
\STATE {{\bf ${\mathcal W}$ Label Update}: minimize Eq.~(\ref{Eq_word_optimization}) w.r.t. ${\mathcal L}_{w}$.}
\STATE {{\bf ${\mathcal W}$ Model Update}:  update $q(d_m, w_i)$.}
\vspace{0.05in}
\STATE {Compute cost change $\delta$ using Eq.~(\ref{Eq_objective_fuction}).}
\ENDWHILE
\end{algorithmic}
}
\end{algorithm}

\subsection{Alternating Optimization}

Since global optimization of all the latent labels as well as the
approximate function $q(\cdot, \cdot)$ is intractable, we perform an alternating optimization shown in Algorithm \ref{Alg_triiHMRFITCC}.
We iterate the process to optimize the labels of documents, words, and entities.
Meanwhile, we update the function $q(\cdot, \cdot)$ for the corresponding types.

For example, to find label $l_{d_m}$ of document $d_m$, we have:
\begin{equation}\label{Eq_document_optimization}
\begin{array}{ll}
l_{d_m}  = \arg \mathop {\min }\limits_{l_{d_m} = k_d} &D_{KL}(p({\mathcal W} | d_m) || p({\mathcal W} | \hat{d}_{k_d})) + \\
& \sum_{t=1}^TD_{KL}(p({\mathcal E^t} | d_m) || p({\mathcal E^t} | \hat{d}_{k_d})).
\end{array}
\end{equation}

To find label $l_{w_i}$ of word $w_i$, we have:
\begin{equation}\label{Eq_word_optimization}
\begin{array}{ll}
l_{w_i}  =  \arg \mathop {\min }\limits_{l_{w_i} = k_w} D_{KL}(p({\mathcal D} | w_i) || p({\mathcal D} | \hat{w}_{k_w})).
\end{array}
\end{equation}

To find the label $l_{e^t_{i}}$, we use the iterated conditional mode (ICM) algorithm~\cite{basu2004probabilistic} to iteratively assign a label to the entity. We update one label $l_{e^t_{i}}$ at a time, and keep all the other labels fixed:
\begin{equation}\label{Eq_entity_optimization}
\begin{array}{ll}
l_{e^t_{i}}  &= \arg \mathop {\min }\limits_{l_{e^t_{i}} = k_{e^t}} D_{KL}(p({\mathcal D} | e^t_{i}) || p({\mathcal D} | \hat{e^t}_{k_{e^t}})) \\
&+ \sum_{s=1}^T D_{KL}(p({\mathcal E}^s | e^t_{i}) || p({\mathcal E}^s | \hat{e^t}_{k_{e^t}}))
\\ & + \sum\nolimits_{
{
\begin{array}{l}
e^t_{i'}\in {\mathcal M}_{e^t_{i}}; \\ I_{l_{e^t_{i}} \neq l_{e^t_{i'}}}
\end{array}} }
w_{\mathcal M} D_{KL}(p({\mathcal D} | e^t_{i}) || p({\mathcal D} | e^t_{i'}))
\\ & + \sum\nolimits_{
{
\begin{array}{l}
e^t_{i'}\in {\mathcal C}_{e^t_{i}}; \\ I_{l_{e^t_{i}} = l_{e^t_{i'}}}
\end{array}} }
w_{\mathcal C} \left( D^t_{max} - D_{KL}(p({\mathcal D} | e^t_{i}) || p({\mathcal D} | e^t_{i'})) \right).
\end{array}
\end{equation}
To transfer the original objective function~(\ref{Eq_objective_fuction}) to Eq.~(\ref{Eq_entity_optimization}), we should follow Eq.~(\ref{Eq_objective_fuction_factor}) where we replace the document and word notations to the entity notations.
To understand why Eq.~(\ref{Eq_objective_fuction_factor}) holds, we suggest to refer to the original ITCC for detailed derivation~\cite{dhillon2003information}.

Then, with the labels ${\mathcal L}_d$, ${\mathcal L}_{e^t}$ and ${\mathcal L}_{w}$ fixed, we update the model function $q(d_m, w_i)$, $q(d_m, e^t_i)$, and $q(e^s_j, e^t_i)$.
The update of $q$ is not influenced by the must-links and cannot-links.
Thus we can modify them the same as ITCC~\cite{dhillon2003information} and only show the update of $q(d_m, e^t_i)$ here:
\begin{equation} \label{Eq_para_fuction1}
q(\hat{d}_{k_d}, \hat{e^t}_{k_{e^t}}) = \sum_{l_{d_m} = k_d}\sum_{l_{e^t_i}
= k_{e^t}} p(d_m, e^t_i);
\end{equation}
\begin{equation} \label{Eq_para_fuction2}
q(d_m | \hat{d}_{k_d}) = \frac{q(d_m)}{q(l_{d_m} = k_d)} \;\; [q(d_m
| \hat{d}_{k_d})=0 {\rm \; if \;} l_{d_m} \neq k_d];
\end{equation}
\begin{equation} \label{Eq_para_fuction3}
q(e^t_i | \hat{e^t}_{k_{e^t}}) = \frac{q(e^t_i)}{q(l_{e^t_i} = k_{e^t})} \;\; [q(e^t_i
| \hat{e^t}_{k_{e^t}})=0 {\rm \; if \;} l_{e^t_i} \neq k_{e^t}];
\end{equation}
where {\small $q(d_m) = \sum_{e^t_i} p(d_m, e^t_i)$, $q(e^t_i) = \sum_{d_m} p(d_m,
e^t_i)$},  {\small $q(\hat{d}_{k_d}) = \sum_{k_{e^t}} p(\hat{d}_{k_d},
\hat{e^t}_{k_{e^t}})$} and  {\small $q(\hat{e^t}_{k_{e^t}}) = \sum_{k_d} p(\hat{d}_{k_d},
\hat{e^t}_{k_{e^t}})$}.

Algorithm \ref{Alg_triiHMRFITCC} summarizes the main steps in the
procedure. The objective function~(\ref{Eq_objective_fuction}) with our alternating update monotonically decreases to a local optimum.
This is because the ICM algorithm decreases the non-negative
objective function~(\ref{Eq_objective_fuction}) to a local optimum given a fixed $q$ function. Then the
update of $q$ is monotonically decreasing as guaranteed by the theorem
proven in~\cite{song2013constrained}. {Besides, the original proof of the decrease of $q$ is shown by THEOREM 4.1 in~\cite{dhillon2003information}.}

The time complexity of Algorithm~\ref{Alg_triiHMRFITCC} is
$O(n_{{\mathcal D}, {\mathcal W}} \cdot (K_d+K_w) + \sum_{t=1}^T n_{{\mathcal D}, {\mathcal{E}^t}} \cdot (K_d+K_{e^t}) + \sum_{t=1}^T\sum_{s=1}^T (n_{{\mathcal{E}^t}, {\mathcal{E}^s}} + (n_c * iter_{ICM})) \cdot (K_{e^t}+K_{e^s})) \cdot iter_{AO}$,
where $n_{\cdot, \cdot}$ is the total number of non-zero elements in the corresponding co-occurrence matrix,
$n_c$ is the number of constraints, $iter_{ICM}$ is the number of ICM iterations, $K_d, K_w$ and $K_{e^t}$ are the number of document clusters, word clusters and entity clusters of type $t$, and $iter_{AO}$ is the number of the alternating optimization iterations.

{{\sl Discussion:} The major factors contribute to the time complexity of Algorithm~\ref{Alg_triiHMRFITCC} are the numbers of non-zero elements in the corresponding matrices. We have about $20,000$ documents and $60,000$ words in 20NG. From Figure~\ref{fig:expcn}, we find around $20,000$ entities specified from Freebase with $79$ types of entities. In contrary, in 20NG, the number of document clusters, word clusters and named entity clusters are empirically set to be 20, 40 and 79, according to the number of categories of documents, twice the number of document clusters following~\cite{song2013constrained} and the total number of top level named entity types in Freebase, respectively. Compared to the number of non-zero elements in the matrix (\eg, hundreds of thousands), the number of relevant clusters can be ignored. In such situation, if the network schema contains more types of named entities, the running time of the algorithm will not increase that much, compared to the impacts caused by the size of document sets and vocabulary, as well as the named entity sets. There could be lots of ways to improve the performance of machine learning algorithms in large-scale datasets, \eg, distributed computation. However, this beyond the scope of this paper, we leave it for future work.}

\section{Experiments}
\label{sec:exp}

In this section, we show the experimental results to demonstrate the effectiveness and efficiency of our approach on document clustering with world knowledge as indirect supervision.

\subsection{Datasets}
\label{sec:data}
We use the following two benchmark datasets to evaluate domain dependent document clustering.
For both datasets we assume the numbers of document clusters are given.

{\bf 20Newsgroups (20NG):} The 20newsgroups dataset~\cite{lang1995newsweeder} contains about 20,000 newsgroups
documents evenly distributed across 20 newsgroups.\footnote{\url{http://qwone.com/~jason/20Newsgroups/}}
We use all the 20 groups as 20 classes.

{\bf RCV1:} The RCV1 dataset is a dataset containing manually labeled newswire stories from Reuter Ltd~\cite{lewis2004rcv1}. The news documents are categorized with respect to three controlled vocabularies: industries, topics and regions. There are 103 categories including all nodes except for root in the hierarchy. The maximum depth is four, and 82 nodes are leaves. We select top categories MCAT (Markets), CCAT (Corporate/Industrial) and ECAT (Economics) in one portion of the test partition to form three clustering tasks. The three clustering tasks are summarized in Table~\ref{tab:rcv1}. We use the original source of this data, and use the leaf categories in each task as the ground-truth classes.

\begin{table}[htbp]
\centering
\caption{RCV1 dataset statistics. \#(Categories) is the number of all categories; \#(Leaf Categories)  is the number of leaf categories; 
\#(Documents) is the number of documents.}
\begin{tabular}{|c|c|c|c|}
\hline
  & \#(Categories) & \#(Leaf Categories) & \#(Documents)\\
\hline
MCAT & 9 & 7 & 44,033\\
\hline
CCAT & 31 & 26 & 47,494\\
\hline
ECAT & 23 & 18 & 19,813\\
\hline
\end{tabular}
\label{tab:rcv1}
\end{table}

\subsection{World Knowledge Bases}
Then we introduce the knowledge bases we use.

{\bf Freebase:} Freebase\footnote{\url{https://developers.google.com/freebase/}} is a publicly available knowledge base consisting of entities and relations collaboratively collected by its community members. Now, it contains over 2 billions relation expressions between 40 millions entities.  We convert a logical form generated by our unsupervised semantic parser of the world knowledge specification approach introduced in Section~\ref{sec:framework} into a SPARQL query and execute it on our copy of Freebase using the Virtuoso engine.

{\bf  YAGO2:} YAGO2\footnote{\url{http://www.mpi-inf.mpg.de/departments/databases-and-information-systems/research/yago-naga/yago/}} is also a semantic knowledge base, derived from Wikipedia, WordNet and GeoNames. Currently, YAGO2 has knowledge of more than 10 million entities (like persons, organizations, cities, \etc) and contains more than 120 million facts about these entities. Similar to Freebase, we also convert a logical form into a SPARQL query and execute it on our copy of YAGO2 using the Virtuoso engine.

In Table~\ref{tab:kbs}, we show some statistics about Freebase and YAGO2.

\begin{table}[htbp]
\centering
\caption{Statistics of Freebase and YAGO2. \#(Entity Types) is the number of entity types; \#(Entity Instances)  is the number of entity instances; \#(Relation Types) is the number of relation types; \#(Relation Instances) is the number of relation instances.}\label{tab:kbs}
\vspace{-0.05in}
\begin{threeparttable}
\begin{tabular}{|c|c|c|}
\hline
 Name & Freebase & YAGO2 \\ \hline
 \#(Entity Types) & 1,500$^a$ & 350,000 \\ \hline
 \#(Entity Instances) & 40 millions & 10 millions \\ \hline
 \#(Relation Types) & 35,000 & 100 \\ \hline
 \#(Relation Instances) & 2 billions & 120 millions  \\ \hline
\end{tabular}
\small
\begin{tablenotes}
\item[$^a$] The number of 1,500 types is reported in \cite{dong2014knowledge}. In our downloaded dump of Freebase, we found 79 domains,  2,232 types, and 6,635 properties.
\end{tablenotes}
\end{threeparttable}
\label{tab:kbstats}
\end{table}

\begin{figure}[h]
\centering
\includegraphics[width=0.6\textwidth]{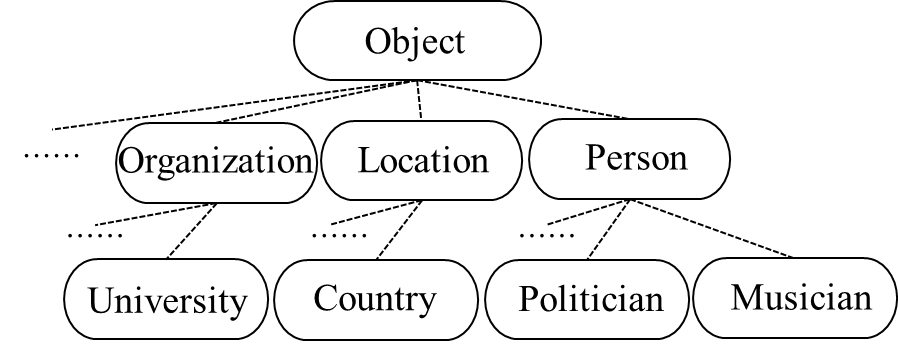}
\caption{Hierarchy of entity types.}
\label{fig:concepthierarchy}
\end{figure}

Note that in most knowledge bases, such as Freebase and YAGO2, entities types are often organized in a hierarchical manner.
For example, {\it Politician} is a sub-type of {\it Person}. {\it University} is a sub-type of {\it Organization}.  All the types or attributes share a common root, called {\it Object}. Figure~\ref{fig:concepthierarchy} depicts an example of hierarchy of types. In general, we use the highest level under the root object as the entity types (\eg, {\it Person}) as specified world knowledge incorporated in the HIN, and the direct children (\eg, {\it Politician}) as entity constraints. In the following experiments, we select {\it Person}, {\it Organization}, and {\it Location} as the three entity types in the HIN, because they are popular in both Freebase and YAGO2.

{{\sl Discussion:} In all the following experiments, we conduct experiments based on different combinations of world knowledge bases (\ie, Freebase, YAGO2) introduced in this section with the four datasets (\ie, 20NG, MCAT, CCAT and ECAT) described in Section~\ref{sec:data}. In total, we illustrate eight different world knowledge base and document dataset combinations for both world knowledge specification and the clustering algorithms. For example, we show eight clustering results of our clustering algorithm named CHINC in Table~\ref{tab:clus_result}(b). As shown in the results, for instance, the clustering result of CHINC on 20NG dataset using Freebase as world knowledge source is 0.631.}

\subsection{Effectiveness of World Knowledge Specification}
Before applying the specified world knowledge to downstream text analytics tasks, such as document clustering in our case, we need to evaluate whether our world knowledge specification approach could produce the correct specified world knowledge.

In order to test the effectiveness of our world knowledge specification approach, we first sample 200 documents from 20 newsgroups, i.e., 10 documents from each category. Second, we split the documents into sentences. After post-processing, 3,232 sentences are generated for human evaluation. Third, we use our world knowledge specification approach in Section~\ref{sec:framework} with three different semantic filtering modules to generate the specified world knowledge for each sentence, which consists of relation triplets in the form of $(e_1, r, e_2)$ with the type information.
{Notice that we use Freebase as the world knowledge source. For FBSF, in each document, we decide the type of an entity in that document by choosing the one with the largest frequency in the document according to Section~\ref{subsec:sf}; For DFBSF, similar to that of FBSF, the type with the largest document frequency is selected as the correct type of the entity in all the documents; For CBSF, we set the number of entity type clusters as 79, which is the number of top types (in Freebase, \ie, domains) as shown in Table~\ref{tab:kbstats}, since we assume the document set including all the possible entity types in the world knowledge base.}
Afterwards, we ask three annotators to label the specified world knowledge according two criterion: (1) whether the boundaries of $e_1$ and $e_2$ are correctly recognized or not; (2) whether the entity type of $e_1$ and $e_2$ are correct or not. It is annotated as correct if both (1) and (2) are satisfied. We check the mutual agreement of the human annotation, which is around $91.3\%$ accuracy.

\begin{table}[htbp]
\centering
\caption{Precision of different semantic filtering results. FBSF represents frequency based semantic filter; DFBSF represents document frequency based semantic filter; CBSF represents conceptualization based semantic filter.}
\begin{tabular}{|c|c|c|c|}
\hline
 Semantic Filter & FBSF & DFBSF & CBSF \\
\hline
Precision & $0.751$ & $0.890$ & $0.916$ \\
\hline
\end{tabular}
\label{tab:sppre}
\end{table}

\begin{table*}[htbp]
\centering
\caption{Error analysis of specified world knowledge generated by the world knowledge specification approach with three different semantic filters. FBSF represents frequency based semantic filter; DFBSF represents document frequency based semantic filter; CBSF represents conceptualization based semantic filter.}
\begin{tabular}{|P{2cm}P{4cm}C{2cm}C{2cm}C{2cm}|}
\hline
Type of error & Example sentence & \multicolumn{3}{c|} {Number and percentage of errors} \\
& & FBSF (805) & DFBSF (359) & CBSF (272) \\
\hline
Entity \newline Recognition & ``Einstein 's theory of relativity explained mercury 's motion.'' & 179 (22.2\%) & 129 (35.9\%) & 105 (38.6\%) \\
\hline
Entity \newline Disambiguation & ``Bill said all this to make the point that Christianity is eminently.'' & 537 (66.7\%) & 182 (50.7\%) & 130 (47.8\%) \\
\hline
Subordinate Clause & ``Bruce S. Winters, worked at United States Technologies Research Center, bought a Ford.'' & 89 (11.1\%) & 48 (13.4\%) & 37 (13.6\%) \\
\hline
\end{tabular}
\label{tab:error}
\end{table*}

\begin{table*}[htbp]
\centering
\caption{Performance of different clustering algorithms on 20NG and RCV1 data. CHINC is our proposed method. BOW, FB (Freebase), or YG (YAGO2) represent bag of word features, the entities generated by our world knowledge specification approach based on Freebase or YAGO2, respectively. We compared all the numbers of HINC and CHINC with CITCC, which is the strongest baseline. The percentage in the brackets are the relative number compared to CITCC. CITCC uses 250K constraints generated based on ground-truth labels of documents.}
\subfloat[{Performance of Kmeans, ITCC, and CITCC with different features.}]
{
\begin{tabular}{|c|c|c|c|c|c|c|c|}
\hline
  & \multicolumn{3}{|c|}{Kmeans} & \multicolumn{3}{|c|}{ITCC} & CITCC \\
 \hline
\underline{Features}  & BOW & BOW & BOW & BOW & BOW & BOW & BOW \\
Data  &  & +FB & +YG & & +FB & +YG & \\
\hline
 20NG & 0.429 & 0.447 & 0.437 & 0.501 & 0.525 & 0.513 & 0.569 \\
\hline
MCAT & 0.549 & 0.575 & 0.559 & 0.604 & 0.630 & 0.619 & 0.652 \\
\hline
CCAT & 0.403 & 0.419 & 0.410 & 0.472 & 0.494 & 0.481 & 0.535 \\
\hline
ECAT & 0.417 & 0.436 & 0.424 & 0.493 & 0.516 & 0.505 & 0.562 \\
\hline
\end{tabular}
}
\\
\subfloat[{Performance of HINC and CHINC with different world knowledge sources.}]
{
\begin{tabular}{|c|c|c|c|c|}
\hline
  & \multicolumn{2}{|c|}{HINC} & \multicolumn{2}{|c|}{CHINC} \\
 \hline
\underline{Features}  & FB & YG & FB & YG \\
Data & & & & \\
\hline
 20NG & 0.571 $\mathbf{(+0.4\%)}$ & 0.541 $\mathbf{(-4.9\%)}$ & 0.631 $\mathbf{(+10.9\%)}$ & 0.600 $\mathbf{(+5.5\%)}$\\
\hline
MCAT & 0.645 $\mathbf{(-1.1\%)}$ & 0.625 $\mathbf{(-4.1\%)}$ & 0.698 $\mathbf{(+7.1\%)}$ & 0.685 $\mathbf{(+5.1\%)}$\\
\hline
CCAT & 0.542 $\mathbf{(+1.3\%)}$ & 0.515 $\mathbf{(-3.7\%)}$ & 0.606 $\mathbf{(+13.3\%)}$ & 0.574 $\mathbf{(+7.3\%)}$\\
\hline
ECAT & 0.561 $\mathbf{(-0.2\%)}$ & 0.530 $\mathbf{(-5.7\%)}$ & 0.624 $\mathbf{(+11.0\%)}$ & 0.588 $\mathbf{(+4.6\%)}$ \\
\hline
\end{tabular}
}
\label{tab:clus_result}
\end{table*}

We then test the precision of three different specified world knowledge generated by the corresponding semantic filtering method. The results are shown in Table~\ref{tab:sppre}. From the results we can see that, CBSF outpeforms the other two ways to generate the correct semantic meaning. The main reason is that, conceptualization based method is able to use the context information to help judge the real semantic of the text rather than only taking the statistics of the data into account.
Here we only care about precision because we wish to use world knowledge as indirect supervision. The recall will not be very important.

\nop{
\begin{table}[htbp]
\centering
\caption{Precision of different semantic filtering results. FBSF represent frequency based semantic parser; DFBSF represents inverted document frequency based semantic parser; CBSF represents conceptualization based semantic parser.}
\vspace{-0.1in}
\begin{tabular}{|c|c|c|c|c|}
\hline
 Semantic Filter & Precision \\
\hline
FBSF & $0.751$ \\
\hline
DFBSF & $0.890$ \\
\hline
CBSF & $0.916$\\
\hline
\end{tabular}
\label{tab:sppre}
\end{table}
}

\subsubsection*{Error Analysis}
To further investigate what triggers the errors in our semantic parsing and semantic filtering pipelines, we analyze the cause of errors for the incorrect specified world knowledge. {The errors are collected from the error cases based on the annotation generated by the three annotators. Then we ask the annotators to try to classify the errors. Finally, we summarize the following three error categories as shown in Table~\ref{tab:error}.}

{\bf Entity Recognition:} In semantic parsing, entities can be extracted incorrectly.
Long entities are composed of multiple simple entities. For example, ``Einstein 's theory of relativity'' may be extracted as ``Einstein'' and ``theory of relativity.'' Paraphrasing and misspelling entities cause their textual expressions to deviate from
any knowledge base entries. Idiomatic expressions are incorrectly picked up as entities. Using a larger mapping from text to knowledge base phrases, or paraphrasing techniques will help avoid some errors. However, this is out of the scope of this article.

{\bf Entity Disambiguation:} Selecting an incorrect entity out of multiple matching candidates causes this
error, \eg, ``Bill'' in our example sentence can be ``Bill Clinton'' or ``Bill Gates.'' Primarily due to two reasons: first, entity disambiguation is a tough research problem in NLP community. Second, the type information of relations are not sufficient to futher prune out mismatching entities during semantic filtering process. Notice that, entity disambiguation is the major cause of the errors. By using CBSF, the number of incorrect entities caused by disambiguation can be dramatically reduced.

{\bf Subordinate Clause:} Semantic parsing sometimes produces {wrong relation phrases} in the subordinate clauses. For example, in the example wrong case shown in Table~\ref{tab:error}, it takes the relation phrase ``worked at'' meaning the working place of ``Bruce S. Winters,'' ignores the phrase ``bought,'' which could be more informative for the target clustering domain. This could be resolved by adding more concrete rules in the semantic parsing grammar.

{{\sl Discussion:} Notice that both 20NG and RCV1 have relatively larger number of named entities
specified from knowledge bases, thus we show more significantly improved clustering results. When encountering the text with little named entities, our algorithm would be very similar to
that of the original CITCC, since the schema of the document based HIN constructed in Section~\ref{sec:framework}
and Section~\ref{sec:representation} would tend to be similar as the schema with only two types of entities, i.e., documents and entities. Besides, CITCC only uses the constraints built upon the words and documents to cluster
the documents, while our approach explores the constraints built based on not only documents and
words, but also the named entities from the knowledge base.}

\begin{figure}[htp]
\centering
\includegraphics[scale=0.20]{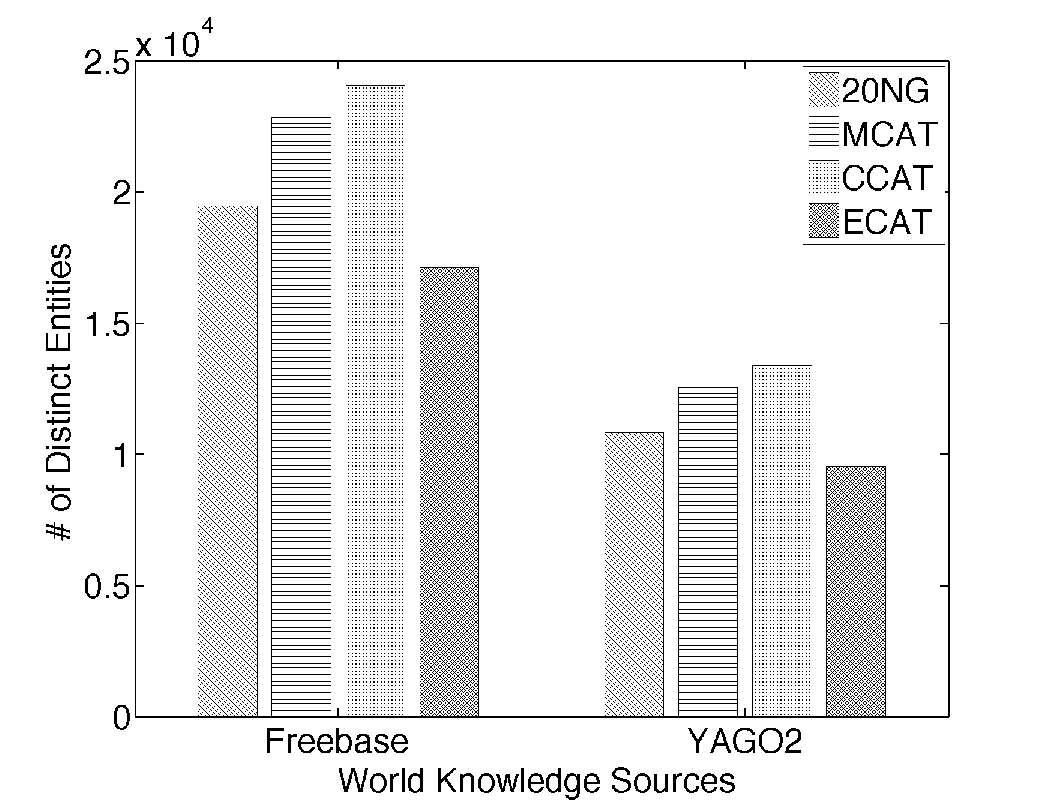}
\caption{Statistics of the number of entities in different document datasets with different world knowledge sources.}
\label{fig:expcn}
\end{figure}

\nop{
The errors are divided into two types according to (i) whether the entities are correctly recognized; (ii) whether the types of entities are correctly recognized. For type (i), the reason for such error is the mistake made by {\bf named entity recognizer} (NER), in the future, a more powerful NER could help improve the performance. For example, given a sentence ``the celebration of Apple inc. is touchful'', our world knowledge specification framework only recognizes the entity as ``Apple inc'' rather than ``celebration of Apple inc.''

For type (ii), we analyzed the sample sentences and find two main causes of the error: (1) Entity disambiguation is a fundamental research problem in natural processing language. In our case, because of the error of NER previously mentioned, the type of ``Apple inc.'' is {\it Organization} rather than include the noun phrases around it as {\it Event}. (2) Our unsupervised semantic parser sometimes find {\bf wrong relation phrase} in the subordinate clauses. For example,  in ``winnipeg is the team that win stanley cup playoffs,'' the parser uses the verb phrase as ``is the team'' when parsing, ignore the right relation phrase ``win'', which leads to the type of ``winnipeg'' to be {\it Person} rather than {\it SportsTeam}.
}

In the following experiments, we use the world knowledge specification approach with CBSF, because it performs the best among the three semantic filtering methods.

\subsection{Clustering Result}
\label{subsec:clus}
\nop{
In Figure~\ref{}, we show the impact of clustering number in CHINC on the clustering performance NMI.
In Figure~\ref{}, we show the impact of iteration number of the alternating optimization algorithm on the clustering performance and execution time.
}

\nop{
According to the different source of document datasets and world knowledge, we use ``Freebase + 20NG'', ``Freebase + RCV1-MCAT'', ``Freebase+RCV1-CCAT'', ``Freebase+RCV1-ECAT'' to denote the clustering task on four document datasets using Freebase as world knowledge source. Similarly, we use ``YAGO2 + 20NG'', ``YAGO2 + RCV1-MCAT'', ``YAGO2+RCV1-CCAT'', ``YAGO2+RCV1-ECAT'' to denote the clustering task on four document datasets using YAGO2 as world knowledge source.
}

\begin{figure*}[htbp]
\centering
\subfloat[{\scriptsize ``CHINC + Freebase'' for 20NG.}]
{
\label{fig:ent_clus_results}
\includegraphics[width=0.5\textwidth,height=0.28\textheight]{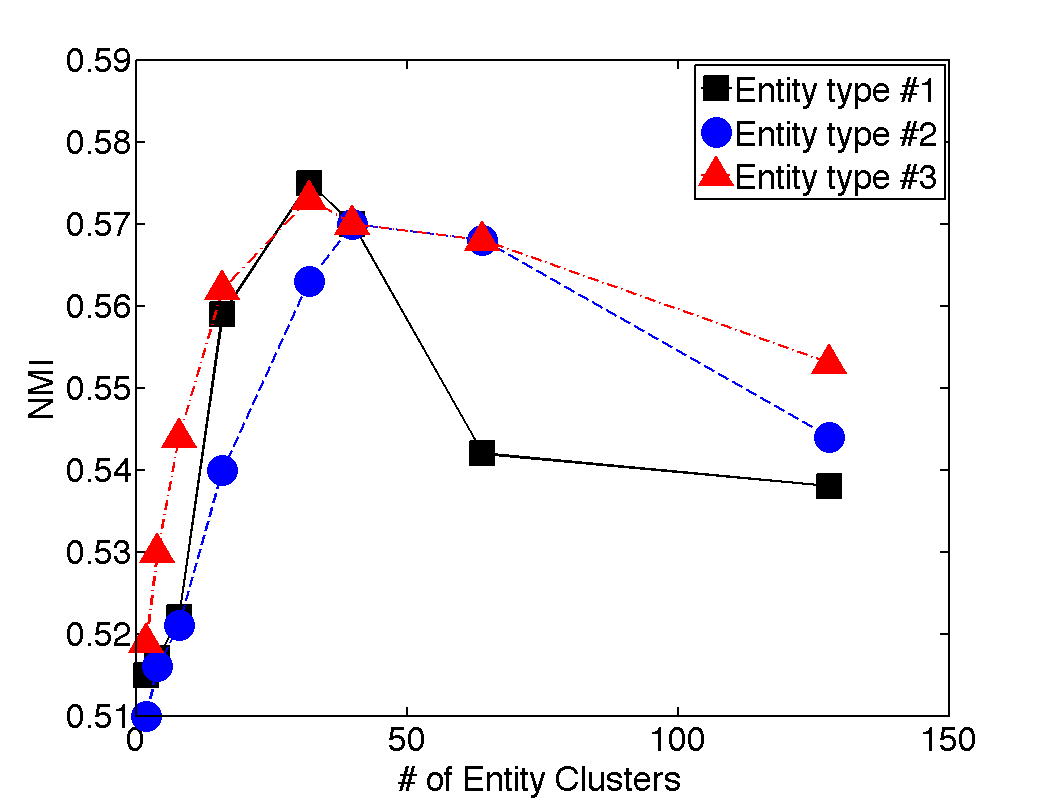}
}
\subfloat[{\scriptsize ``CHINC + Freebase'' for MCAT.}]
{
\label{fig:mcat_clus_results}
\includegraphics[width=0.5\textwidth,height=0.28\textheight]{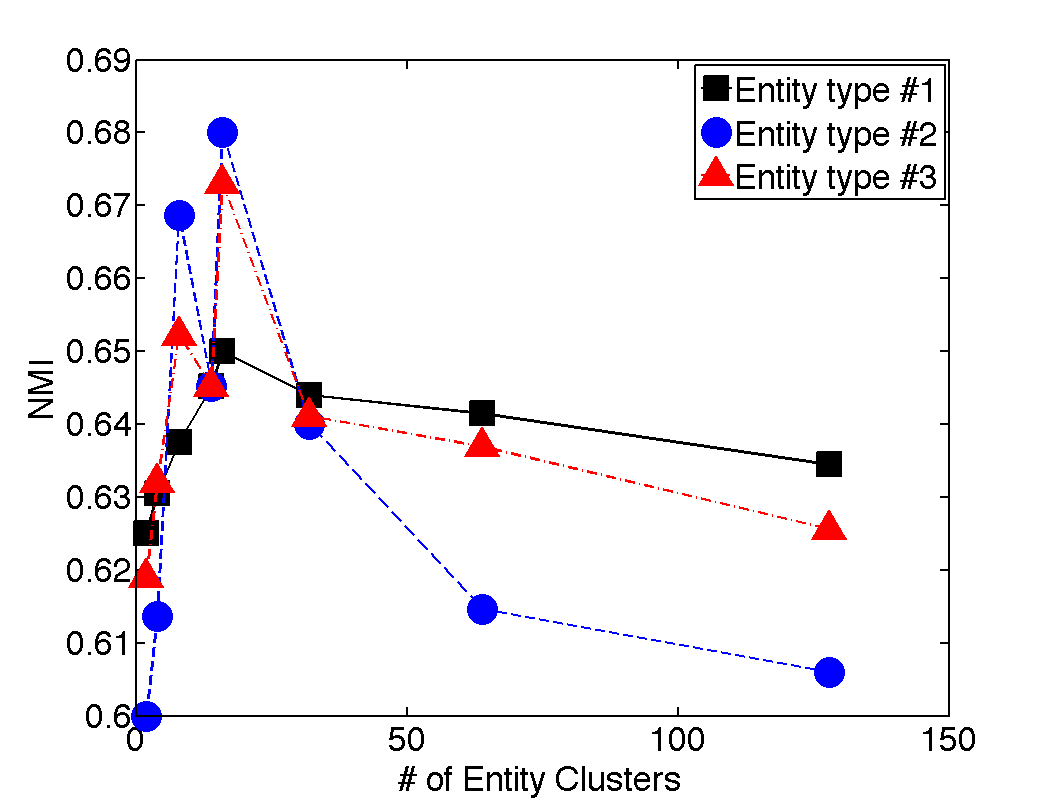}
}
\\
\subfloat[{\scriptsize ``CHINC + Freebase'' for CCAT.}]
{
\label{fig:ccat_clus_results}
\includegraphics[width=0.5\textwidth,height=0.28\textheight]{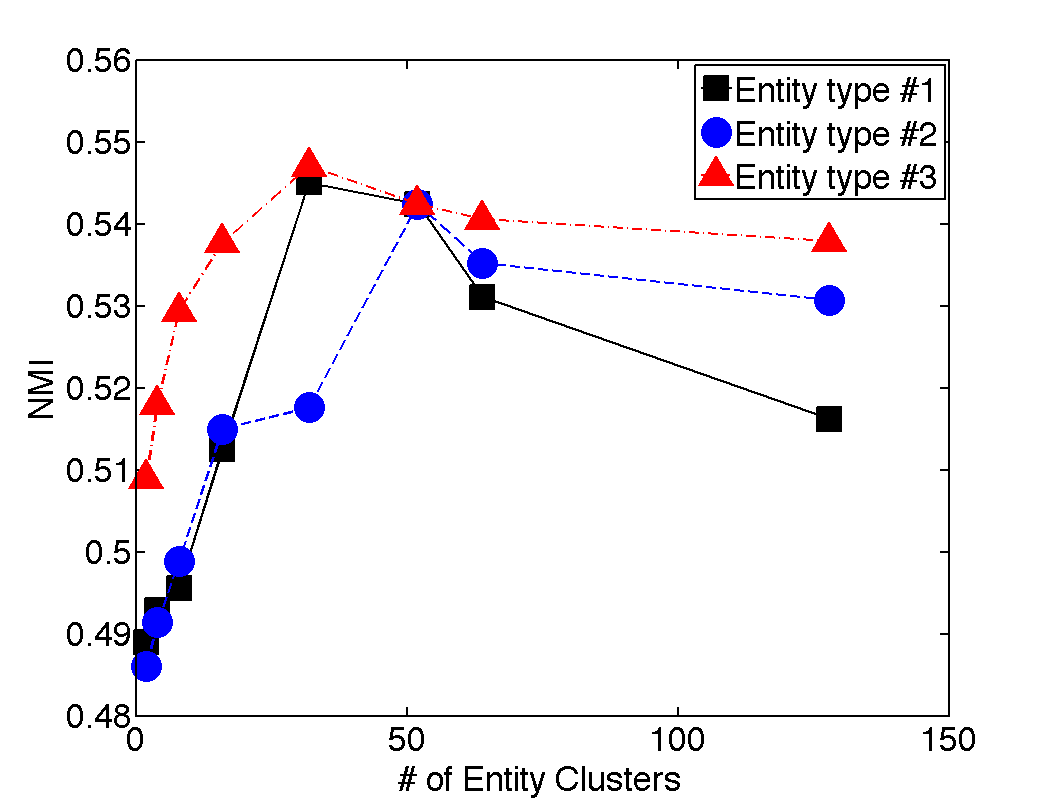}
}
\subfloat[{\scriptsize ``CHINC + Freebase'' for ECAT.}]
{
\label{fig:ecat_clus_results}
\includegraphics[width=0.5\textwidth,height=0.28\textheight]{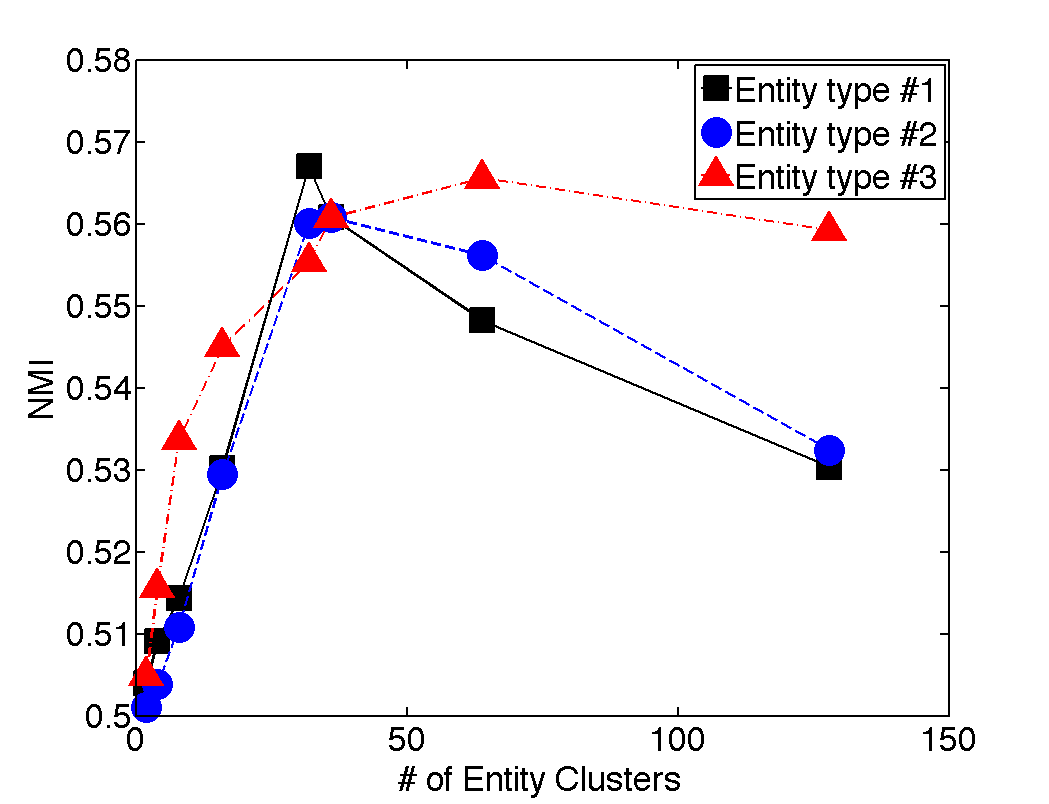}
}
\caption{{Effect of number of entity clusters of each entity type on document clustering on different datasets with Freebase as world knowledge source.}}
\end{figure*}

\begin{figure*}[htbp]
\centering
\subfloat[{\scriptsize ``CHINC + YAGO2'' for 20NG.}]
{
\label{fig:entyg_clus_results}
\includegraphics[width=0.5\textwidth,height=0.28\textheight]{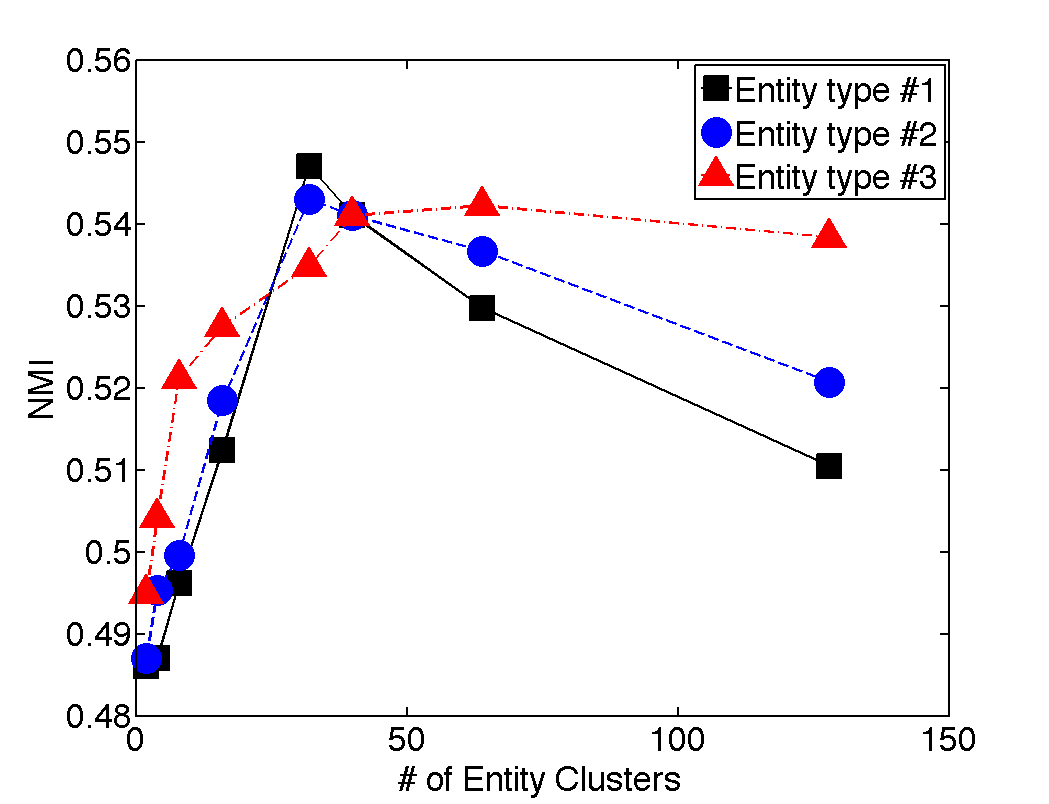}
}
\subfloat[{\scriptsize ``CHINC + YAGO2'' for MCAT.}]
{
\label{fig:mcatyg_clus_results}
\includegraphics[width=0.5\textwidth,height=0.28\textheight]{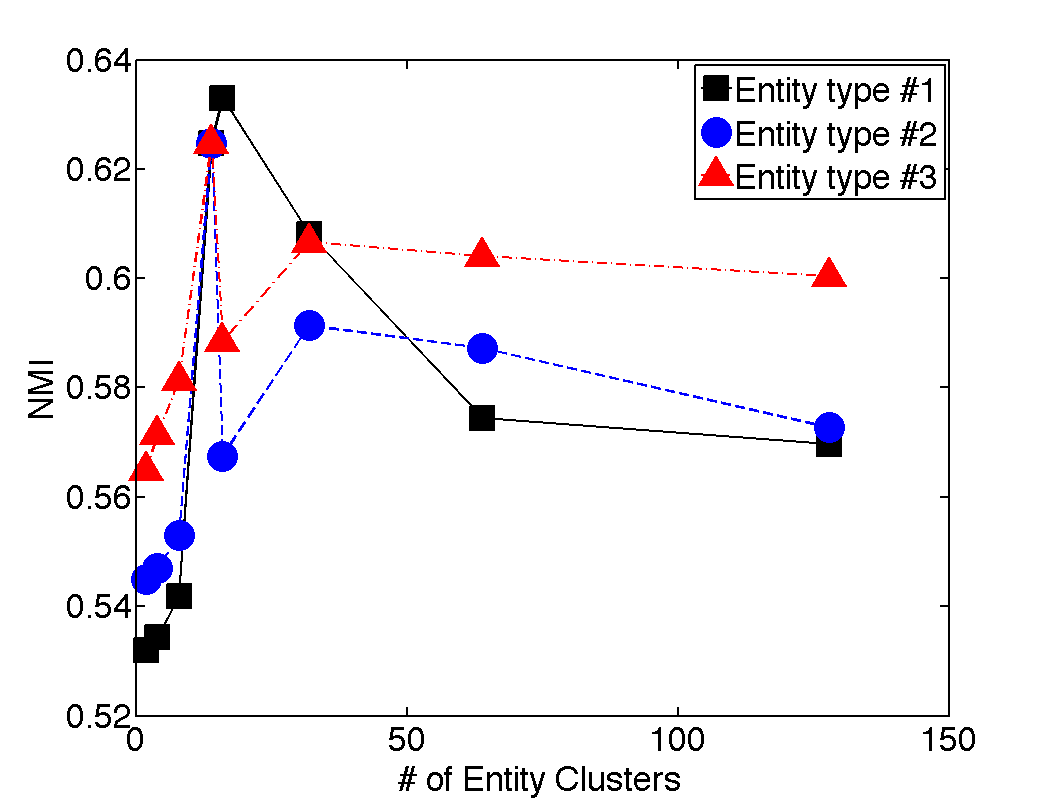}
}
\\
\subfloat[{\scriptsize ``CHINC + YAGO2'' for CCAT.}]
{
\label{fig:ccatyg_clus_results}
\includegraphics[width=0.5\textwidth,height=0.28\textheight]{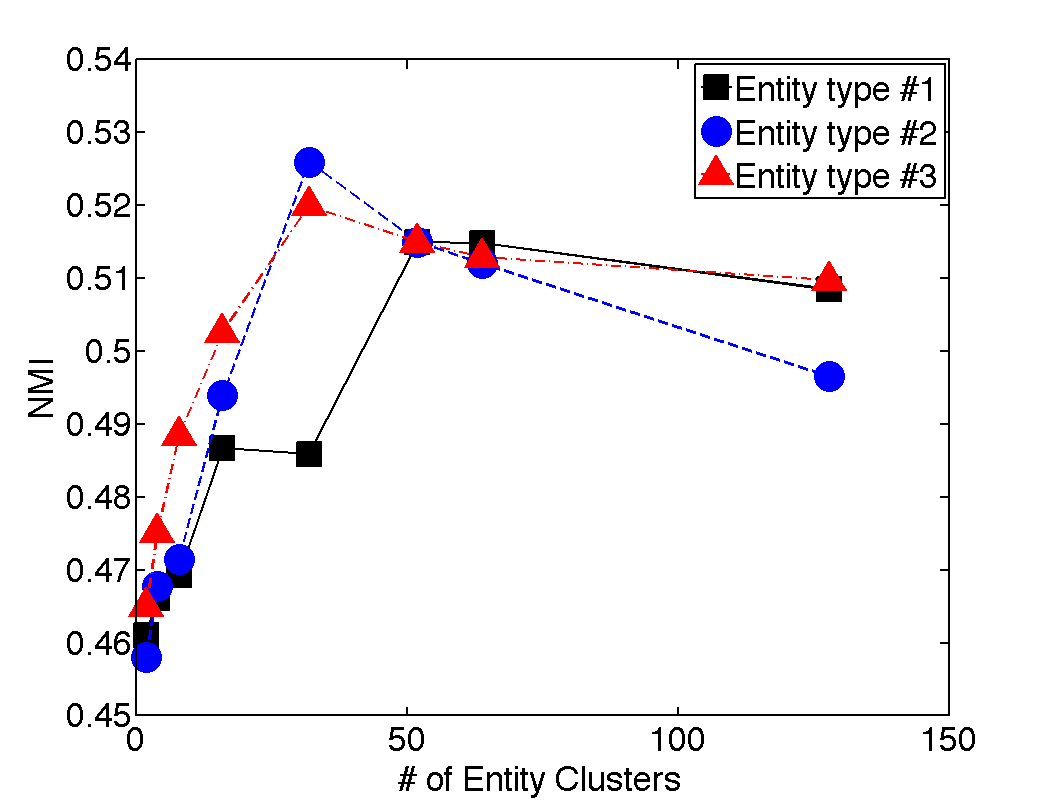}
}
\subfloat[{\scriptsize ``CHINC + YAGO2'' for ECAT.}]
{
\label{fig:ecatyg_clus_results}
\includegraphics[width=0.5\textwidth,height=0.28\textheight]{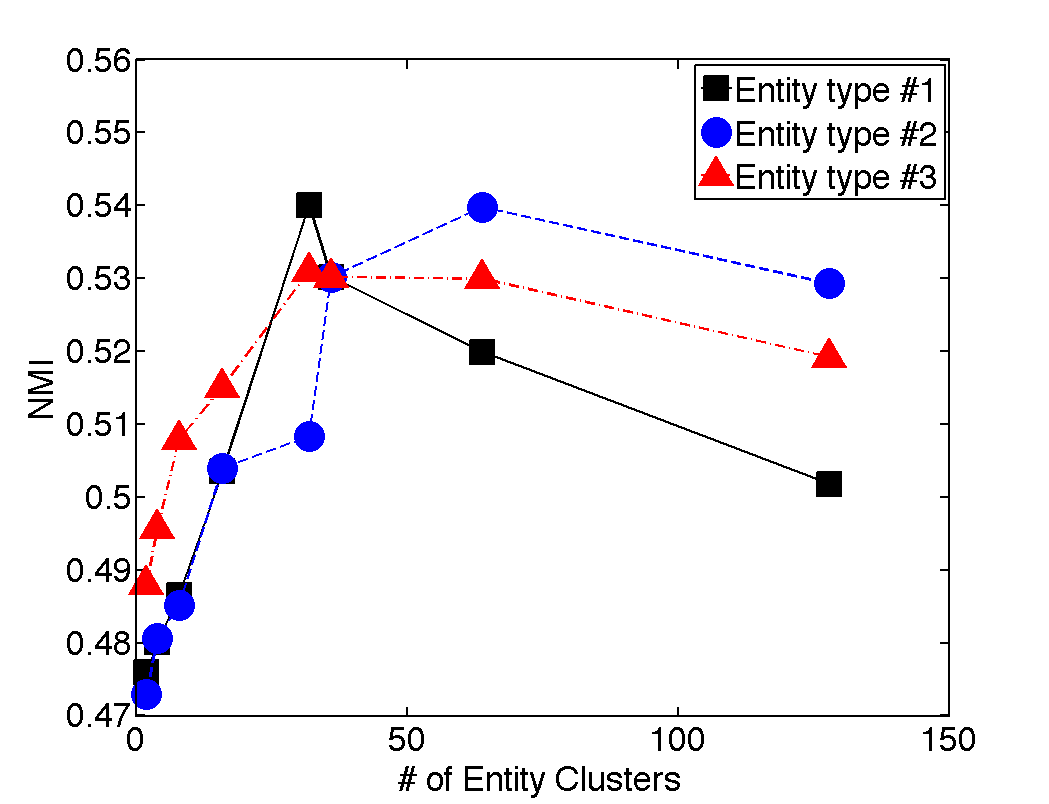}
}
\caption{{Effect of number of entity clusters of each entity type on document clustering on different datasets with YAGO2 as world knowledge source.}}
\end{figure*}

In this experiment, we compare the performance of our model, constrained heterogeneous information network clustering (CHINC), with several representative clustering algorithms such as Kmeans, ITCC~\cite{dhillon2003information} and CITCC~\cite{song2013constrained}. The parameters used in CHINC to control the constraints are $w_{\mathcal M}$ and $w_{\mathcal C}$. {We set them as $\frac{1}{\sum_{t=1}^T|{\mathcal E}^t|}$ following the rules tested in~\cite{song2013constrained}, where $|{\mathcal E}^t|$ is the number of entities in ${\mathcal E}^t$.\footnote{In~\cite{song2013constrained}, the experiment shows the parameter study on the weights of the constraints varying from $1E-8$ to 100, and conclude this is one of the best settings.}}
We also denote our algorithm without constraints as HINC. {For both CHINC and HINC, we set the numbers of document clusters, word clusters and named entity clusters according to the numbers of categories of the document set, twice the number of document clusters following~\cite{song2013constrained} and the total number of top level named entity types in the world knowledge base, respectively. The number of constraints used in each combination of document set and knowledge base are the same, which equals to $3*10^8$. The constraints are randomly selected from all the constraints generated by the method introduced in Section~\ref{subsec:const}.} ``FB'' and ``YG'' represent two different world knowledge sources, Freebase and YAGO2, respectively.
We re-implement all the above clustering algorithms. Notice that, for CITCC, we follow~\cite{song2013constrained} to generate and add constraints for documents and words.
We also use the specified world knowledge as features to enhance the Kmeans and ITCC. The feature settings are defined as below:
\begin{itemize}
\item BOW: Traditional bag-of-words model with the tf-idf weighting mechanism.
\item BOW+FB: BOW integrated with additional features from entities in specified world knowledge of Freebase.
\item BOW+YG: BOW integrated with additional features from entities in specified world knowledge of YAGO2.
\end{itemize}

We employ the widely-used normalized mutual information (NMI)~\cite{strehl2003cluster} as the evaluation measure.
The NMI score is 1 if the clustering results {match the category labels perfectly} and 0 if the clusters are obtained from a random partition.
In general, the larger the scores are, the better the clustering results are.

\nop{Among all the methods we tested, CITCC is constrained clustering algorithms; STriNMF, CSTriNMF, ITCC and CITCC are co-clustering methods; and CSTriNMF and CITCC are constrained co-clustering methods.
}
\nop
{
For document constraints, we added a must-link between two documents if they shared the same category label. We also added a cannot-link if two documents come from different newsgroups. For the word constraints, after stop word removal, we counted the term frequencies of words in each newsgroup, and then chose the top $1,000$ words in each group to randomly generate word pairs to add the word must-links. We did not use any word cannot-links in our experiments. In the following experiments, the number of document clusters is set to 20, the ground-number, the ground-truth number. As suggested in \cite{song2013constrained}, the number of word clusters is fixed to 40, which is as twice as the number of document clusters. Moreover, the trade-off parameters $a_{m_1, m_2}$ and $\bar{a}_{m_1, m_2}$ for constraints in Eqs.~(\ref{Eq_must_engergy}) and (\ref{Eq_cannot_engergy}) are empirically set to $1/\sqrt{M}$ for documents, $1/\sqrt{W}$ for words, and $1/\sqrt{V_I}$ ($I\in \{1, 2, 3\}$) for entities of different types following~\cite{song2013constrained}.
}

In Table~\ref{tab:clus_result}, we show the performance of all the clustering algorithms with different experimental settings. The NMI is the average NMI of five random trials per experiment setting. Overall, among all the methods we test, CHINC consistently performs the best among all the clustering methods.
We can see that HINC+FB and HINC+YG perform better than ITCC with BOW+FB or BOW+YG features, respectively.
This means that by using the structural information provided by the world knowledge, we can further improve the clustering results.
In addition, the algorithms with Freebase consistently outperform the ones with YAGO2, since Freebase has much more facts compared with YAGO2 as shown in Table~\ref{tab:kbs}; besides, one can see in Figure~\ref{fig:expcn} that Freebase could consistently specify more entities than YAGO2 does from all of the document datasets.
CITCC is the strongest baseline clustering algorithm, because it uses the ground-truth constraints derived from category labels based on the human knowledge. We use 250K constraints to perform CITCC.
As shown in Table~\ref{tab:clus_result}, HINC performs competitive with the CITCC.
CHINC significantly outperforms CITCC.
This shows that by automatically using world knowledge, it has the potential to perform better than the algorithm with the specific domain knowledge.

{{\sl Discussion:} Based on the results, we can see that, when performing existing clustering algorithms with world knowledge, it is better to choose the world knowledge sources including relatively larger number of instances of entities and relations. Since, in general, the more entities and relations one knowledge base has, the bigger possibility that more useful entities and relations could be parsed out, and thus impact the final clustering result. Table~\ref{tab:clus_result}  shows such difference between Freebase and YAGO2 on the document clustering task, which marches the difference between the number of specified entities in Figure~\ref{fig:expcn}. Besides, it will be interesting to explore the clustering validation methods~\cite{luo2009information,wu2009adapting,liu2013understanding} to further demonstrate the reliability of the clustering results.}

\subsubsection{Analysis of Number of Entity Clusters}
We also evaluate the effect of varying the number of entity clusters of each entity type in CHINC on the document clustering task.
Figure~\ref{fig:ent_clus_results} shows the results of clustering with different numbers of entity clusters of each entity type on ``CHINC + Freebase'' for the 20NG dataset. The number of entity clusters varies from 2 to 128. The default number of iterations is set as $20$, which will be discussed in Section~\ref{exp:iter}. When testing the effect of the number of entity clusters of one entity type, the numbers of entity clusters of the other two entity types are fixed as twice as the number of document clusters, which are 40 and 40 in 20NG, respectively.
It is shown that for this dataset, more entity clusters may not result in improved document clustering results when a sufficient number of entity clusters is reached. For example, as shown in Figure~\ref{fig:ent_clus_results}, after reaching 32, the NMI scores of CHINC actually decrease when the numbers of entity clusters further increase.
One can also find the effects of the numbers of entity clusters on the clustering performance with the other document dataset and knowledge base combinations in Figures~\ref{fig:mcat_clus_results}--\ref{fig:ecat_clus_results} for Freebase and Figures~\ref{fig:entyg_clus_results}--\ref{fig:ecatyg_clus_results} for YAGO2. From the results, we can conclude that, there exist certain values of the number of entity clusters leading to the best clustering peformance.

\subsubsection{Analysis of Number of Iterations in Alternating Optimization}
\label{exp:iter}
\begin{figure*}[htbp]
\centering
\subfloat[{\scriptsize ``CHINC + Freebase'' for 20NG.}]
{
\label{fig:alg_iter}
\includegraphics[width=0.50\textwidth,height=0.2\textheight]{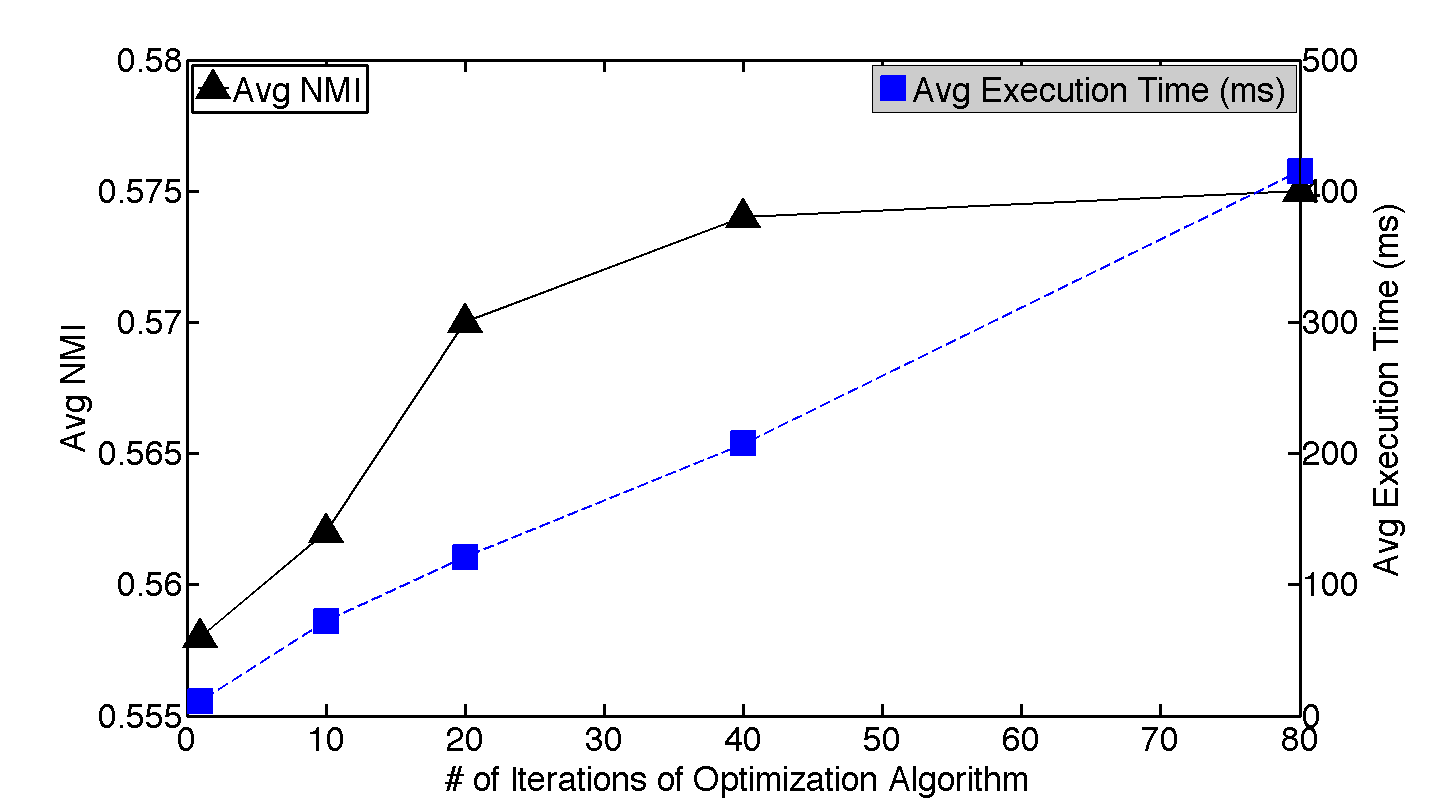}
}
\subfloat[{\scriptsize ``CHINC + Freebase'' for MCAT.}]
{
\label{fig:mcat_clus_iter}
\includegraphics[width=0.50\textwidth]{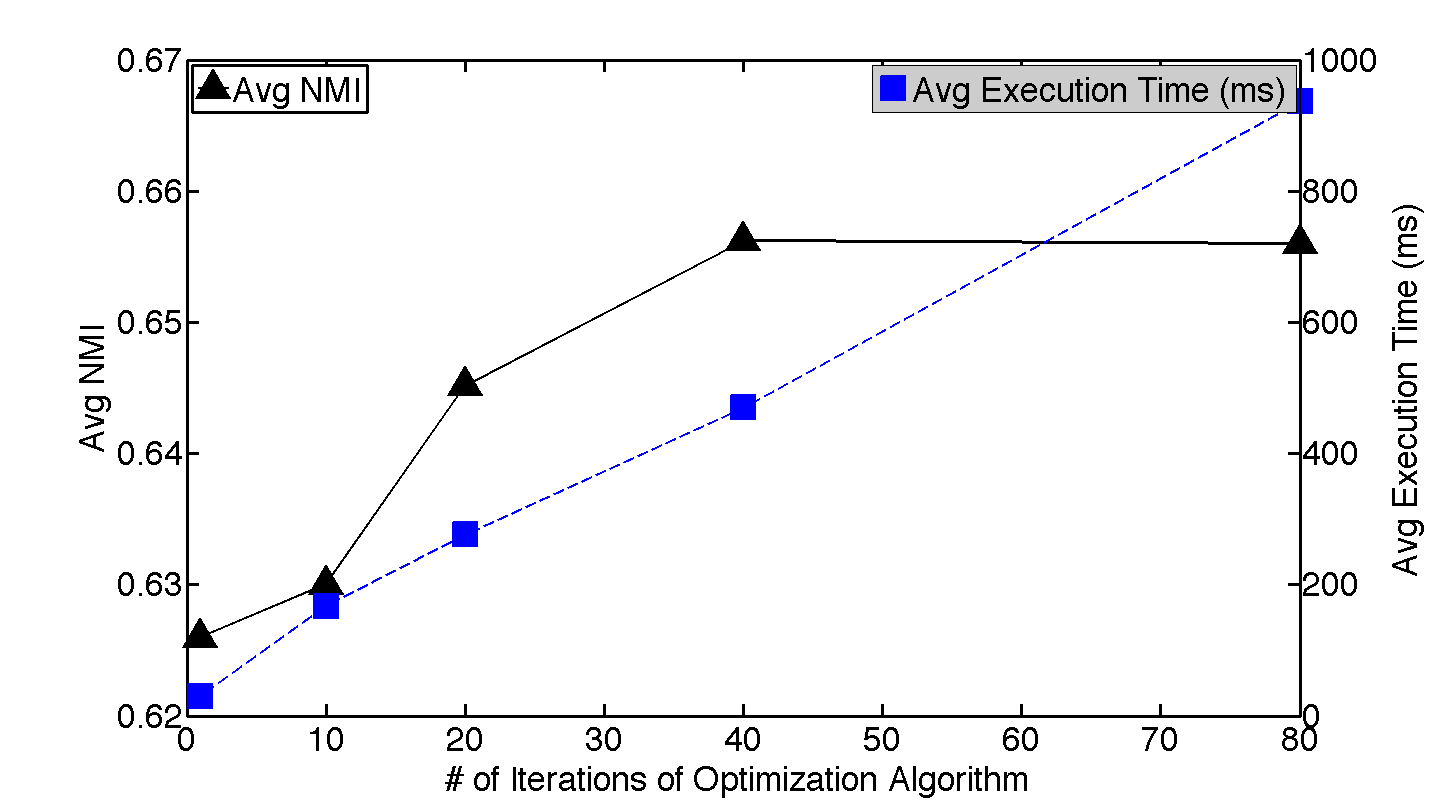}
}
\\
\subfloat[{\scriptsize ``CHINC + Freebase'' for CCAT.}]
{
\label{fig:ccat_clus_iter}
\includegraphics[width=0.50\textwidth]{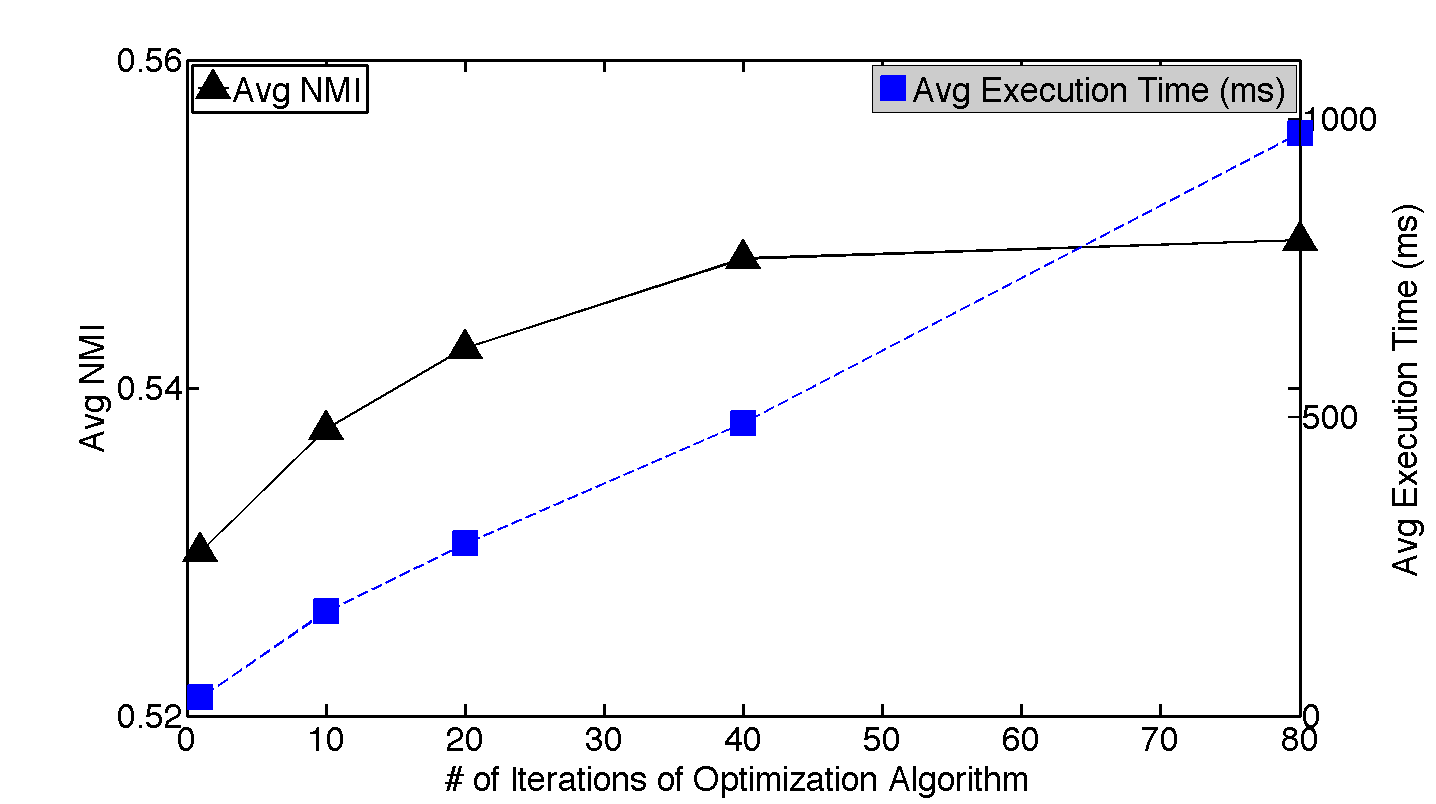}
}
\subfloat[{\scriptsize ``CHINC + Freebase'' for ECAT.}]
{
\label{ecat_clus_iter}
\includegraphics[width=0.50\textwidth]{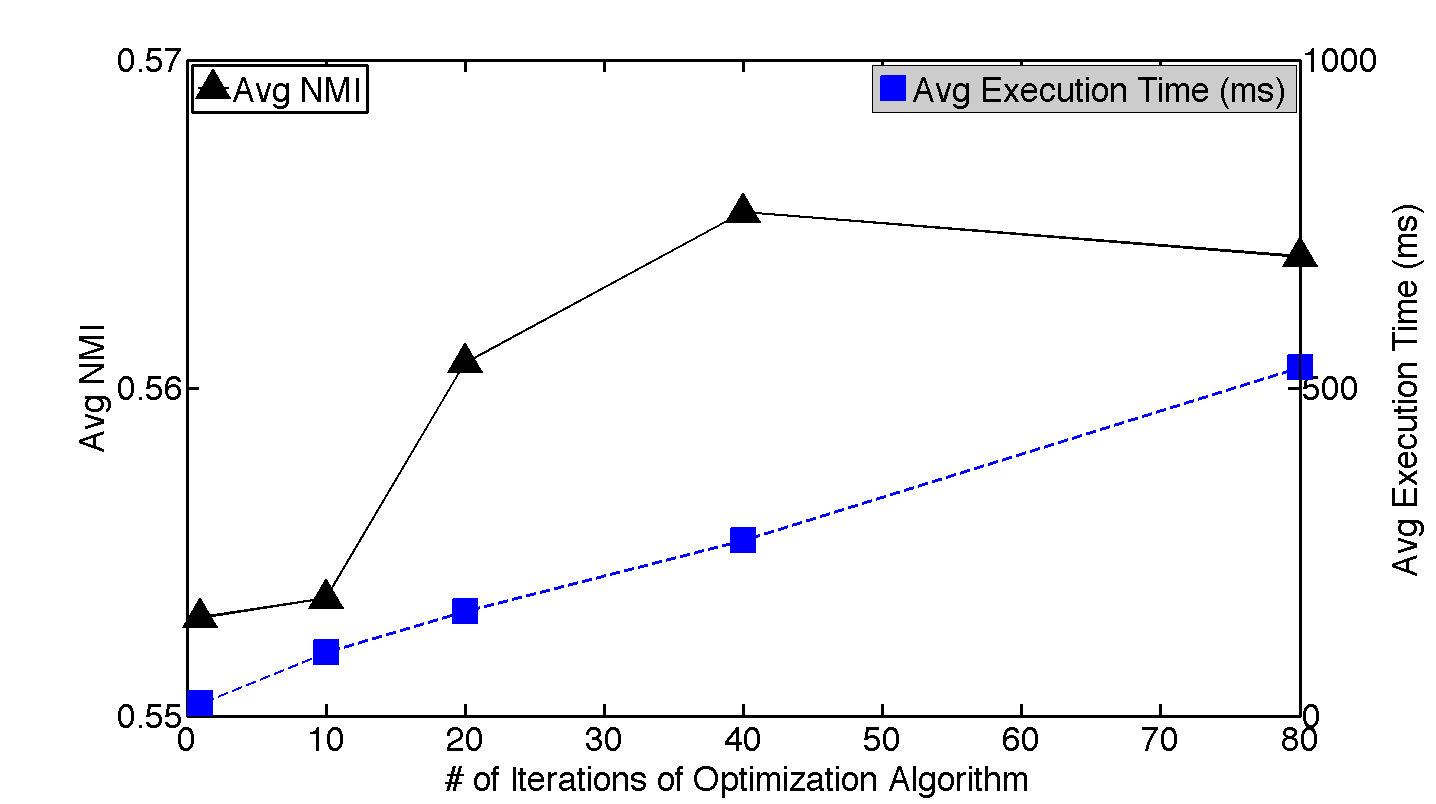}
}
\caption{{Analysis of \# of iterations in alternating optimization algorithm on different datasets with Freebase as world knowledge source. Left $y$-axis: average NMI; Right $y$-axis: average execution time (ms).}}
\label{fig:CHINC_iter}
\end{figure*}

\begin{figure*}[htbp]
\centering
\subfloat[{\scriptsize ``CHINC + YAGO2'' for 20NG.}]
{
\label{fig:algyg_iter}
\includegraphics[width=0.50\textwidth]{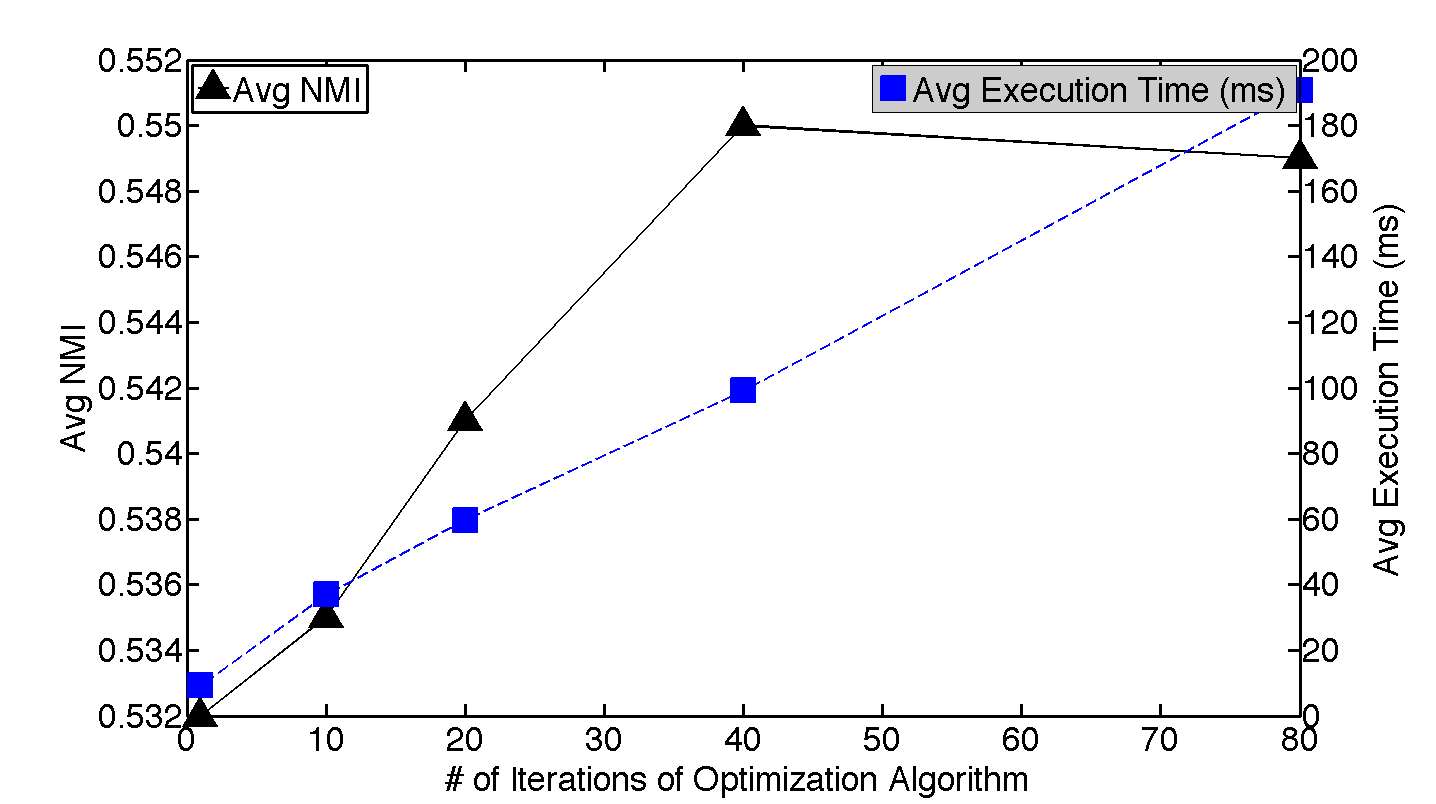}
}
\subfloat[{\scriptsize ``CHINC + YAGO2'' for MCAT.}]
{
\label{fig:mcatyg_clus_iter}
\includegraphics[width=0.50\textwidth]{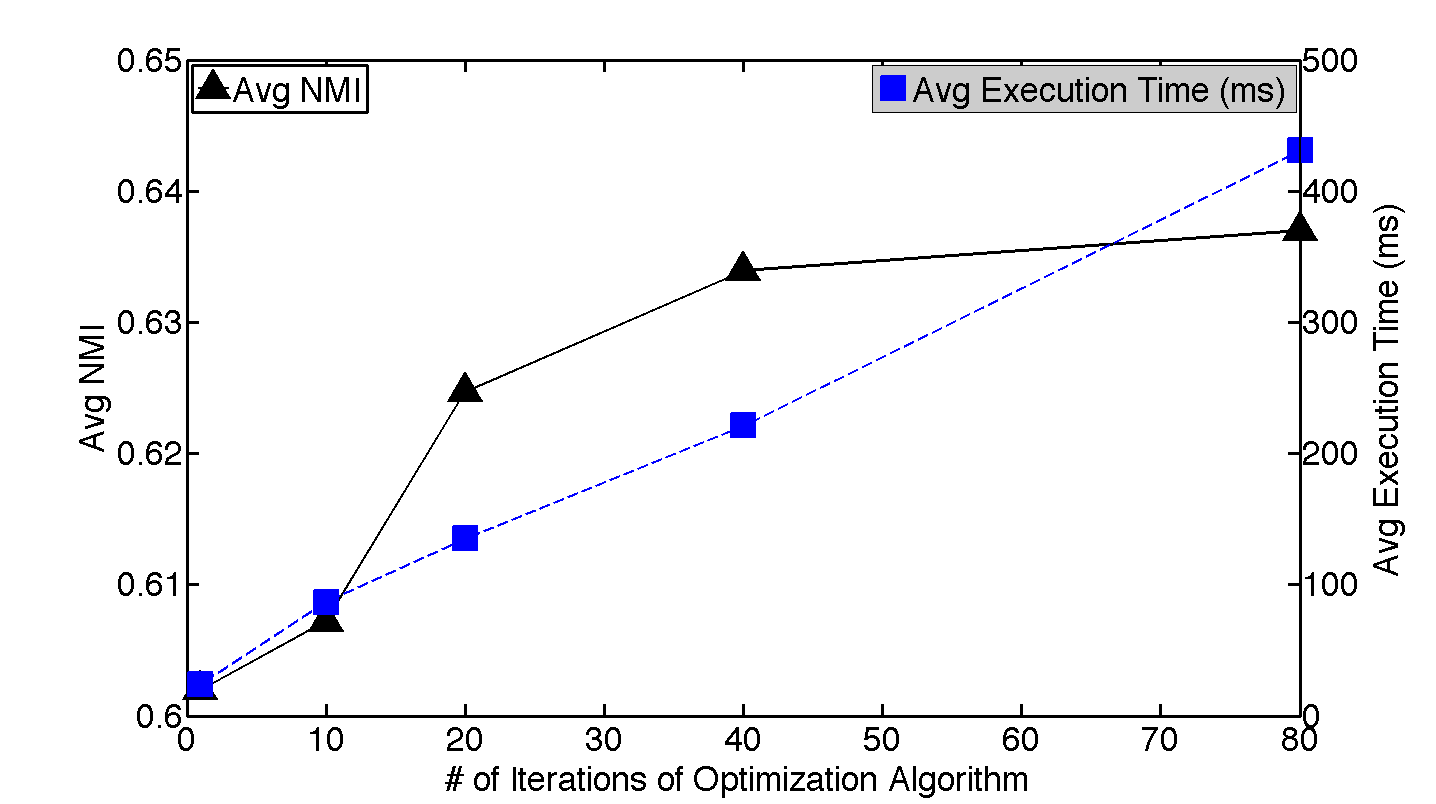}
}
\\
\subfloat[{\scriptsize ``CHINC + YAGO2'' for CCAT.}]
{
\label{fig:ccatyg_clus_iter}
\includegraphics[width=0.50\textwidth]{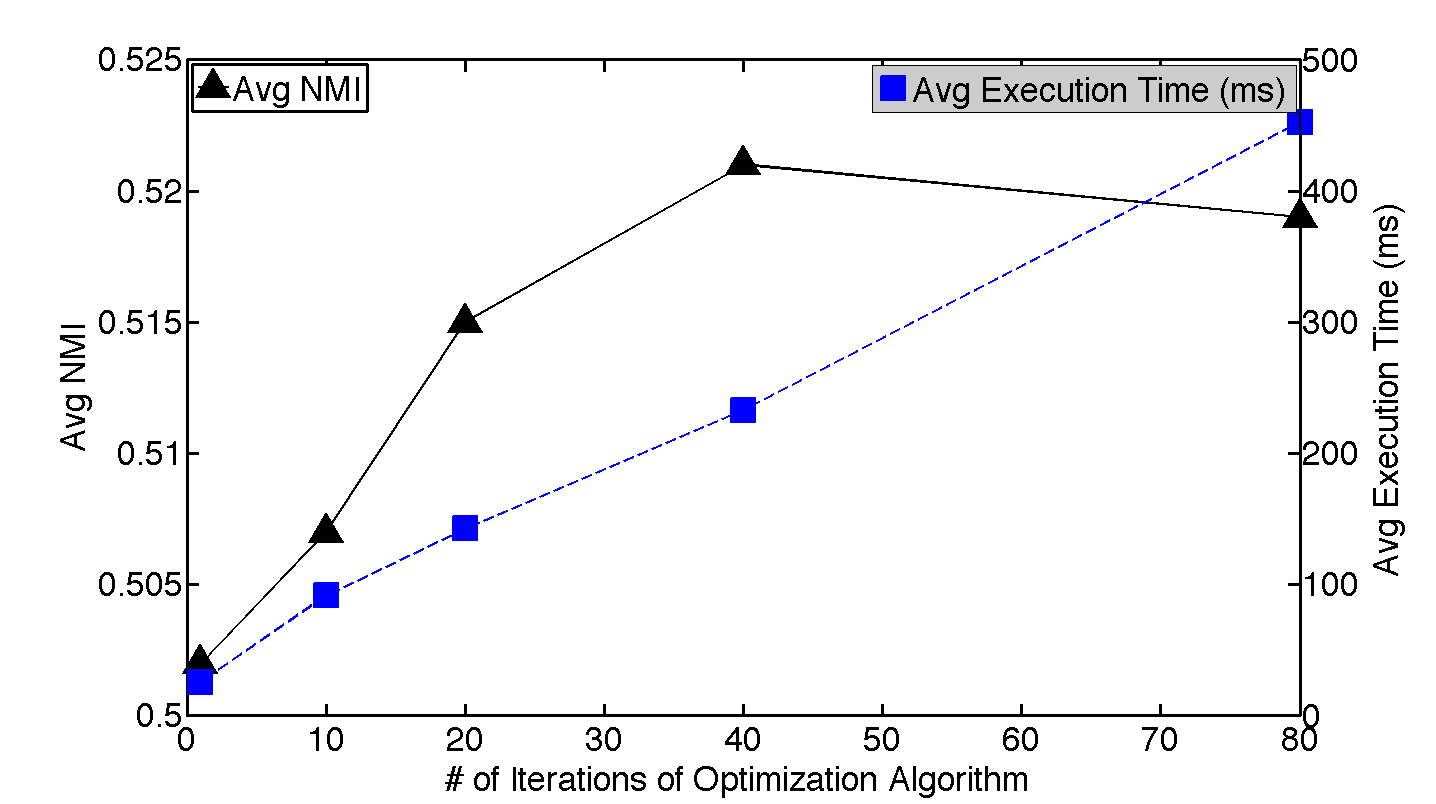}
}
\subfloat[{\scriptsize ``CHINC + YAGO2'' for ECAT.}]
{
\label{fig:ecatyg_clus_iter}
\includegraphics[width=0.50\textwidth]{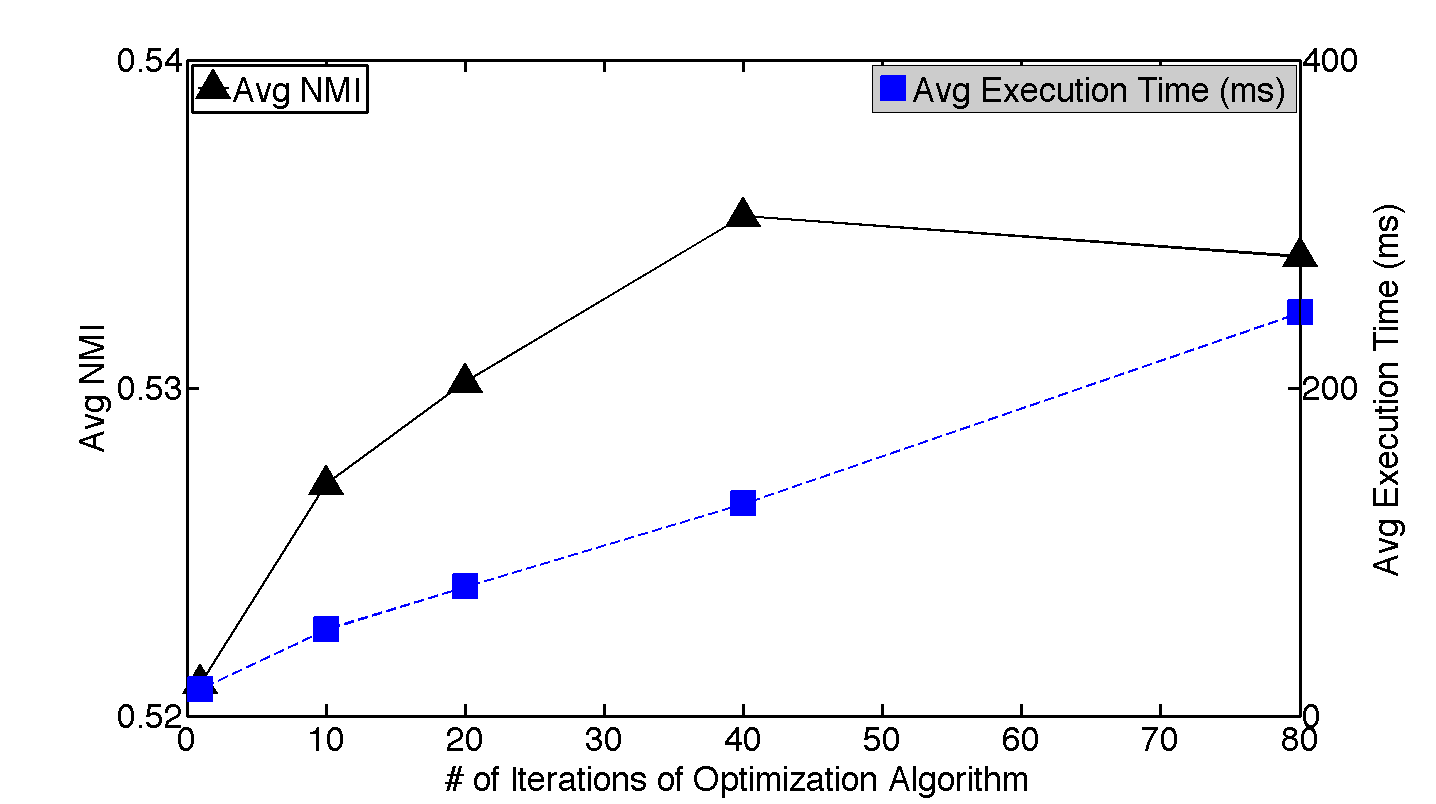}
}
\caption{{Analysis of \# of iterations in alternating optimization algorithm on different datasets with YAGO2 as world knowledge source. Left $y$-axis: average NMI; Right $y$-axis: average execution time (ms).}}
\label{fig:CHINC_iter1}
\end{figure*}

We evaluate the impact of the number of iterations of the alternating optimization (Algorithm~\ref{Alg_triiHMRFITCC}) on CHINC in relation to the execution time of the optimization algorithm as well as the clustering performance. We increase the number of iterations from 1 to 80. For example, for each number of iterations, we run CHINC five trials, and the average execution time and NMI are summarized in Figures~\ref{fig:CHINC_iter}--\ref{fig:CHINC_iter1}. From the result, one can conclude that the larger number of iterations is, the more significant the improvement on clustering performance. This improvement eventually drops, tapers out, and becomes stable.
The reason is that, with the increase of the number of iterations, the alternating optimization algorithm comes to covergence. However, the execution time still increase in a nearly linear manner. For example, as shown in Figure~\ref{fig:alg_iter}, after reaching $20$, the performance stays stable. Thus, we set the number of iterations as $20$ in the remaining experiments with the consideration of both performance and efficiency. As shown in Figure~\ref{fig:mcat_clus_iter}--\ref{fig:ecatyg_clus_iter}, we set the number of iterations as $20$ when conducting experiments on the other combinations of document datasets and world knowledge bases.

\subsubsection{Analysis of Specified World Knowledge based Constraints}
\label{subsec:const}
Rather than using human knowledge as constraints, we use the specified world knowledge automatically generated by our approach as constraints in CHINC. Based on the specified world knowledge, it is straightforward to design constraints for entities.

{\it Entity constraints.} {\bf (1) Must-links.} If two entities belong to the same entity sub-type, we add a must-link. For example, the entity sub-types of ``Obama'' and ``Bush'' are both {\it Politician}, we thus build a must-link between them. {\bf (2) Cannot-links.} If two entities belong to different entity sub-types, we add a cannot-link. For example, the entity sub-types of ``Obama'' and ``United States'' are {\it Politician} and {\it Country} respectively. In this case, we add a cannot-link to them.

\begin{figure*}[htbp]
\centering
\subfloat[{\scriptsize Constraints of type \#1 of ``CHINC + Freebase'' for 20NG.}]
{
\label{fig:CHINC_ent1_cons}
\includegraphics[width=0.32\textwidth]{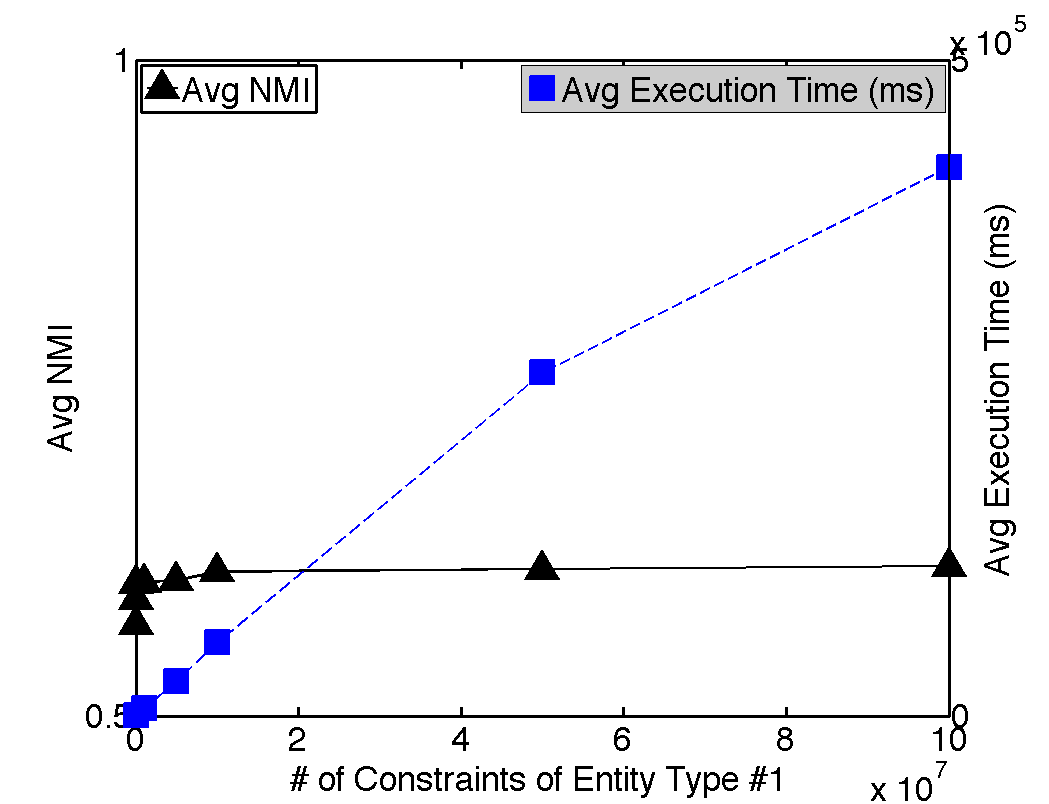}
}
\subfloat[{\scriptsize Constraints of types \#1+\#2 of ``CHINC + Freebase'' for 20NG.}]
{
\label{fig:CHINC_ent12_cons}
\includegraphics[width=0.32\textwidth]{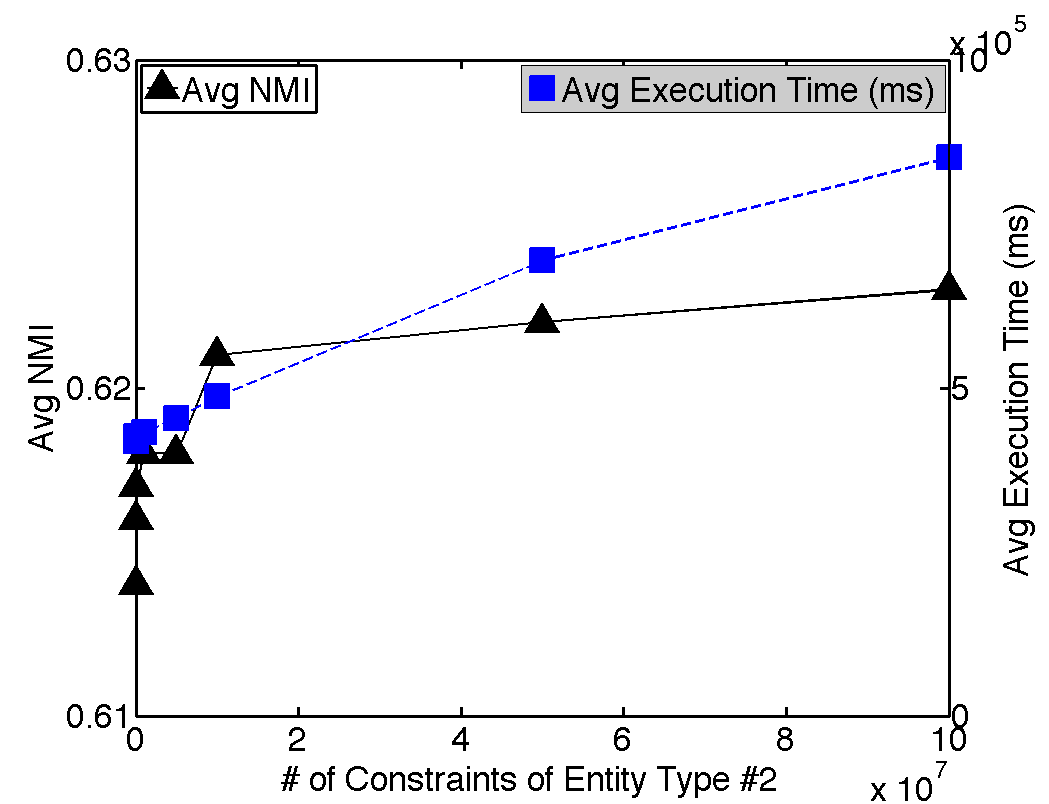}
}
\subfloat[{\scriptsize Constraints of types \#1+\#2+\#3 of ``CHINC + Freebase'' for 20NG.}]
{
\label{fig:CHINC_ent123_cons}
\includegraphics[width=0.32\textwidth]{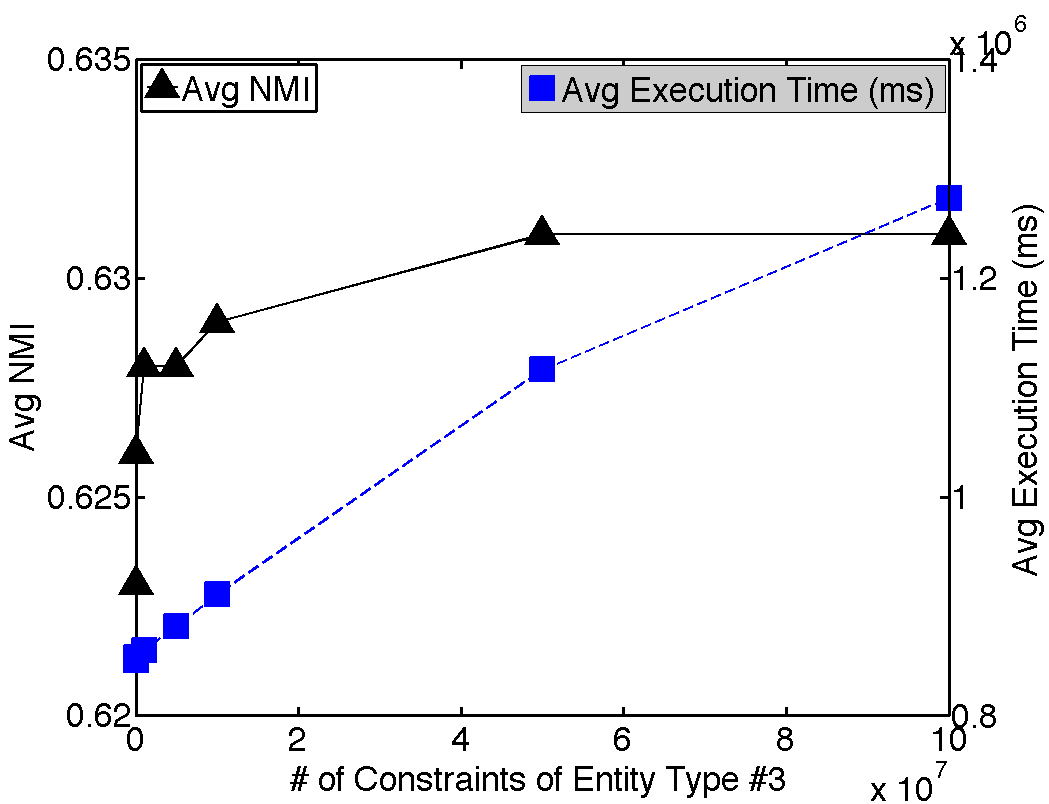}
}
\\
\subfloat[{\scriptsize Constraints of type \#1 of ``CHINC + Freebase'' for MCAT.}]
{
\label{fig:CHINC_mcat_ent1_cons}
\includegraphics[width=0.32\textwidth]{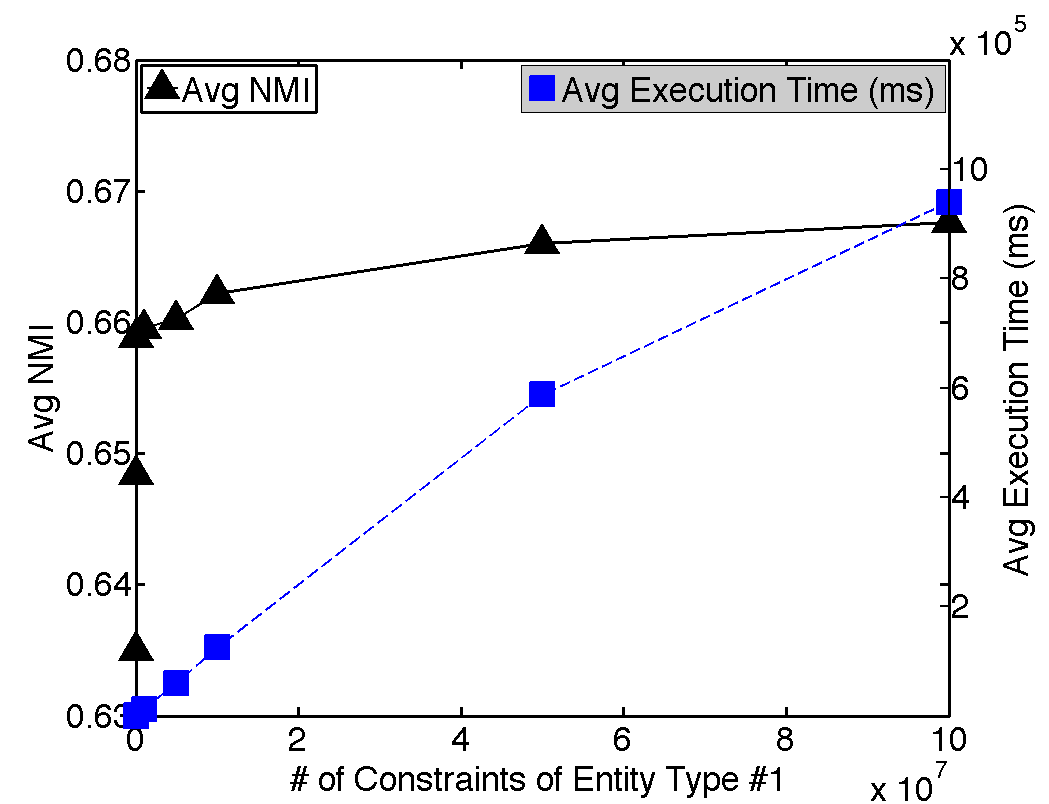}
}
\subfloat[{\scriptsize Constraints of types \#1+\#2 of ``CHINC + Freebase'' for MCAT.}]
{
\label{fig:CHINC_mcat_ent12_cons}
\includegraphics[width=0.32\textwidth]{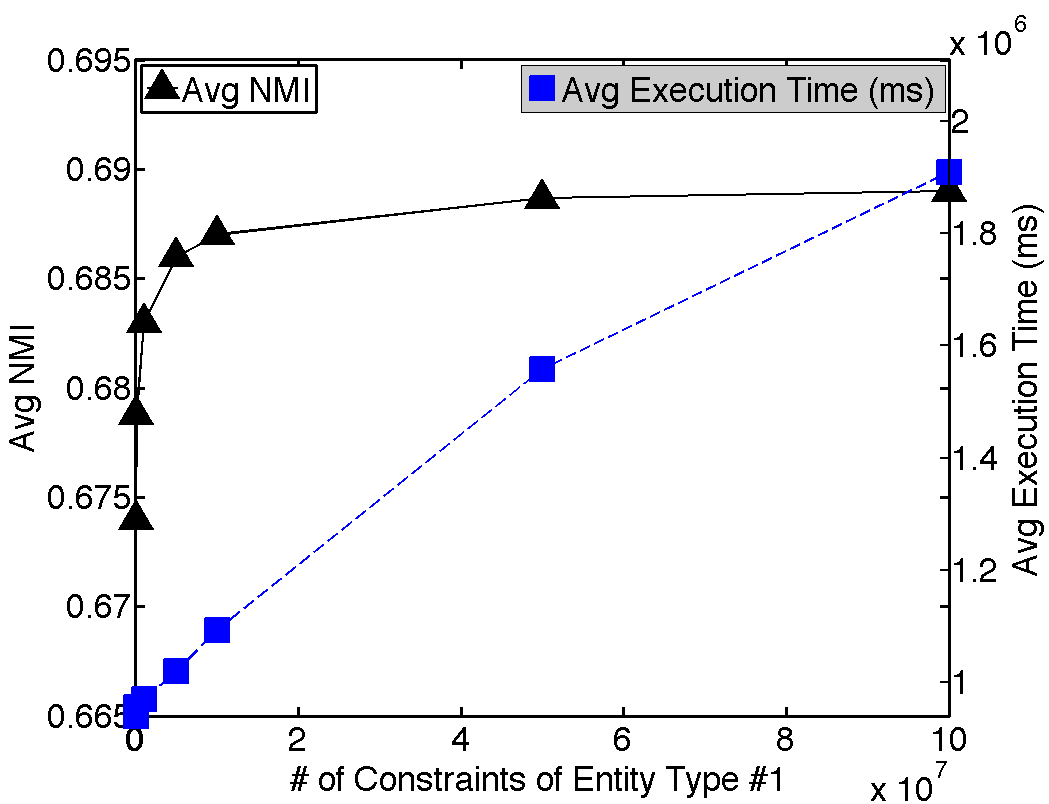}
}
\subfloat[{\scriptsize Constraints of types \#1+\#2+\#3 of ``CHINC + Freebase'' for MCAT.}]
{
\label{fig:CHINC_mcat_ent123_cons}
\includegraphics[width=0.32\textwidth]{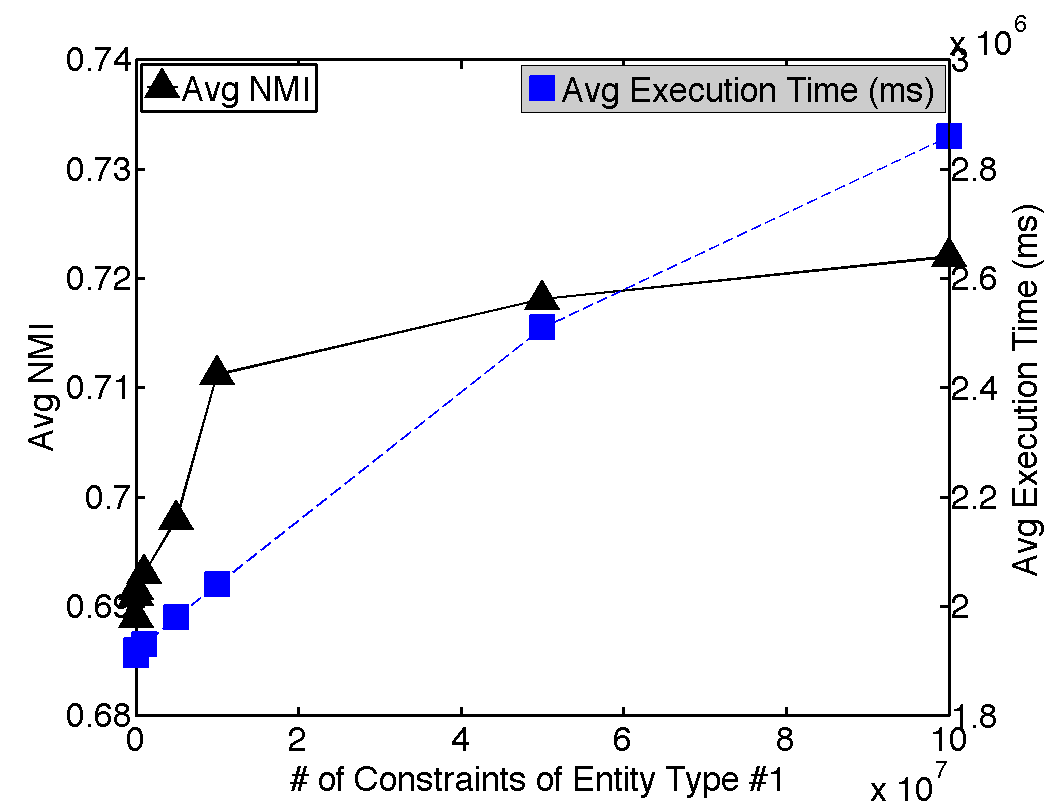}
}
\\
\subfloat[{\scriptsize Constraints of type \#1 of ``CHINC + Freebase'' for CCAT.}]
{
\label{fig:CHINC_ccat_ent1_cons}
\includegraphics[width=0.32\textwidth]{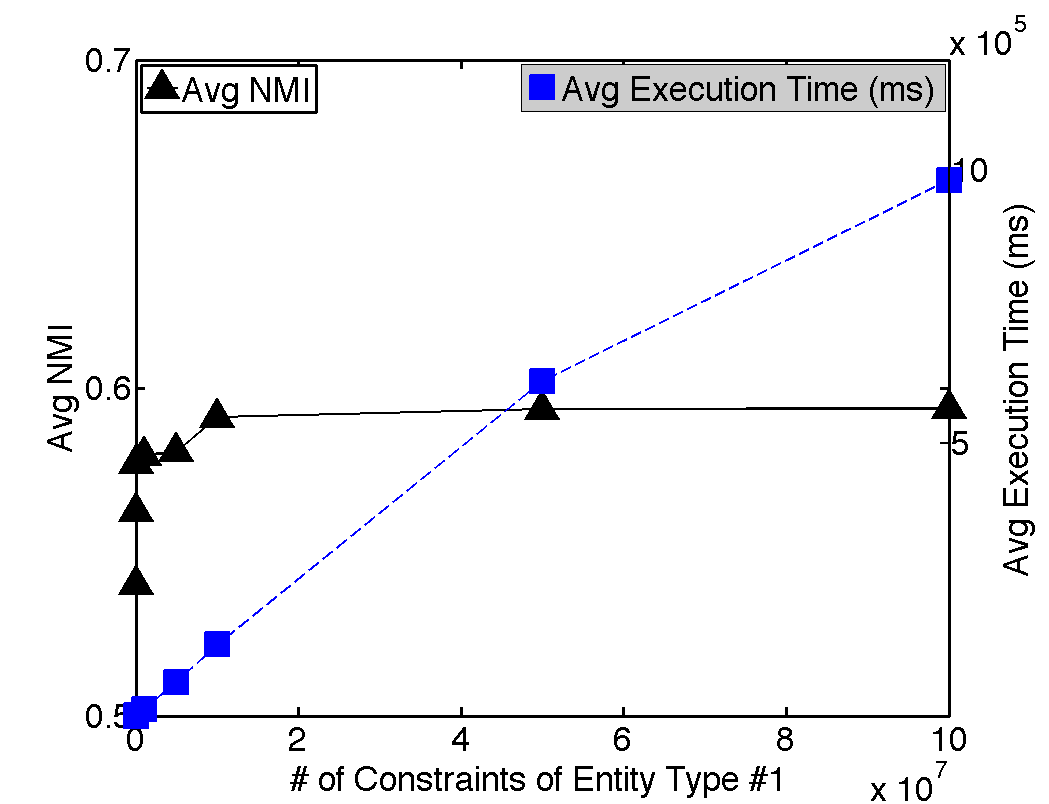}
}
\subfloat[{\scriptsize Constraints of types \#1+\#2 of ``CHINC + Freebase'' for CCAT.}]
{
\label{fig:CHINC_ccat_ent12_cons}
\includegraphics[width=0.32\textwidth]{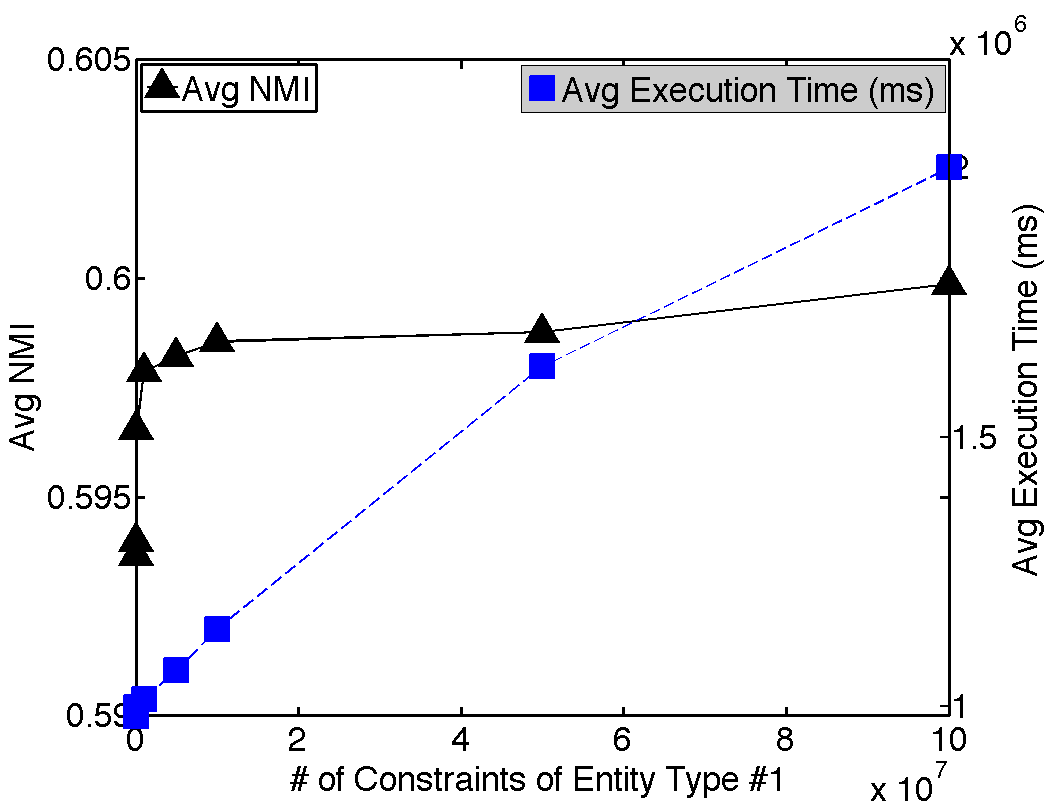}
}
\subfloat[{\scriptsize Constraints of types \#1+\#2+\#3 of ``CHINC + Freebase'' for CCAT.}]
{
\label{fig:CHINC_ccat_ent123_cons}
\includegraphics[width=0.32\textwidth]{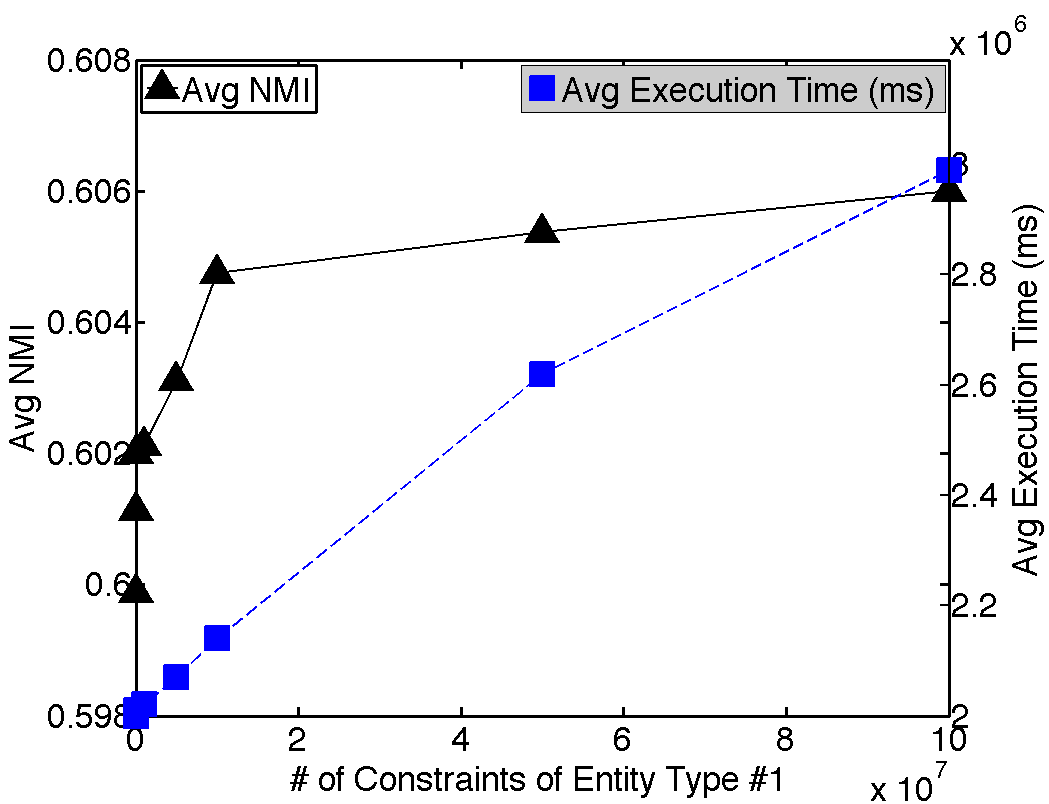}
}
\\
\subfloat[{\scriptsize Constraints of type \#1 of ``CHINC + Freebase'' for ECAT.}]
{
\label{fig:CHINC_ecat_ent1_cons}
\includegraphics[width=0.32\textwidth]{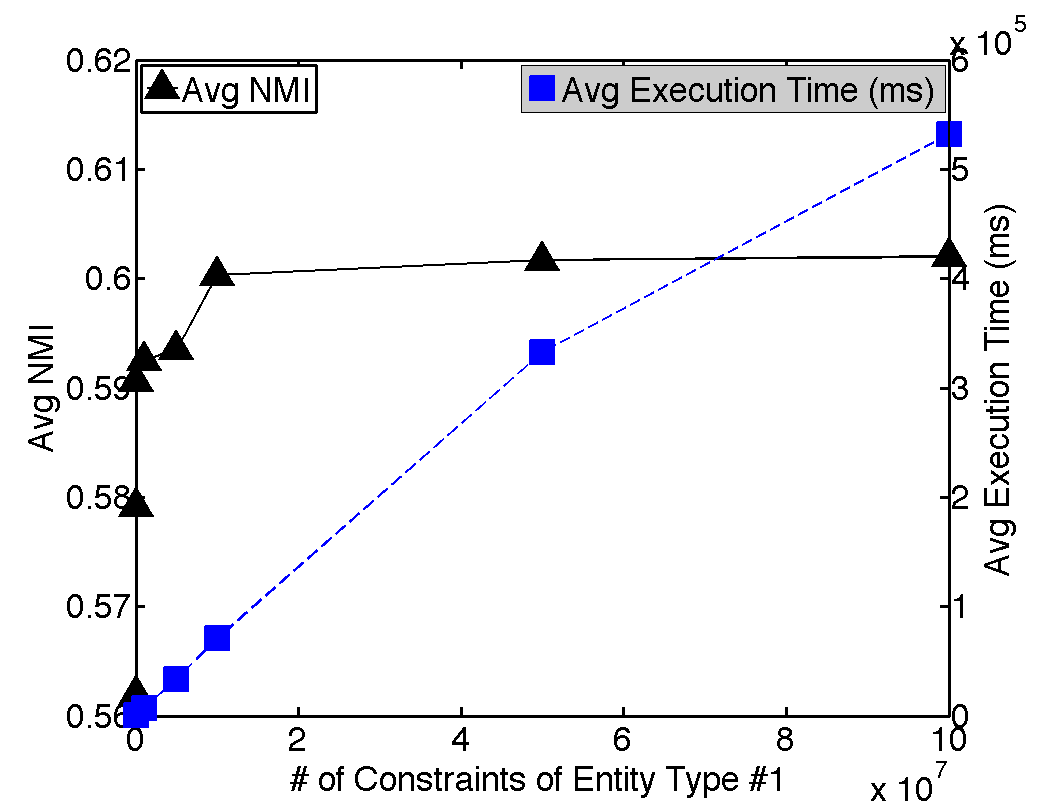}
}
\subfloat[{\scriptsize Constraints of types \#1+\#2 of ``CHINC + Freebase'' for ECAT.}]
{
\label{fig:CHINC_ecat_ent12_cons}
\includegraphics[width=0.32\textwidth]{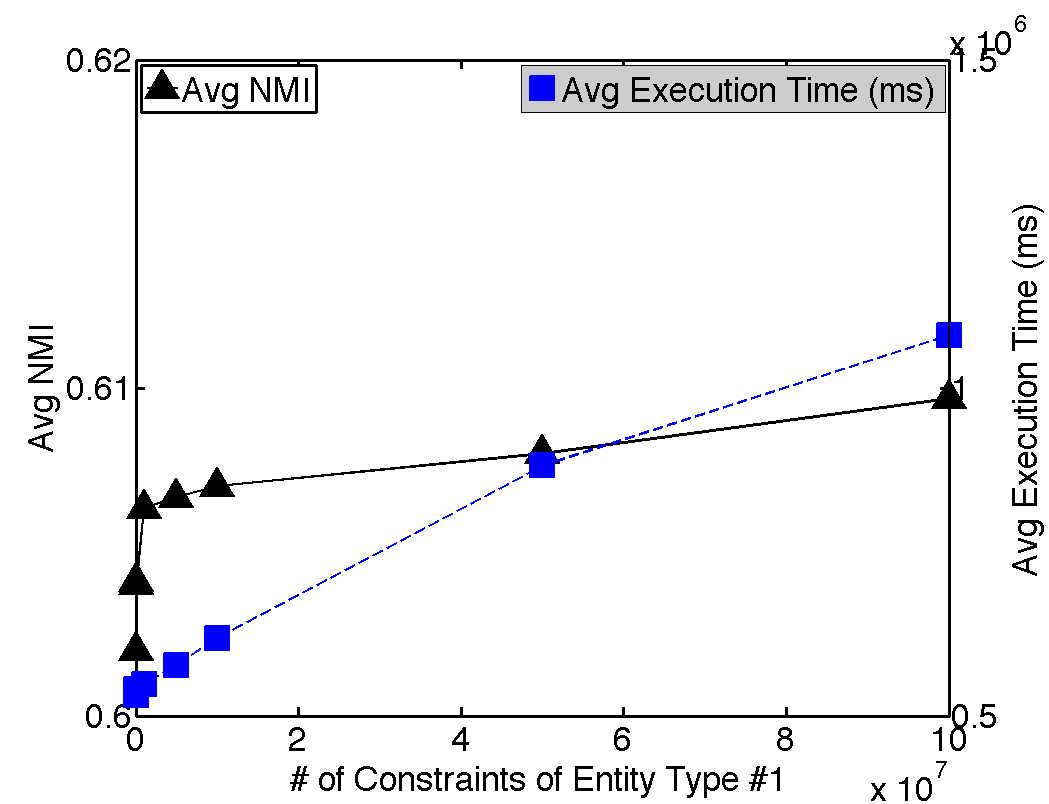}
}
\subfloat[{\scriptsize Constraints of types \#1+\#2+\#3 of ``CHINC + Freebase'' for ECAT.}]
{
\label{fig:CHINC_ecat_ent123_cons}
\includegraphics[width=0.32\textwidth]{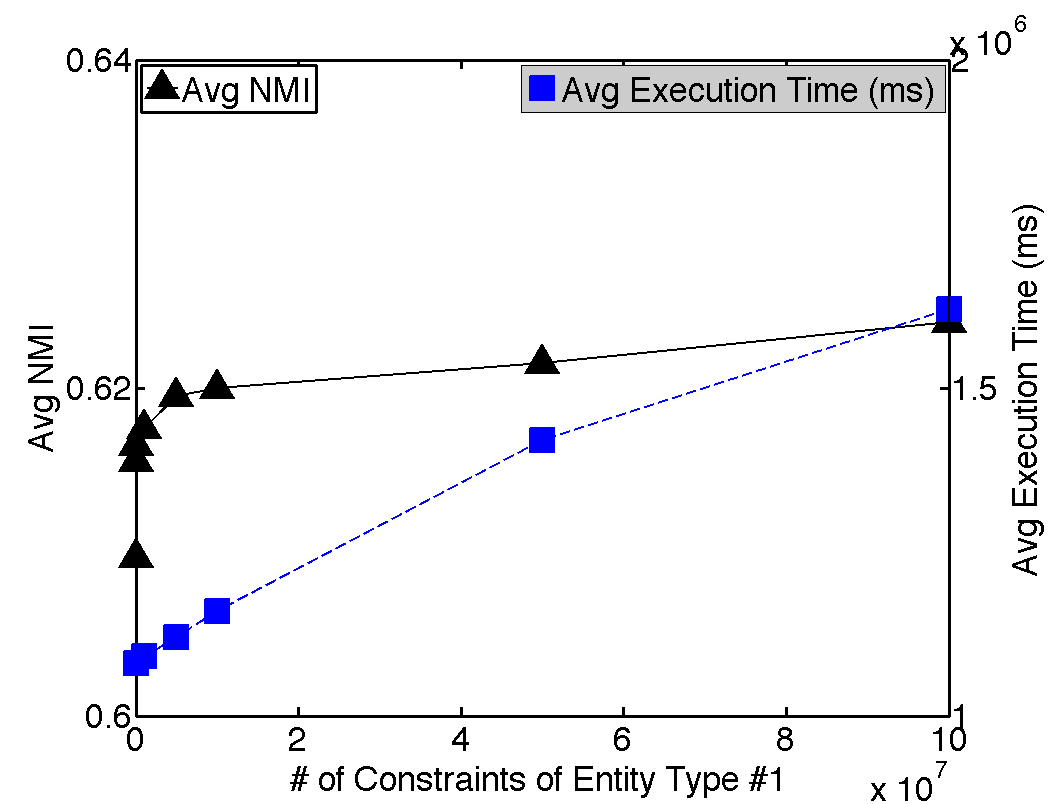}
}
\caption{{Effects of entity constraints and world knowledge source (Freebase). Left $y$-axis: average NMI; Right $y$-axis: average execution time (ms).}}
\label{fig:CHINC_ent_cons1}
\end{figure*}

\begin{figure*}[htbp]
\centering
\subfloat[{\scriptsize Constraints of type \#1 of ``CHINC + YAGO2'' for 20NG.}]
{
\label{fig:CHINC_ngyg_ent1_cons}
\includegraphics[width=0.32\textwidth]{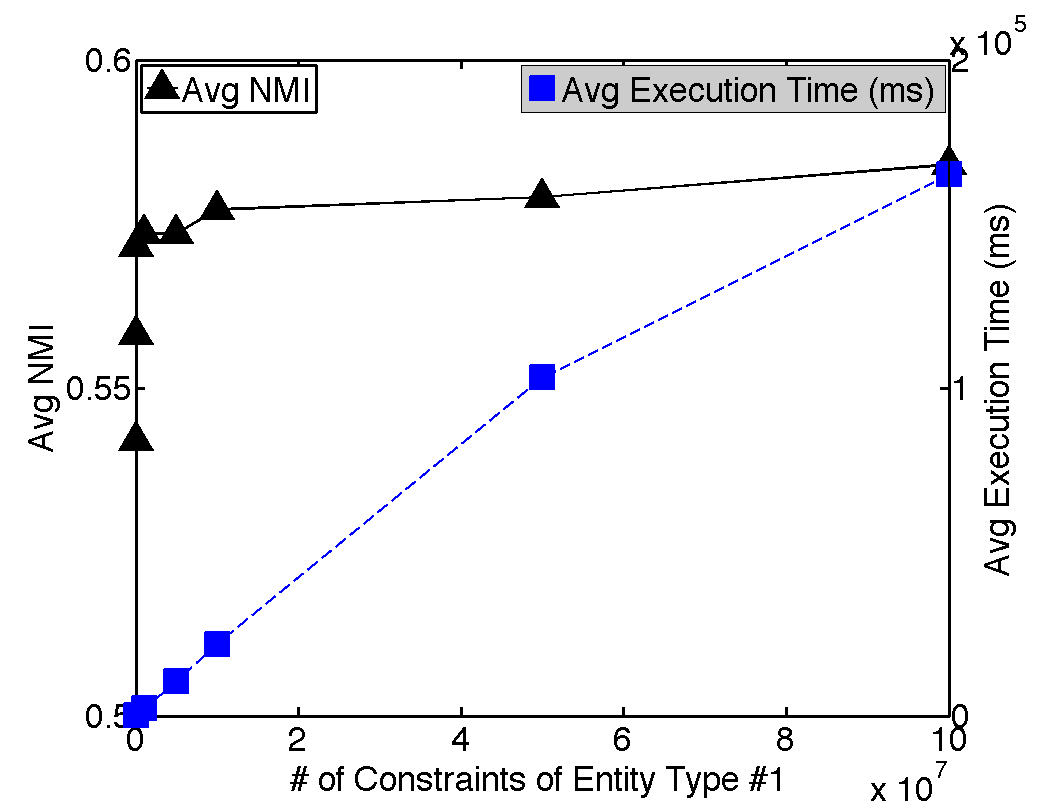}
}
\subfloat[{\scriptsize Constraints of types \#1+\#2 of ``CHINC + YAGO2'' for 20NG.}]
{
\label{fig:CHINC_ngyg_ent12_cons}
\includegraphics[width=0.32\textwidth]{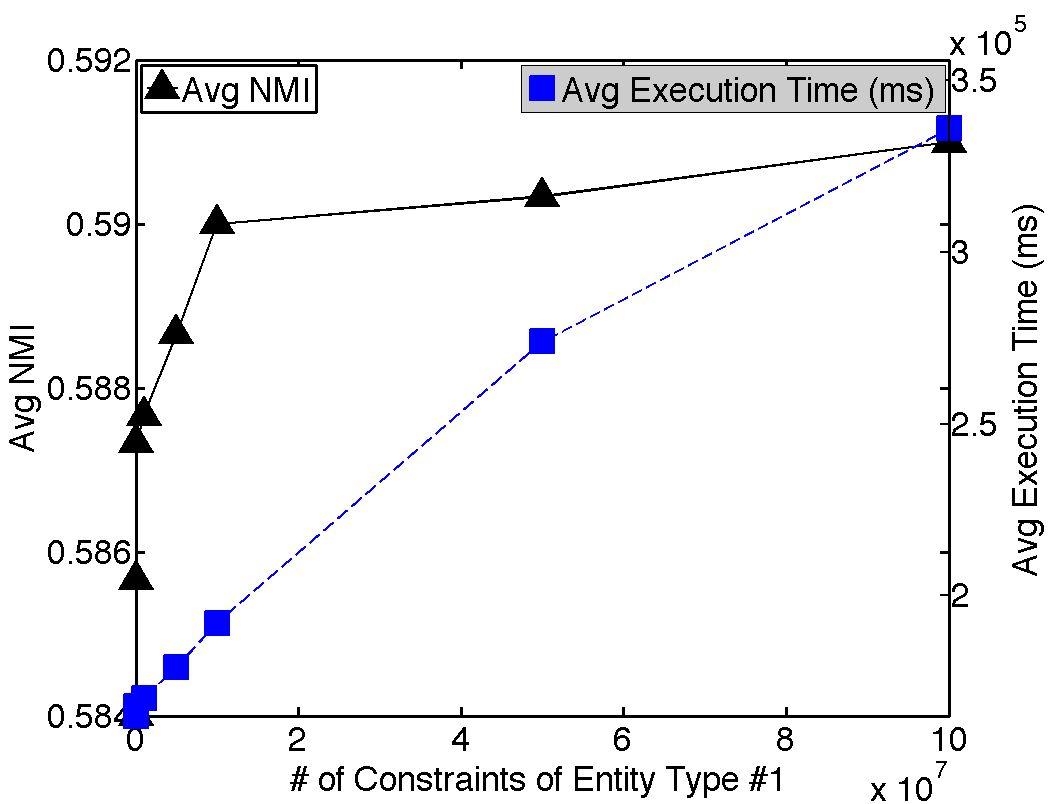}
}
\subfloat[{\scriptsize Constraints of types \#1+\#2+\#3 of ``CHINC + YAGO2'' for 20NG.}]
{
\label{fig:CHINC_ngyg_ent123_cons}
\includegraphics[width=0.32\textwidth]{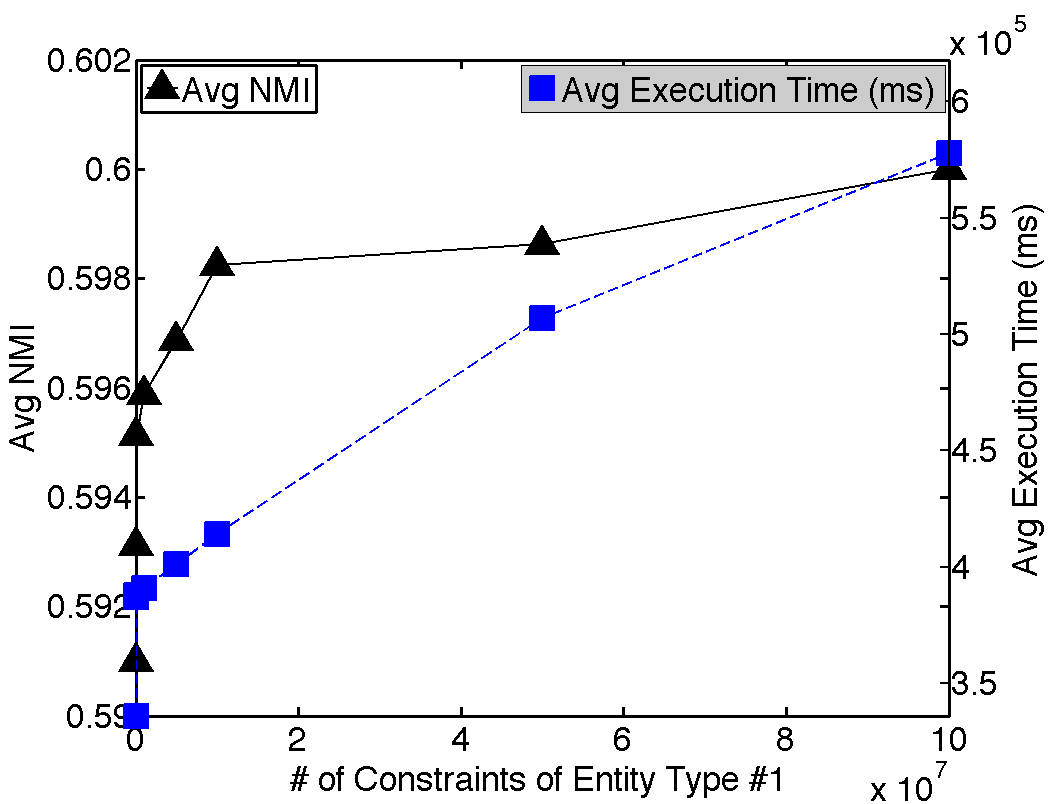}
}
\\
\subfloat[{\scriptsize Constraints of type \#1 of ``CHINC + YAGO2'' for MCAT.}]
{
\label{fig:CHINC_mcatyg_ent1_cons}
\includegraphics[width=0.32\textwidth]{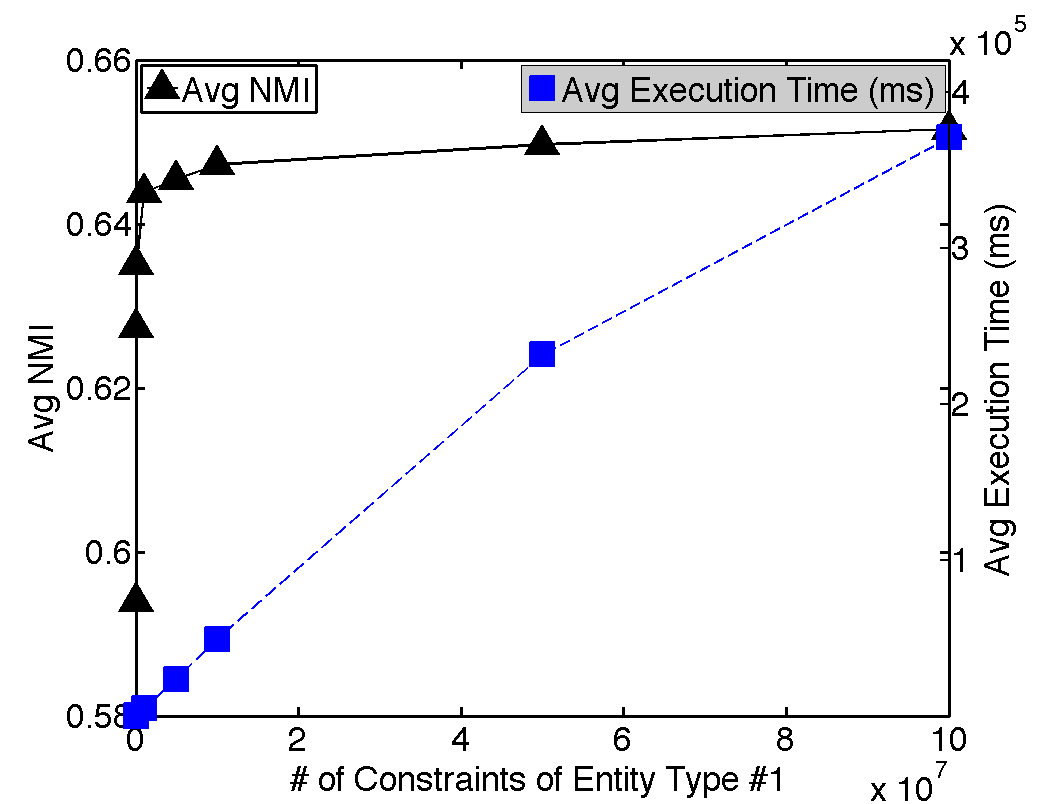}
}
\subfloat[{\scriptsize Constraints of types \#1+\#2 of ``CHINC + YAGO2'' for MCAT.}]
{
\label{fig:CHINC_mcatyg_ent12_cons}
\includegraphics[width=0.32\textwidth]{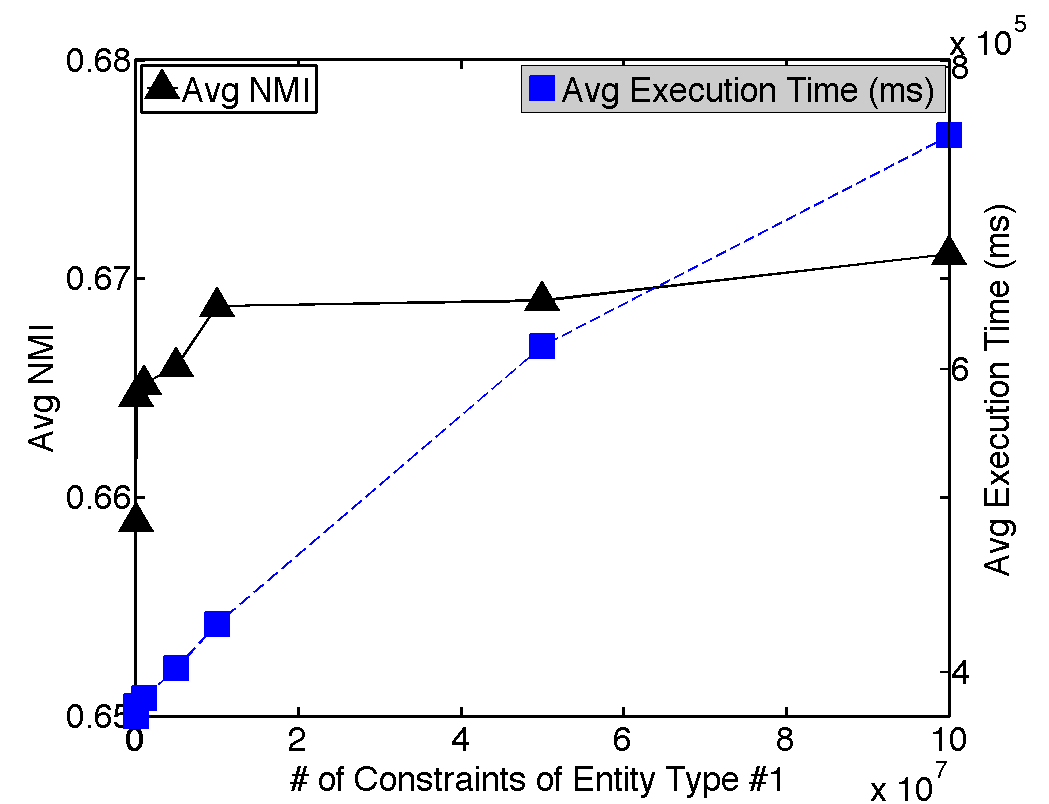}
}
\subfloat[{\scriptsize Constraints of types \#1+\#2+\#3 of ``CHINC + YAGO2'' for MCAT.}]
{
\label{fig:CHINC_mcatyg_ent123_cons}
\includegraphics[width=0.32\textwidth]{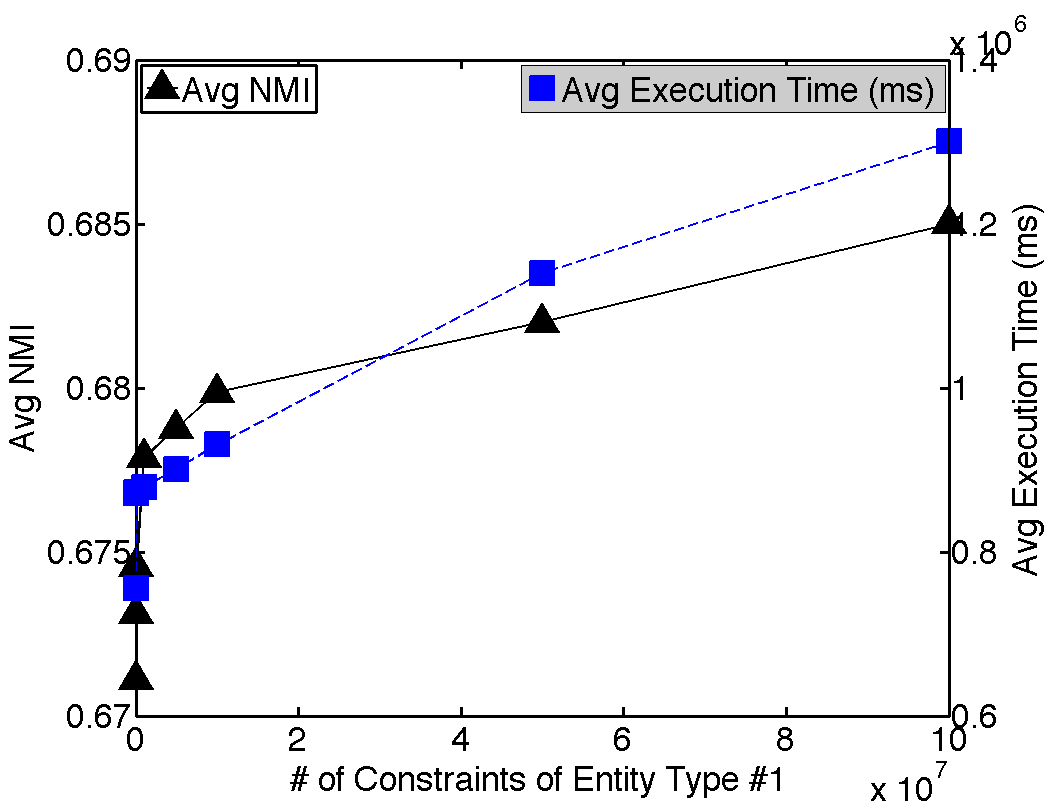}
}
\\
\subfloat[{\scriptsize Constraints of type \#1 of ``CHINC + YAGO2'' for CCAT.}]
{
\label{fig:CHINC_ccatyg_ent1_cons}
\includegraphics[width=0.32\textwidth]{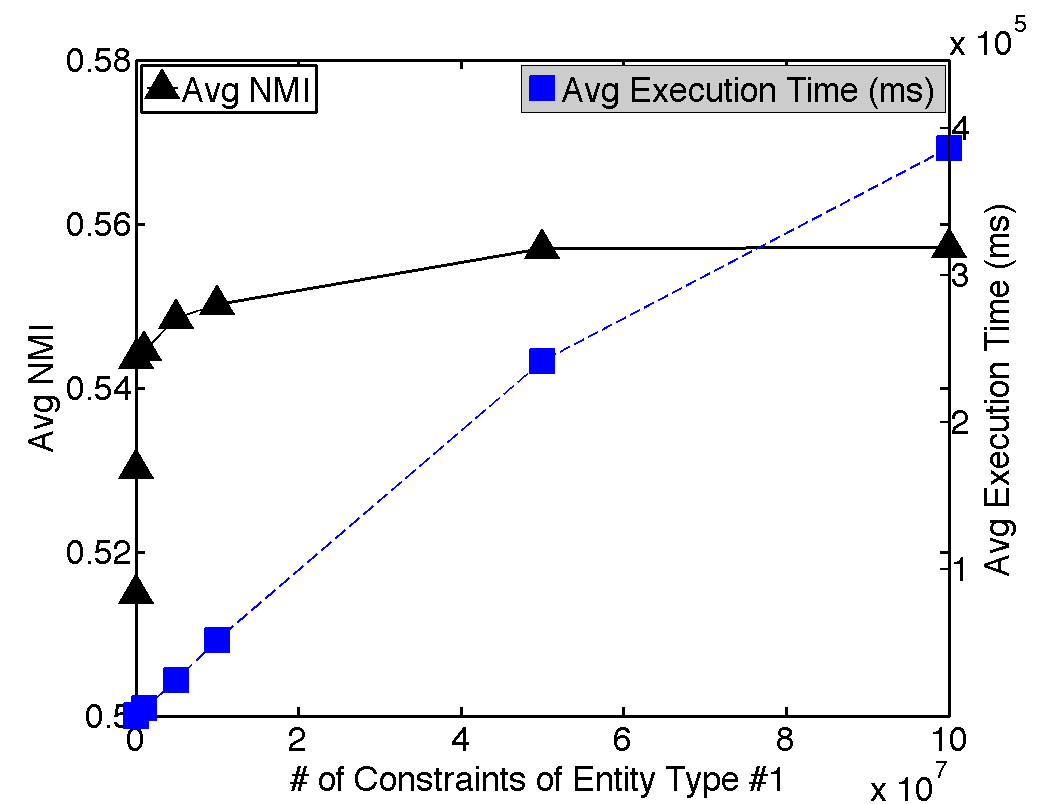}
}
\subfloat[{\scriptsize Constraints of types \#1+\#2 of ``CHINC + YAGO2'' for CCAT.}]
{
\label{fig:CHINC_ccatyg_ent12_cons}
\includegraphics[width=0.32\textwidth]{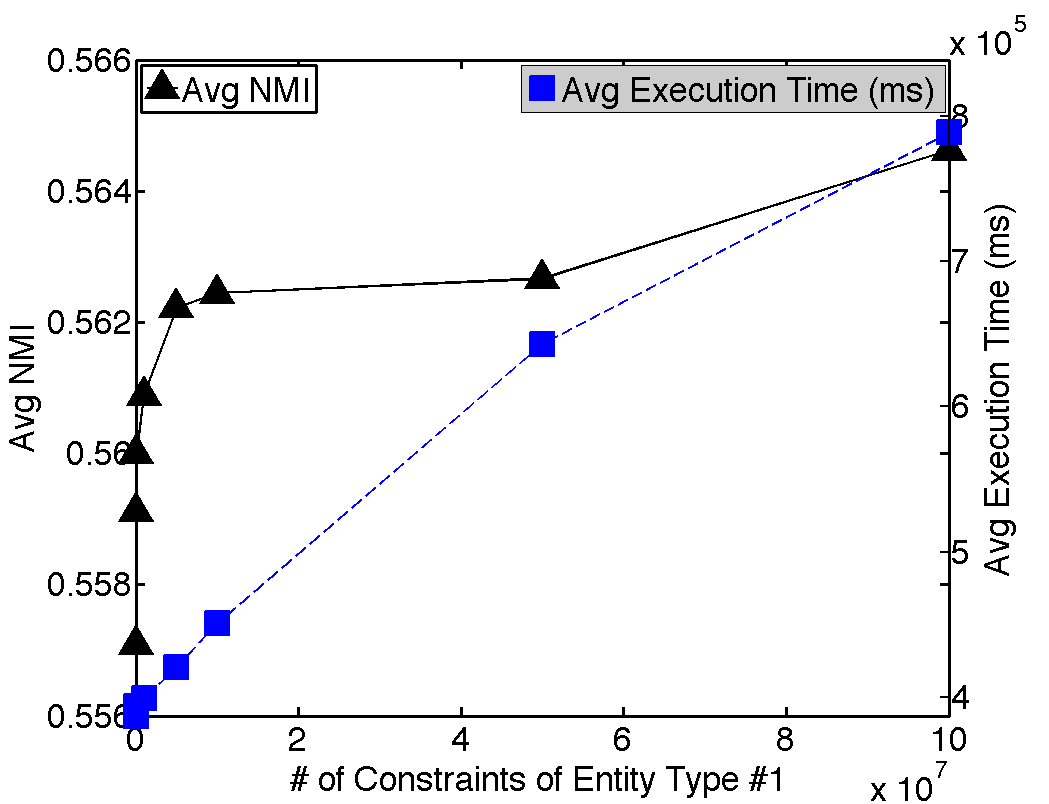}
}
\subfloat[{\scriptsize Constraints of types \#1+\#2+\#3 of ``CHINC + YAGO2'' for CCAT.}]
{
\label{fig:CHINC_ccatyg_ent123_cons}
\includegraphics[width=0.32\textwidth]{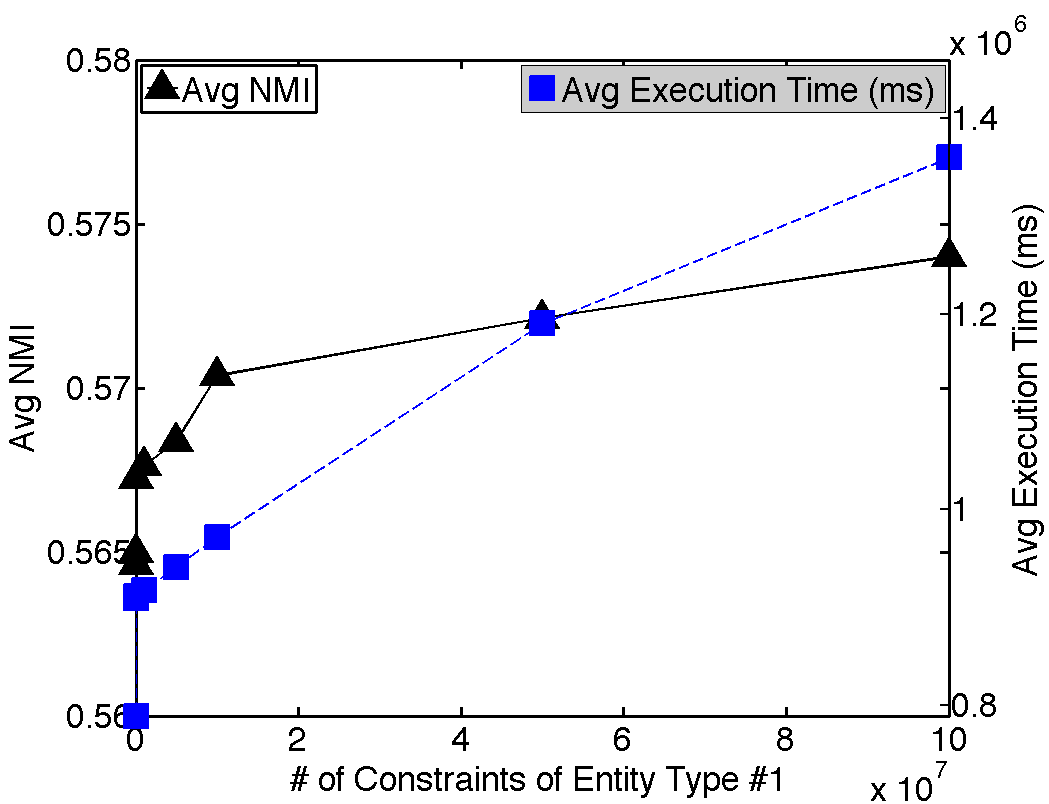}
}
\\
\subfloat[{\scriptsize Constraints of type \#1 of ``CHINC + YAGO2'' for ECAT.}]
{
\label{fig:CHINC_ecatyg_ent1_cons}
\includegraphics[width=0.32\textwidth]{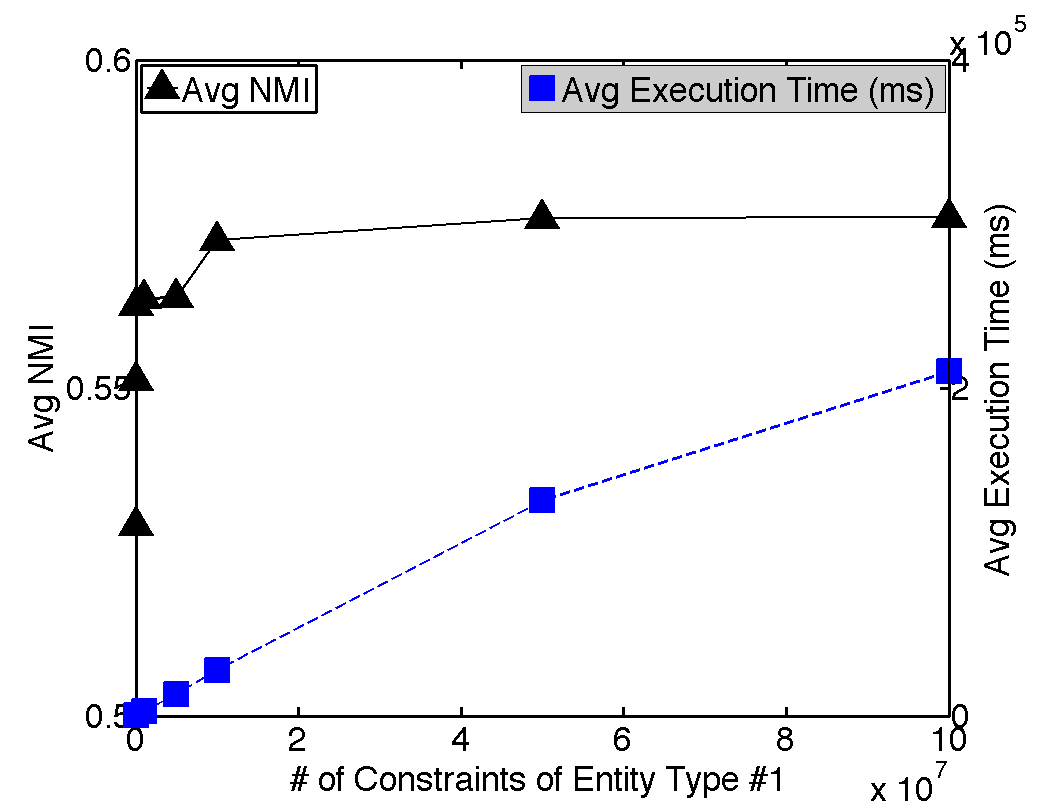}
}
\subfloat[{\scriptsize Constraints of types \#1+\#2 of ``CHINC + YAGO2'' for ECAT.}]
{
\label{fig:CHINC_ecatyg_ent12_cons}
\includegraphics[width=0.32\textwidth]{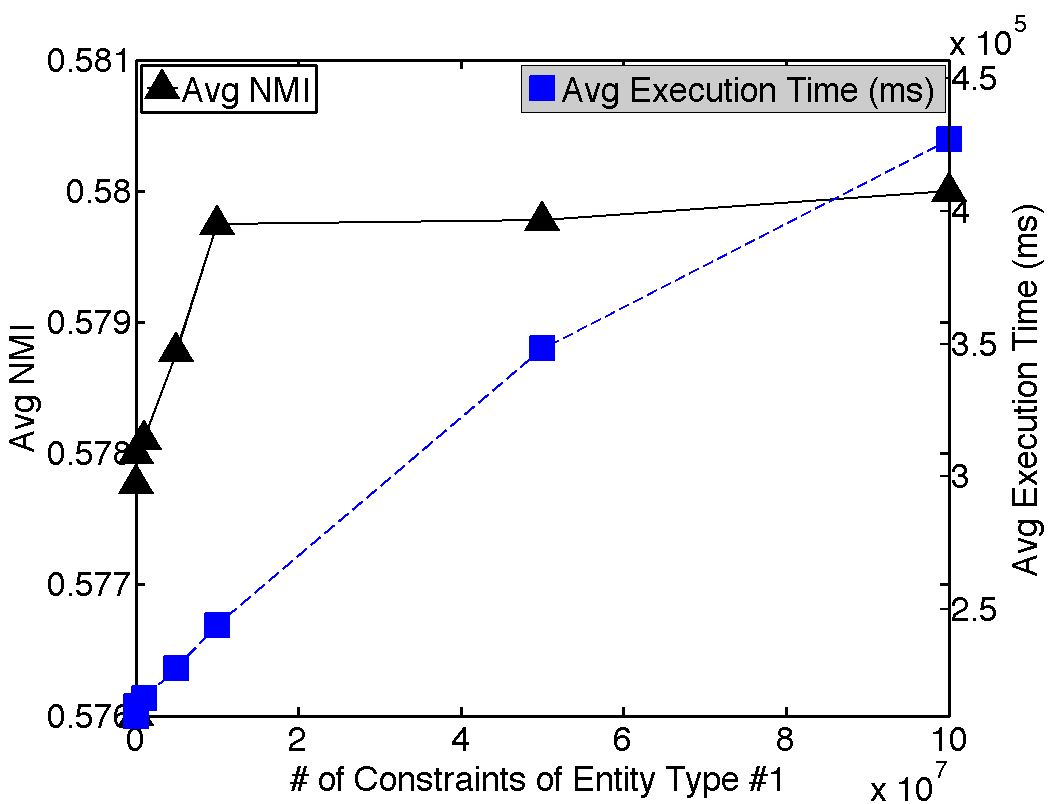}
}
\subfloat[{\scriptsize Constraints of types \#1+\#2+\#3 of ``CHINC + YAGO2'' for ECAT.}]
{
\label{fig:CHINC_ecatyg_ent123_cons}
\includegraphics[width=0.32\textwidth]{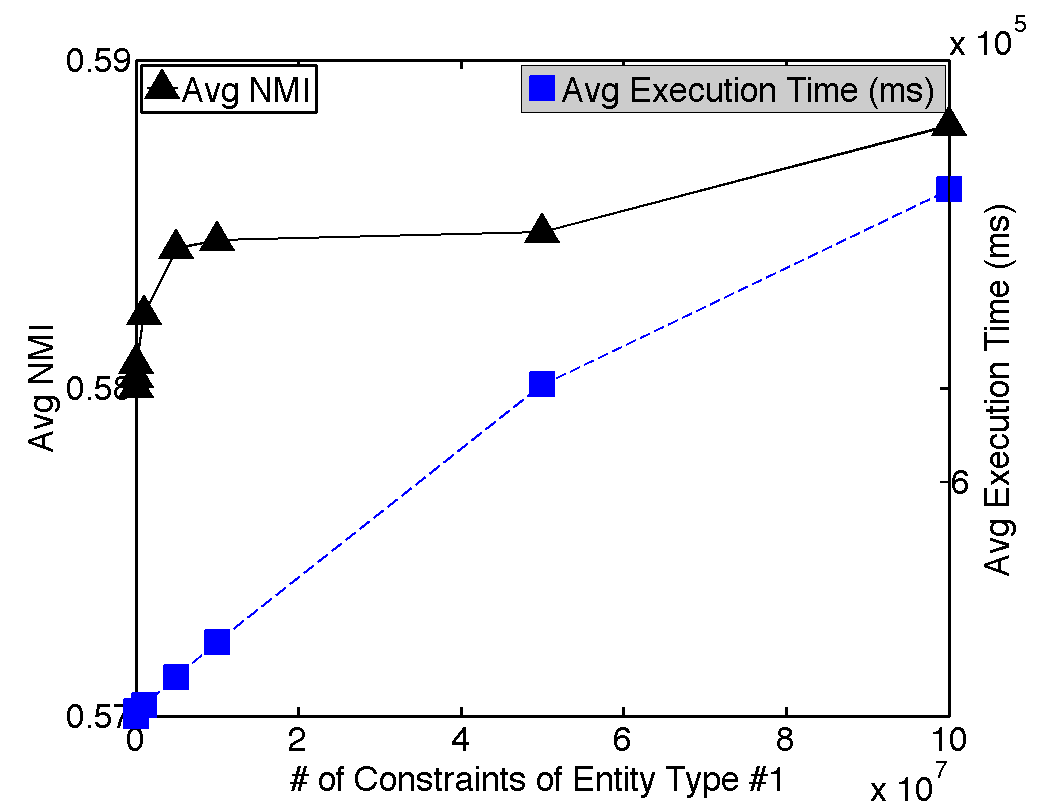}
}
\caption{{Effects of entity constraints and world knowledge source (YAGO2). Left $y$-axis: average NMI; Right $y$-axis: average execution time (ms).}}
\label{fig:CHINC_ent_cons3}
\end{figure*}

\nop{{\color{red} We then test the performance of our proposed CHINC by using none human knowledge used in the above clustering algorithm, but use the domain-specific knowledge as constraints in the algorithm. I cannot understand this.}
}
We then test the performance of our proposed CHINC by using the specified world knowledge as constraints described above.
We show the experiments on all of the different combinations of datasets and world knowledge sources in Figure~\ref{fig:CHINC_ent_cons1}--\ref{fig:CHINC_ent_cons3}. Each $x$-axis represents the number of entity type constraints used in each experiment, and $y$-axis is the average NMI of five random trials. For example, the constraints derived from entity type \#1, \#2, and \#3 are eventually added to CHINC as shown in Figure~\ref{fig:CHINC_ent1_cons}, Figure~\ref{fig:CHINC_ent12_cons} and Figure~\ref{fig:CHINC_ent123_cons} respectively, when using Freebase as world knowledge and testing on 20NG dataset.
We can see that CHINC outperforms the best clustering algorithm with the human knowledge as shown in Table~\ref{tab:clus_result} (CITCC: $0.569$) with even no constraints (HINC: $0.571$). By adding more and more constraints, the clustering result of CHINC is significantly better.
So CHINC is able to use information in world knowledge specified in the HIN, and the entity sub-type information can be transferred to the document side. The results show the power of modeling data as heterogeneous information networks, as well as the high quality of constraints derived from world knowledge.

From Figures~\ref{fig:CHINC_ent_cons1}--\ref{fig:CHINC_ent_cons3}, by increasing the number of constraints, we find that the average execution time of five trials increases linearly, and the clustering performance measured by NMI is increasing as mentioned before. For example, Figure~\ref{fig:CHINC_ent123_cons} shows the effects of the constraints of all the three entity types on the clustering performance as well as the execution time, when Freebase is used as world knowledge and CHINC is tested on 20NG dataset. After the number of constraints reach around 50M, the increase of performance drops and stays stable. At this point, the execution time is around 1.2M (ms). In Figure~\ref{fig:CHINC_mcat_ent1_cons}--\ref{fig:CHINC_ecatyg_ent123_cons}, we can see the similar results on the other combinations of document datasets and knowledge bases. We also find that the average execution time of our algorithm with Freebase as world knowledge source is greater than that with YAGO2. As shown in Figure~\ref{fig:expcn}, the reason is that each document datasets with Freebase could be specified much more entities than that with YAGO2. From the results, we can see that our algorithm is scalable to use the large scale specified world knowledge as constraints, and cluster large amounts of documents.

\begin{figure}[h]
\centering
\includegraphics[scale=0.20]{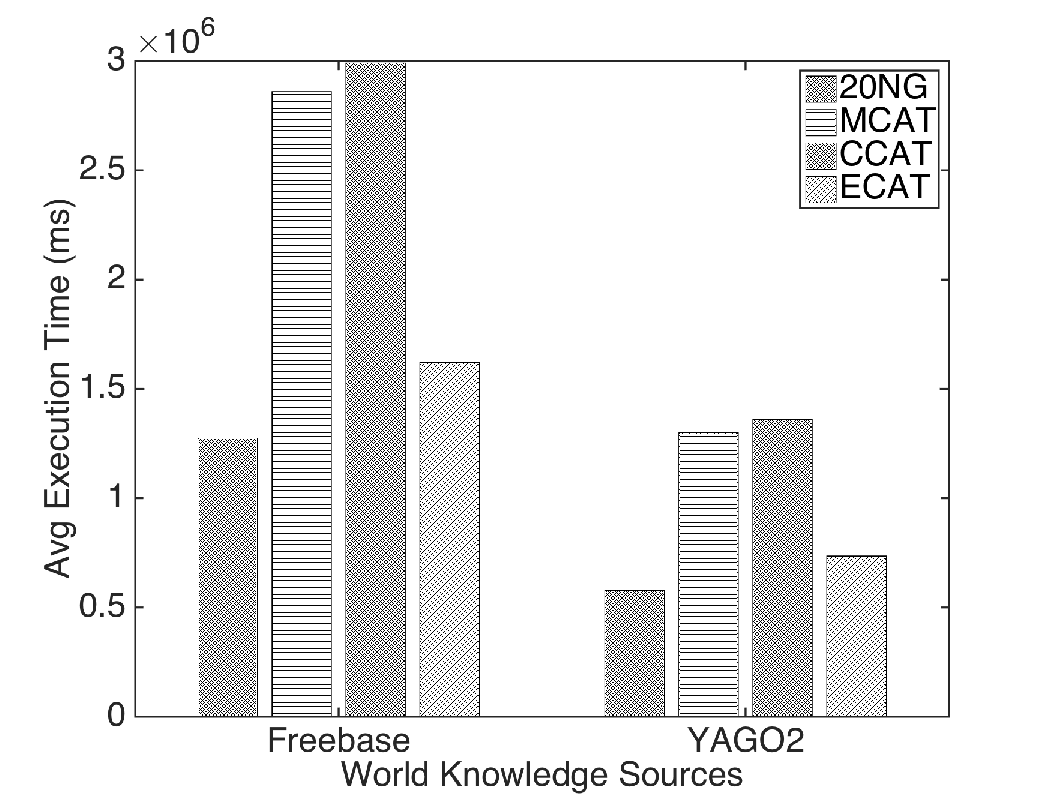}
\vspace{-0.1in}
\caption{Analysis of the efficiency of our algorithm on different document datasets with different world knowledge sources.}
\label{fig:expt}
\end{figure}

\nop{
\begin{figure}[h]
\centering
\includegraphics[width=0.4\textwidth]{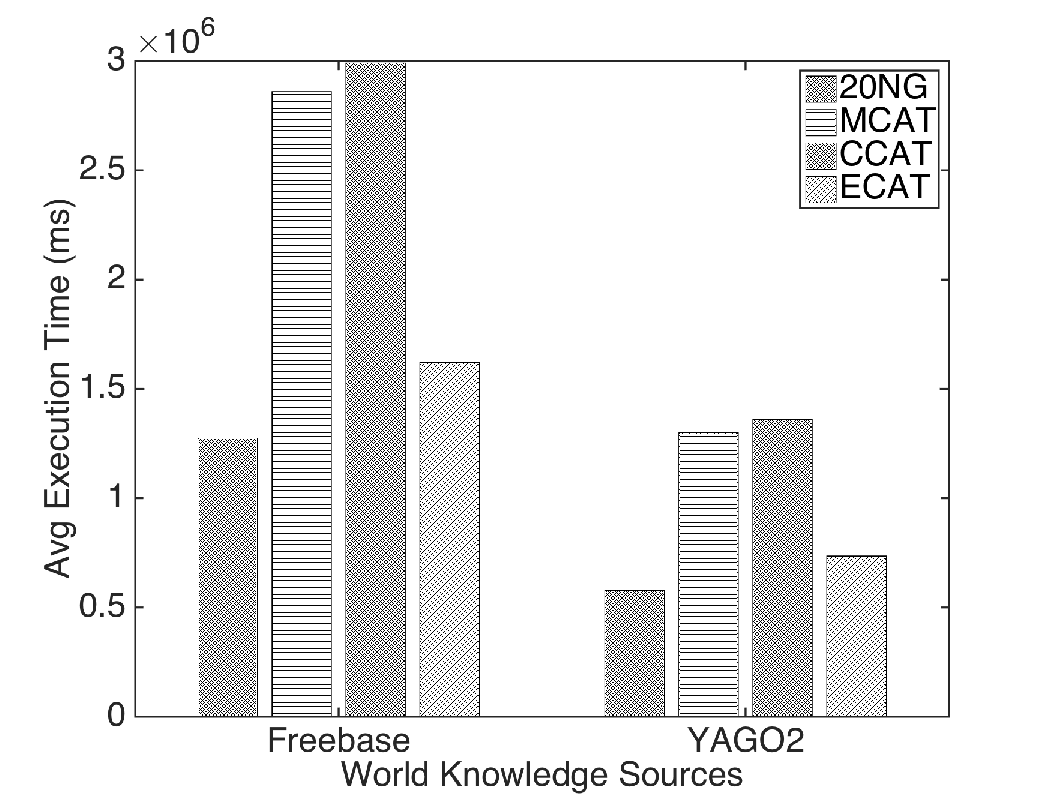}
\vspace{-0.1in}
\caption{Analysis of the efficiency of our algorithm on different document datasets with different world knowledge sources.}
\label{fig:expt}
\end{figure}
}

\nop{
\subsubsection{User Study of the Clustering Results}
We show the clusters generated by CHINC with the comparison with the K-means, HINC, \etc to show the insight why our method outperforms the other methods.

By showing the above user study of the clustering results, we find that our discovered constraints are very useful to overcome some errors that the baseline clustering algorithms encountered.
}

\nop{
Besides, we show the execution time and clustering performance when using YAGO or Freebase or WordNet as world knowledge source, we find that the NMI of YAGO or Freebase is much greater than using WordNet, even WordNet could use less execution time, but the increase of execution time is not significant.
}

\section{Related Work}
\label{sec:related}

In this section, we review the related work on document clustering, machine learning with domain and world knowledge, as well as heterogeneous information networks.

\subsection{Document Clustering}
Document clustering has been studied for many years.
We can use traditional one-dimensional clustering algorithms (e.g., Kmeans) to cluster the documents.
If we treat the document and corresponding words as a bipartite graph, we can use co-clustering algorithms~\cite{dhillon2003information} to cluster the documents.
Moreover, with the help of labeled seed documents, semi-supervised clustering can be used~\cite{Basu02}.
When the seeds are not available, we can use side information as constraints to guide clustering algorithms~\cite{basu2004probabilistic}.
When the supervision from target domain is not available, we can also perform transfer learning to transfer the labeled information from other domains to the target domain~\cite{Dai2007,Wang2009}.
All the above clustering algorithms with supervision need domain or relevant domain knowledge.
When there are diverse domains and the supervision is needed, they will still be very costly to ask a lot of different domain experts to label.

\subsection{Machine Learning with Domain Knowledge}
The general idea of incorporating domain knowledge into machine learning algorithms has already been studied extensively in natural language processing community. Chang, Ratinov and Roth~\cite{chang2012structured} presented constrained conditional models (CCMs) that allow to inject high-level knowledge as soft constraints into linear models, such as Hidden Markov Models and Structured Perceptron, for lots of natural language processing tasks, including information extraction~\cite{roth2004linear,roth2007global}, semantic role labeling~\cite{roth2005integer,punyakanok2008importance}, summarization~\cite{clarke2006constraint}, and co-reference resolution~\cite{denis2007joint}. Posterior Regularization (PR)~\cite{ganchev2010posterior} works on incorporating indirect supervision via constraints on posterior
distributions of probabilistic models with latent variables. The key difference between CCM based learning and PR is that CCMs allows the use of hard constraints, while PR uses expectation constraints. Samdani et al.~\cite{samdani2012unified} have proposed an unified Expectation-Maximization algorithm that can cover CCM based learning and PR. Besides, lots of models have been studied to incorporate the domain knowledge for better performance. Markov Logic Network (MLN)~\cite{richardson2006markov} is proposed to integrate first-order logic with Markov Random Field. A combination of the Bayesian network model with a collection of deterministic constraints is presented in~\cite{dechter2004mixtures}. However, all of the mentioned work use domain knowledge to improve the performance of the corresponding machine learning algorithms. Different from their work, we explore a more general learning framework with world knowledge. Given domain-dependent data, we aim to automatically generate domain knowledge by specifying the world knowledge, and represent the domain knowledge in an unified format for more general machine learning. It will be interesting to use our learning framework in more learning models, to conduct empirical experiments to compare to CCM and PR based models in various domain-dependent tasks.

{Besides, transfer learning~\cite{pan2010survey} is another direction on leveraging domain knowledge for better machine learning. The main idea of transfer learning is to leverage the domain knowledge in source domain to help the learning tasks in the target domain. The key intuition is that the labeled data is relatively easier to collect in the source domain than in the target domain. Unsupervised transfer learning~\cite{dai2008self} has been adapted for developing new clustering algorithms. Self-taught clustering~\cite{dai2008self} is an instance of unsupervised transfer learning, which aims at clustering a small collection of unlabeled data in the target domain with the help of a large amount of unlabeled data in the source domain. Besides unsupervised transfer learning, there are inductive transfer learning~\cite{dai2007boosting,mihalkova2007mapping} and transductive transfer learning~\cite{arnold2007comparative,ling2008spectral} for the problem of regression and classification. Recently, source-free transfer learning~\cite{lu2014source} and selective transfer learning~\cite{lu2013selective} have been proposed to address the text classification and cross domain recommendation problems, and shown improved results respectively. Lifelong learning~\cite{eaton2013ella,chen2014topic} is also a framework of machine learning with domain knowledge. Lifelong learning test~\cite{liaaai2015} has been proposed to take both the current performance and the performance growth rate into consideration. However, both transfer learning and lifelong learning are focusing on leveraging domain knowledge rather than world knowledge, whereas our framework is focusing on first how to specify the world knowledge to automatically generate domain knowledge and then modeling the learning task(s) with the help of the specified world knowledge. Again, it will be of great interests to build an end-to-end machine learning system by using the automatically generated domain knowledge while conducting transfer learning or lifelong learning.}

\subsection{Machine Learning with World Knowledge}
Most of the existing usage of world knowledge is to enrich the features beyond bag-of-words representation of documents.
For example, by using the linguistic knowledge base WordNet to resolve synonyms and introduce WordNet concepts, the quality of
document clustering can be improved~\cite{Hotho2003}.
The first paper using the term ``world knowledge'' \cite{gabrilovich2005feature} extends the bag-of-words features with the categories in Open Directory Project (ODP), and shows that it can help improve text classification with additional knowledge.
Following this, by mapping the text to the semantic space provided by Wikipedia pages or other ontologies, it has been proven to be useful for short text classification~\cite{gabrilovich2006overcoming,gabrilovich2007computing,GabrilovichM09}, clustering~\cite{HuJian2008,Hu2009EIE,HuXiaohua2009,Fodeh2011}, and information retrieval~\cite{EgoziMG11}.
After that, a bunch of research also uses another knowledge base of taxonomy, Probase~\cite{wu2011taxonomy}, to enrich the features of short text or keywords for classification~\cite{Wang14}, clustering~\cite{song2011short,song2015short}, information retrieval~\cite{Hua2013,Song14,wangrelsim}, or mines knowledge from text for information retrieval~\cite{wang2013paraphrasing}.
All of the above approaches just consider to use world knowledge as a source of features.
However, the knowledge in the knowledge bases indeed has annotations of types, categories, etc..
Thus, it can be more effective to consider this information as ``supervision'' to supervise other machine learning algorithms and tasks.
Along this research direction, recent work~\cite{wang2015sim,wang2016text} apply our world knowledge enabled machine learning framework for the document similarity computation and classification tasks.

Distant supervision uses the knowledge of entities and their relationships from world knowledge bases, e.g., Freebase, as supervision for the task of entity and relation extraction~\cite{mintz2009distant,wang2014knowledge,xu2014rc}. It considers to use knowledge supervision to extract more entities and relations from new text or to generate a better embedding of entities and relations.
Thus, the application of direct supervision is limited to entities and relations themselves.

Song et al.~\cite{song2013constrained} consider using fully unsupervised method to generate constraints of words using an external general-purpose knowledge base, WordNet.
This can be regarded as an initial attempt to use general knowledge as indirect supervision to help clustering.
However, the knowledge from WordNet is mostly linguistically related.
It lacks of the information about named entities and their types.
Moreover, their approach is still a simple application of constrained co-clustering, where it misses the rich structural information in the knowledge base.

\subsection{Heterogeneous Information Network}
A heterogeneous information network (HIN) is defined as a graph of multi-typed entities and relations~\cite{Han2010MKD,sun2012mining}.
Different from traditional graphs, HIN incorporates the type information which can be useful to identify the semantic meaning of the paths in the graph~\cite{Yizhou11}.
This is a good property to perform graph search and matching~\cite{he2006closure,yan2004graph,zhang2007treepi}.
Original HINs are developed for the applications of scientific publication network analysis~\cite{zhao2009p,sun2011co,Yizhou11,SunNHYYY12}.
Then social network analysis also leverages this representation for user similarity and link prediction~\cite{KongZY13,ZhangKY13,ZhangKY14}.
Seamlessly, we can see that the knowledge in world knowledge bases, e.g., Freebase and YAGO2, can be naturally represented as an HIN, since the entities and relations in the knowledge base are all typed.
We introduce this representation to knowledge based analysis, and show that it can be very useful for our document clustering task.
Note that there is also a series of methods called multi-type relational data clustering~\cite{Long2006,Long2007} and collective matrix factorization~\cite{singh2008relational,nickel2011three,bouchard2013convex,klami2013group}.
While they require the data to be structural beforehand (e.g., providing information of authors, co-authors, etc.), our method only needs the input of raw documents.
In addition to the multi-type relational information, we also incorporate the type information provided by the knowledge base as constraints to further improve the clustering results.

\section{Conclusion}
\label{sec:con}
In this paper, we study a novel problem of machine learning with world knowledge. Particularly, we take document clustering as an example and show how to use world knowledge as indirect supervision to improve the clustering results. To use the world knowledge, we show how to adapt the world knowledge to domain dependent tasks by using semantic parsing and semantic filtering. Then we represent the data as a heterogeneous information network, and use a constrained network clustering algorithm to obtain the document clusters. We demonstrate the effectiveness and efficiency of our approach on two real datasets along with two popular knowledge bases. In the future, we plan to use world knowledge to help more text mining and text analytics tasks, such as text classification and information retrieval.

\section*{Acknowledgments}

Chenguang Wang gratefully acknowledges the support by the National Natural Science Foundation of China (NSFC Grant Number 61472006) and the National Basic Research Program (973 Program No. 2014CB340405). The research is also partially supported by the Army Research Laboratory (ARL) under agreement W911NF-09-2-0053, and by DARPA under agreement number FA8750-13-2-0008. Research is also partially sponsored by China National 973 project 2014CB340304, U.S. National Science Foundation IIS-1320617, IIS-1354329 and IIS 16-18481, HDTRA1-10-1-0120, and grant 1U54GM114838 awarded by NIGMS through funds provided by the trans-NIH Big Data to Knowledge (BD2K) initiative (www.bd2k.nih.gov), and MIAS, a DHS-IDS Center for Multimodal Information Access and Synthesis at UIUC.
The views and conclusions contained herein are those of the authors and should not be interpreted as necessarily representing the official policies or endorsements, either expressed or implied by these agencies or the U.S. Government.

{
\bibliographystyle{ACM-Reference-Format-Journals}
\bibliography{sigproc}
}


\medskip
\end{document}